\documentclass{article}

\usepackage[preprint]{neurips_2025}

\usepackage[utf8]{inputenc} 
\usepackage[T1]{fontenc}    
\usepackage{hyperref}       
\usepackage{url}            
\usepackage{booktabs}       
\usepackage{amsfonts}       
\usepackage{amsmath}
\usepackage{nicefrac}       
\usepackage{microtype}      
\usepackage{xcolor}         
\usepackage{cleveref}
\usepackage{comment}
\usepackage{enumitem}
\usepackage{graphicx}
\usepackage{amsthm}
\usepackage{subcaption}

\newtheorem{theorem}{Theorem}
\newtheorem{lemma}{Lemma}

\newtheorem{definition}{Definition}

\newtheorem{proposition}{Proposition}
\newtheorem{corollary}{Corollary}
\theoremstyle{remark}        
\newtheorem*{remark}{Remark} 

\newcommand{\E}{\mathbb{E}}

\newcommand{\R}{\mathbb{R}}

\newcommand{\sS}{\mathcal{S}}
\newcommand{\sA}{\mathcal{A}}

\newcommand{\sD}{\mathcal{D}}

\newcommand{\Tr}{\mathrm{Tr}}

\DeclareMathOperator*{\argmax}{arg\,max}
\DeclareMathOperator*{\argmin}{arg\,min}

\usepackage[svgnames]{xcolor}

\definecolor{lxs}{RGB}{200,0,0}

\title{KL-regularization Itself is Differentially Private in Bandits and RLHF}

\author{%
  Yizhou Zhang \\
  California Institute of Technology \\
  Pasadena, CA 91125 \\
  \texttt{yzhang8@caltech.edu} \\
  \And
  Kishan Panaganti \\
  California Institute of Technology \\
  Pasadena, CA 91125 \\
  \texttt{kpb@caltech.edu} \\
  \And
  Laixi Shi \\
  California Institute of Technology \\
  Pasadena, CA 91125 \\
  \texttt{laixis@caltech.edu} \\
  \And
  Juba Ziani \\
  Georgia Institute of Technology \\
  Atlanta, GA 30332 \\
  \texttt{jziani3@gatech.edu} \\
  \And
  Adam Wierman \\
  California Institute of Technology \\
  Pasadena, CA 91125 \\
  \texttt{adamw@caltech.edu} \\
}

\begin{document}

\maketitle

\begin{abstract}
Differential Privacy (DP) provides a rigorous framework for privacy, ensuring the outputs of data-driven algorithms remain statistically indistinguishable across datasets that differ in a single entry. While guaranteeing DP generally requires explicitly injecting noise either to the algorithm itself or to its outputs, the intrinsic randomness of existing algorithms presents an opportunity to achieve DP ``for free''. In this work, we explore the role of regularization in achieving DP across three different decision-making problems: multi-armed bandits, linear contextual bandits, and reinforcement learning from human feedback (RLHF), in offline data settings. We show that adding KL-regularization to the learning objective (a common approach in optimization algorithms) makes the action sampled from the resulting stochastic policy itself differentially private. This offers a new route to privacy guarantees without additional noise injection, while also preserving the inherent advantage of regularization in enhancing performance.
\end{abstract}

\section{Introduction}
Data-driven methods are becoming an increasingly central part of our society. Online and user data now fuels ad auctions, recommender systems, credit scoring, and many other high-impact applications \citep{FTC2024SocialMedia, fu2025privacy, Fed2024ConsumerCompliance}. Often, such data is private or sensitive, and the volume of personal data driving today's algorithms has increased by orders of magnitude. This makes strong privacy protections—capable of preventing the exposure of sensitive personal information—more crucial than ever~\citep{dinur2003revealing}. 

To address this challenge, Differential Privacy (DP) \citep{dwork2006calibrating} offers a rigorous mathematical framework with precise, quantifiable privacy guarantees. DP has been largely used across various high-stake domains, including government statistics and in particular the 2020 decennial release of the U.S. Census \citep{boyd2019differential2020}, online user data (e.g., Google, Apple, Meta, LinkedIn all use DP in several of their products) \citep{desfontaines2021realworlddp}, finance \citep{maniar2021differential}, cyber-physical systems \citep{hassan2019differential}, healthcare \citep{dankar2013practicing}, energy data \citep{desfontaines2021realworlddp}, and client services \citep{erlingsson2014rappor}. Intuitively, DP provides an information-theoretic guarantee that ensures that the outputs of data-driven solutions remain (nearly) statistically indistinguishable whether or not any individual's data is included. As an example, when a foundation model is fine-tuned via reinforcement learning with human feedback (RLHF) using human labels, it risks memorizing and revealing sensitive personal details. DP guarantees that the model trained with and without a given individual's data are (almost) the same, which implies that the model does not memorize and output such sensitive data~\citep{zhang2025towards, wu2023privately}.

Differential Privacy (DP) inherently requires randomness in the output to prevent identifiable changes when the training data of a single agent varies. Achieving DP in various decision-making tasks usually relies on two main techniques. One is adding calibrated random noise (e.g., Laplace or Gaussian noise) to the training dataset, algorithm, or model parameters to obscure individual data contributions \citep{dwork2006calibrating, dwork2014algorithmic}. The other one, when it comes to loss minimization and utility maximization, is to sample a parameter with some probability that depends on how well this parameter performs---otherwise known as the \emph{Exponential Mechanism}. However, the introduction of either noise or sampling inevitably leads to accuracy and performance degradation, as the injected noise can obscure true data patterns, making it more difficult for algorithms to learn accurate models from the data \citep{chaudhuri2011differentially}. This creates significant challenges in balancing privacy and performance.

The goal of our work is to highlight a natural setting in which \emph{existing randomness in commonly used algorithms is already sufficient to obtain Differential Privacy}, and additional noise or sampling techniques may either not be needed, or may require adding less noise than previously believed. In particular, we focus on the role of \emph{KL-regularization} in machine learning. KL-regularization is a widely used technique in optimization and decision-making. It is naturally integrated into many such algorithms to improve performance by reducing computational costs or increasing the stability of the learning process~\citep{schulman2017proximal,schulman2015trust,haarnoja2018soft,cuturi2013sinkhorn}. In RL settings, for example,  absent regularization, optimal policies tend to be deterministic, yet including regularization induces policies that are inherently stochastic and randomized.

In this work, we investigate whether, and to what extent, sampling from the optimal policy resulting from optimizing a KL-regularized objective naturally and intrinsically yields Differential Privacy (DP) \emph{on its own and without the need for further randomization}. Thus, our goal is not to design a new mechanism for ensuring privacy; instead, we aim to understand the extent to which an existing algorithmic tool (KL-regularization) ensures privacy as an unintended byproduct.

We characterize the level of DP that is intrinsically obtained when sampling from a KL-regularized policy across three types of problems: multi-armed bandits, linear contextual bandits, and reinforcement learning from human feedback (RLHF), within the offline data collection setting. These problems have broad applications in online learning and advertising  \citep{li2010contextual}, healthcare \citep{villar2015multi}, and large language model alignment \citep{bai2022training}. We summarize our contributions as follows:
\begin{itemize} 
\item We show a parallel between KL-regularization and the celebrated \emph{Exponential Mechanism}. We provide $\epsilon_0$-pure DP guarantees for all three settings we consider under certain dataset coverage assumptions. Additionally, we provide $(\epsilon,\delta)$-approximate DP guarantees that are not provided by the Exponential Mechanism. Our results show that, within the practical range $\delta\ll 1$,we obtain an exponential $\epsilon$-$\delta$ trade-off: one can choose any $c\in(0,1)$ and obtain $\epsilon=c\epsilon_0$ with $\delta\sim O\left(\exp(-\sqrt{c})\right)$ in multi-armed bandits, and with $\delta\sim O\left(\exp(-c)\right)$ in linear bandits and RLHF. Notably, our bound achieves DP for multi-armed bandits with partial dataset coverage, outperforming the Exponential Mechanism’s requirement of full coverage.
\item We propose a novel approach to obtaining approximate DP guarantees that combines KL-regularization with the idea of \textit{pessimism} in offline reinforcement learning. We give a refined approximate DP analysis for the Exponential Mechanism through a separate consideration of high-probability and low-probability arms, leveraging pessimism to show a refined bound on the sensitivity of high-probability arms, as well as an exponentially small upper bound on the probability of pulling low-probability arms.
\item We conduct numerical simulations to illustrate the applicability of our bounds under both pure DP (under better coverage assumptions) and approximate DP (under weaker coverage assumptions). We also show that KL-regularization hence our privacy guarantee do not necessarily lead to empirical performance degradation.
\end{itemize}

\section{Related works}
In this section we provide a detailed discussion on related works. 
Our work inscribes itself in a line of work that aims to show that existing randomized algorithms commonly used in machine learning inherently satisfy Differential Privacy \citep{ou2024thompsonsamplingdifferentiallyprivate, blocki2012johnsonlindenstrausstransformpreservesdifferential}. This allows us, in certain cases, to use existing machine learning pipelines as is, without need for complex and costly modifications to provide strong, formal privacy protections.

\paragraph{Regularization in decision-making problems.}
Regularization has been widely used in decision-making problems. In bandits, \citet{fontaine2019regularized} studied the regularized contextual bandits problem with regularization that constrains the policy around some reference policy. \citet{geist2019theory} provided a general theoretical framework for reinforcement learning in a large class of regularizers. KL-regularization and its special case --- entropy regularization, are the most commonly used methods in empirical works of decision-making, and have motivated a wide range of algorithms \citep{schulman2017trustregionpolicyoptimization, schulman2017proximal, haarnoja2017reinforcement, haarnoja2018soft}, and have also been studied from a theoretical perspective on the convergence guarantees of different algorithms \citep{agarwal2020optimality, mei2020global, cen2022fast}. Reinforcement learning with human feedback (RLHF), which maximizes a KL-regularized reward model, serves as one of the most effective ways to learn from human feedback \citep{ouyang2022training, bai2022training}. From a theoretical perspective, \citet{xiong2023iterative} formulated the problem of RLHF as a KL-regularized linear bandit problem and gave theoretical guarantees for both offline and online settings. \citet{song2024importanceonlinedataunderstanding} studied the different dataset coverage assumptions required for online and offline settings, and \citet{zhao2024sharp} provided a sharper sample complexity analysis for KL-Regularized bandits and RLHF.
While the works above illustrated the effectiveness of KL/entropy regularization in these decision-making problems, we focus on the inherent privacy guarantee that KL/entropy regularization brings to these problems.

\paragraph{Differential privacy in decision-making problems.} 
Achieving DP in decision-making problems through explicit noise injection has been studied in various settings.
In bandits, \citet{shariff2018differentially} and \citet{han2021generalized} established joint/local differential privacy for (contextual) linear bandits, giving both privacy guarantee and sublinear regret bounds. \cite{azize2022privacy} characterized inherent privacy-regret trade-offs through minimax lower bounds. \citet{wang2024optimal} and \citet{syrgkanis2016efficientalgorithmsadversarialcontextual} proved regret bounds for bandits where the Follow-the-Perturbed-Leader approach achieves instance-dependent bounds under global DP through Gaussian and Laplace perturbations respectively. \citet{azize2024concentrated} used adaptive noise injection to showcase concentrated DP in linear contextual bandits.
In RL, \citet{vietri2020private} pioneered joint DP guarantees for episodic RL through private optimism-based learning, \citet{qiao2023offlinereinforcementlearningdifferential} studied DP under an offline setting, and \citet{rio2025differentially} extended to DP under policy gradient and trust-region methodologies to enable deep RL algorithms. In RLHF, while \citet{korkmaz2024learning} and \citet{chowdhury2024differentially} established privacy guarantees for reward learning with human feedback, the connection between ubiquitous RLHF regularization techniques and formal privacy guarantees remains uncharacterized.
Our work considers the privacy guarantee for the released action, which resembles the objective in the line of works referred to as ``prediction privacy" \citep{dwork2018privacy, dagan2020pac, van2020trade}.

\paragraph{Differential privacy through regularization and intrinsic randomness.}
Previous works have also studied DP from a regularization perspective, but still with explicit noise injection. \citet{ponnoprat2023dirichlet} showed that a KL divergence minimizer using exponential mechanisms satisfies Rényi DP, and \citet{watson2020stability} showed that regularization-induced stability loss minimization (explicit or implicit) reduces DP noise requirements. Obtaining privacy guarantees from intrinsic noise in the algorithm has also been considered, with \citet{ou2024thompsonsamplingdifferentiallyprivate} investigated the privacy guarantees of Thompson sampling through its intrinsic randomness, and \citet{blocki2012johnsonlindenstrausstransformpreservesdifferential} showed that the classical Johnson-Lindenstrauss transform preserves differential privacy. In comparison, our work combines the properties of regularization and intrinsic randomness in decision-making problems to show that the action sampled through a KL-regularized objective is inherently private.

\section{Preliminaries and Problem Formulation}\label{sec:prelims_and_settings}

In this section, we formally introduce the three problem settings that we consider in this paper. We also formally introduce the notion of differential privacy. Throughout this paper, we use $\langle x,y\rangle_\Sigma$ to denote the induced inner product $x^T\Sigma y$ by a positive-definite matrix $\Sigma$, and $\|x\|_\Sigma$ to denote the corresponding induced norm $\sqrt{x^T\Sigma x}$. When context is clear, we use $\langle\cdot,\cdot\rangle$ and $\|\cdot\|$ to denote the $\ell^2$-inner product and norm respectively. For a finite set $\sA$, we use $\Delta(\sA)$ to denote the probability simplex over $\sA$, and $|\sA|$ to denote the number of elements in $\sA$. We use $\mathbb{I}[\cdot]$ to denote the indicator function that equals $1$ if the expression $[\cdot]$ holds and $0$ otherwise.

\subsection{Multi-armed bandits and linear contextual bandits with KL-regularization}

\paragraph{Multi-armed bandits with KL-regularization.} We begin with a fundamental setting: the multi-armed bandit problem. A decision maker faces a set of arms (also referred to as actions) $\sA$ and aims to learn a policy $\pi\in \Delta(\sA)$, where $\pi(a)$ denotes the probability of selecting $a\in \sA$. When an arm $a$ is pulled, it will generate a random reward signal $R(a)$ satisfying $\E[R(a)]=\mu(a)$ for some $\mu:\sA\rightarrow [0,1]$. In the decision-making process, the decision maker is given an offline dataset $\sD=\{(a_i,r_i)\}_{i=1}^n$ with $n$ data points, where each data point is a tuple of the pulled arm $a_i$ and the observed reward signal $r_i\sim R(a_i)$ from pulling arm $a_i$. For an action $a\in\sA$, we denote $N(a;\sD)=\sum_{i=1}^n \mathbb{I}[a_i=a;\sD]$ as the number of data points associated with action $a$ in $\sD$ (the `coverage'). We assume rewards are uniformly bounded: $r_i \in [0,R]$ for all $i$, and typically $R\approx 1$ in practice. The goal is to learn a policy maximizing the expected reward $\E_{a\sim \pi}\left[\mu(a)\right]$ that the policy obtains. 

Suppose the decision maker has a reference policy $\pi_0\in \Delta(\sA)$ and would like the obtained policy to stay close to it, it is natural to consider the KL-regularized objective:
\begin{equation*}\begin{aligned}
    J(\pi)=&\E_{a\sim\pi}[\mu(a)]-\eta D_{KL}(\pi\|\pi_0)
    =\E_{a\sim \pi}\left[\mu(a)+\eta\log\frac{\pi_0(a)}{\pi(a)}\right],
\end{aligned}\end{equation*}
where $\eta$ is a hyperparameter that controls the level of regularization that larger $\eta$ indicates a stronger restriction of the policy around $\pi_0$\footnote{In practice, the value of $\eta$ is usually obtained through evaluating the generalization performance of the obtained policy, which has a typical value between $[0.01, 1]$.}. Notice that if we take $\pi_0$ to be the uniform distribution over $\sA$, KL-regularization becomes entropy regularization, which is widely used in decision-making problems such as bandits and RL. The optimal policy $\pi^*=\argmax_\pi J(\pi)$ can be shown to have the following form:
\begin{equation}\label{eq:regularized_closed_form_optimal_policy}
    \pi^*(a)\propto \pi_0(a)\exp\left(\frac{\mu(a)}{\eta}\right),
\end{equation}
such that the probability of choosing action $a$ is proportional to $\pi_0(a)$ (as a prior induced by the reference policy) and the exponentiated reward. Throughout the paper we assume the reference policy $\pi_0$ is $\sigma$-regular:
\begin{equation}
    \max_{a,a'\in\sA,x\in \mathcal{X}} \log \frac{\pi_0(a|x)}{\pi_0(a'|x)}\leq \sigma.
\end{equation}

\paragraph{Linear contextual bandits with KL-regularization.}
The problem of linear contextual bandits is a generalization of multi-armed bandits. The decision maker aims to obtain a policy $\pi:\mathcal{X}\rightarrow \Delta(\sA)$ such that, on observing a context $x$ in some context space $\mathcal{X}$ (which describes some ``situation" that the reward function may depend on), the policy (possibly randomly) pulls an arm trying to maximize the expected reward, which may depend on both the observed context $x$ and the pulled arm $a$. We assume that the reward model satisfies some linear structure, such that there exists a known feature map $\phi:\mathcal{X}\times \sA\rightarrow \R^d$ and an unknown parameter $\theta^*\in \R^d$, such that the observed reward signal when pulling arm $a\in \sA$ under context $x\in\mathcal{X}$ satisfies $\E[R(x,a)]=(\theta^*)^T\phi(x,a)$. Without loss of generality we take $\theta^*$ and $\phi$ to be bounded, i.e., $\max_{x,a} \|\phi(x,a)\|\leq 1$, $\|\theta^*\|\leq B$.
The dataset $\sD$ has the form $\sD=\{(x_i,a_i,r_i)\}_{i=1}^n$, where $r_i\sim R(x_i,a_i), r_i\in[0,R]$ is the observed reward signal when pulling arm $a_i$ under context $x_i$. In linear contextual bandits, the KL-regularized objective has the form:
\begin{equation*}
    J(\pi)=\E_{x\sim \rho}\E_{a\sim \pi} \left[(\theta^*)^T\phi(x,a)+\eta\log\frac{\pi_0(a|x)}{\pi(a|x)}\right],
\end{equation*}
where $\rho$ is some distribution of interest on $\mathcal{X}$ \citep{zhao2024sharp}.
Similar to the multi-armed bandit case, the optimal policy $\pi^*$ also has the form $\pi^*(a|x)\propto \pi_0(a|x)\exp\left((\theta^*)^T\phi(x,a)/\eta\right)$.
Given a dataset $\sD$ and a fixed $\lambda$, we define the sample covariance matrix as:
\begin{equation}\label{eq:linear_contextual_bandits_SigmaD}
    \Sigma_\sD=\lambda I+\sum_{i\in \sD} \phi(x_i,a_i)\phi(x_i,a_i)^T,
\end{equation}
which measures the coverage in different directions of the feature space. Here $\lambda I$ is a Ridge regression regularizer---commonly used in linear bandits/RL \citep{abbasi2011improved, kazerouni2017conservative}.

\subsection{RL from human feedback}

Reinforcement learning from human feedback (RLHF) \citep{ziegler2020finetuninglanguagemodelshuman, ouyang2022training} is a framework for aligning pretrained machine learning models, especially large language models (LLMs), with human preferences. RLHF enables models to produce responses that are both fluent and aligned with human values, instructions, and quality standards. RLHF starts from a model that has already been trained on a large dataset (often text, images, or speech) and adapts it to a specific downstream task. Human comparisons between multiple outputs of the pretrained model
are then collected to train a reward model that predicts how much a human would prefer a given response. The final model is trained under this reward model through reinforcement learning, usually with regularization to stay close to the pretrained model.

We consider a theoretical framework for RLHF, following the presentation of \citet{pmlr-v202-zhu23f} and \citet{xiong2023iterative}. Let $\mathcal{X}$ denote the space of prompts to the LLM, and $\sA$ denote the space of its responses (both being a subset of infinite-length token sequences), the pretrained LLM can be seen as a policy $\pi_0:\mathcal{X}\rightarrow \Delta(\sA)$, which takes a prompt $x\in\mathcal{X}$ and randomly generates a response $a\in \sA$. Assume there exists a ground-truth reward model $r^*:\mathcal{X}\times\sA\rightarrow \R^+$ that reflects the true underlying quality of the response, for a given prompt distribution $\rho$, in order to restrain the fine-tuned LLM to be close to the pretrained LLM, we consider optimizing KL-regularized objective: 
\begin{equation}\label{eq:RLHF_objective_function}
    J(\pi)=\E_{x\sim \rho}\E_{a\sim \pi} \left[r^*(x,a)+\eta\log\frac{\pi_0(a|x)}{\pi(a|x)}\right],
\end{equation}
as a standard technique in RLHF \citep{ziegler2020finetuninglanguagemodelshuman, NEURIPS2022_b1efde53, rafailov2024directpreferenceoptimizationlanguage}. We follow a widely used linear reward model assumption in the theoretical analyses of RLHF \citep{kong2022provably, pmlr-v202-zhu23f, xiong2023iterative}: namely, that the reward functions we consider are parameterized by $\theta\in \R^d$, with $r_\theta(x,a)=\theta^T\phi(x,a)$ for some known feature map $\phi:\mathcal{X}\times\sA\rightarrow \R^d$, and the ground truth reward $r^*$ is also parametrized, here as  $r^*(x,a)=(\theta^*)^T\phi(x,a)$. We also maintain the same boundedness assumptions stating that $\max_{x,a} \|\phi(x,a)\|\leq 1$ and $\|\theta^*\|\leq B$.

In RLHF, given a prompt $x$, two response samples $a^1,a^2\in\sA$ are generated; then, a human expert labels their preference $y=\mathbb{I}[a^1\succ a^2]$ (i.e. $a^1$ is preferred over $a^2$). We assume that, given the ground-truth reward function $r^*$, the preference generation process satisfies the Bradley-Terry model \citep{bradley1952rank}:
\begin{equation*}
    \Pr[y=1|x,a^1,a^2]=\mathsf{sigmoid}\big(e^{r^*(x,a^1)}-e^{r^*(x,a^2)}\big),
\end{equation*}
where $\mathsf{sigmoid}(x)=1/(1+\exp(-x))$. The goal of RLHF is to learn a policy that maximizes the objective function $J(\pi)$ in \eqref{eq:RLHF_objective_function} from a dataset $\sD=\{(x_i,a_i^1,a_i^2,y_i)\}_{i=1}^n$ consisting of $n$ prompt-response-preference tuples. When context is clear, in the RLHF setting, we overload the notation $\Sigma_\sD$ as:
\begin{equation*}\begin{aligned}
    \Sigma_\sD=\lambda I+\sum_{i\in \sD}\left(\phi(x_i,a_i^1)-\phi(x_i,a_i^2)\right)\left(\phi(x_i,a_i^1)-\phi(x_i,a_i^2)\right)^T.
\end{aligned}\end{equation*}

\subsection{Differential privacy}

Throughout this work, we adopt differential privacy (DP) as the primary quantitative measure of privacy. Here, we present respective privacy notions for the three problem settings above and a well-known DP mechanism closely related to KL-regularization.

\paragraph{Differential privacy for bandits.}

A general notion of DP is defined as:
\begin{definition}[Differential Privacy \citep{dwork2006calibrating}]\label{def:differential_privacy}
    A randomized mechanism $\mathcal{M}$ provides $(\epsilon,\delta)$-differential privacy if, for all datasets \( \mathcal{D}_1 \) and \( \mathcal{D}_2 \) differing in at most one element, and all subsets of outputs $S \subseteq \text{Range}(\mathcal{M}) $, the following condition holds:
\begin{equation*}
    \Pr[\mathcal{M}(\mathcal{D}_1) \in S] \leq e^{\epsilon} \cdot \Pr[\mathcal{M}(\mathcal{D}_2) \in S]+\delta,
\end{equation*}
where $\epsilon,\delta \geq 0$ are privacy parameters. When $\delta=0$, we say it's pure DP and use $\epsilon$-differential privacy as an abbreviation, when $\delta>0$, we say it's approximate DP.
\end{definition}
The notion of DP considers a randomized algorithm (mechanism) $\mathcal{M}$ that takes a dataset $\sD$ as input and randomly outputs $a\in \text{Range}(\mathcal{M})$, suggesting that the output distribution does not change significantly when the input $\sD$ is perturbed by a single data point. In practice, it is preferable to have a privacy level with $\epsilon<1$ and an exponentially small $\delta$ compared to the dataset size (e.g., $\delta<10^{-6}$ for $n=1000$) in most cases. In \Cref{def:differential_privacy},  neighboring datasets ``differing in one element'' can be interpreted as: $\sD_2$ is obtained (i) by adding/removing one data point from $\sD_2$ (add–remove DP) \citep[Definition 2.4]{dwork2014algorithmic} or (ii) by substituting one $\sD_1$ data point with another (swap DP) \citep{dwork2006calibrating}. These notions are equivalent up to constant factors. For simplicity, we adopt the add–remove definition and refer to it simply as `DP'.

\begin{proposition}\label{prop:conversion_between_addremove_swap}
    If a mechanism $\mathcal{M}$ is $(\epsilon,\delta)$-add-remove DP, it is $(2\epsilon,(1+\exp(\epsilon))\delta)$-swap DP. Meanwhile, if $\mathcal{M}$ is $(\epsilon',\delta')$-swap DP, it is $(\epsilon,\exp(\epsilon)\delta)$-add-remove DP.
\end{proposition}

We consider algorithms taking the dataset $\sD$ as input and compute a policy for the given KL-regularized objective. Specifically, we release an action $a$ sampled from the resulting (potentially stochastic) policy as the output. This utilizes intrinsic randomness of the policy and naturally fits privacy-critical applications in bandits (e.g., A/B tests, clinical trials) and RLHF \citep{gershoff2025kanonymousabtesting, liu2024survey}, where each user typically samples from the policy once given the data and context. Although weaker than releasing the policy itself, standard composition theorems can be applied to our analysis to obtain a corresponding $O(\sqrt{T})$ approximate DP for $T > 1$ samples, as discussed in \Cref{sec:extensions}.

\paragraph{Differential privacy for RLHF.} The pretraining phase of an LLM is usually conducted on a massive-scale public dataset from the internet, which can be viewed as public. In contrast, the RLHF phase uses datasets consisting of feedback from human labelers, which we want to keep private. Recall that in RLHF, datasets have the form $\sD=\{(x_i,a_i^1,a_i^2,y_i)\}_{i=1}^n$, following the notion in \Cref{def:differential_privacy} would certainly yield a privacy guarantee strong enough to protect each label $y_i$ generated by human experts through protecting the entire data entry. However, it is usually the case that the prompts and sampled responses are public and do not require privacy protection. Therefore, one may prefer a weaker notion that only protects the labels $y_i$ which contains sensitive information of human labelers, but not the context and response samples $x_i,a_i^1,a_i^2$:
\begin{definition}[Label DP \citep{ghazi2021deep}]\label{def:label-DP}
    Consider datasets whose entries are in the form $(x,a^1,a^2,y)$ where $y\in\{0,1\}$ is the label,
    a randomized mechanism $\mathcal{M}$ is $(\epsilon,\delta)$-label differentially private if, for all datasets \( \mathcal{D}_1 \) and \( \mathcal{D}_2 \) differing only in the label $y$ of at most one element, and all subsets of outputs $S \subseteq \text{Range}(\mathcal{M}) $, the following condition holds:
\begin{equation*}
    \Pr[\mathcal{M}(\mathcal{D}_1) \in S] \leq e^{\epsilon} \cdot \Pr[\mathcal{M}(\mathcal{D}_2) \in S]+\delta.
\end{equation*}
When $\delta=0$, we say $\mathcal{M}$ is $\epsilon$-label DP.
\end{definition}

\paragraph{The Exponential Mechanism.}
We call two datasets $\sD_1$ and $\sD_2$ that differ by one element 
``neighbors'' as is standard in the DP literature. We write  $\|\sD_1-\sD_2\|_1=1$ to indicate that two datasets are neighbors.
Consider a utility-based decision making problem over a discrete action set 
$\sA$ that depends on an input dataset $\sD$. Let $u(\cdot~;\sD):\sA\rightarrow \R$ be a utility function computed from $\sD$. To capture how $u$ changes between neighboring datasets, we define the \textit{sensitivity} of $u$ as
\begin{equation}\label{eq:def_sensitivity}
    \Delta u:=\max_{a\in\sA}\max_{\|\sD_1-\sD_2\|_1=1} |u(a;\sD_1)-u(a;\sD_2)|,
\end{equation}
and the action-data-dependent sensitivity for any action $a\in \sA$ as:
\begin{equation}\label{eq:def_action_dependent_sensitivity}
    \Delta u(a;\sD):=\max_{\|\sD-\sD'\|_1=1} |u(a;\sD)-u(a;\sD')|.
\end{equation}
For such problems, the exponential mechanism has well-understood privacy and utility guarantees:
\begin{definition}[The Exponential Mechanism \citep{4389483}]\label{def:exponential_mechanism}
    Given a desired privacy level $\epsilon$, a dataset $\sD$, the sensitivity $\Delta u$ defined in \eqref{eq:def_sensitivity}, and a rule to compute the utility function $u(\cdot~;\sD)$, the exponential mechanism selects action $a$ with probability:
    \begin{equation}\label{eq:exponential_mechanism}
        \pi(a;\sD)\propto \exp\left(\epsilon u(a;\sD)/(2\Delta u)\right).
    \end{equation}
\end{definition}

The exponential mechanism is $\epsilon$-DP. Its 
utility guarantees can be found in ~\cite{dwork2014algorithmic}.

\section{Differential privacy through KL-regularization}\label{sec:DP_through_KL_regularization}
In this section, we provide privacy guarantees for KL-regularized multi-armed bandits, linear contextual bandits and RLHF.

The key idea underlying our analysis is that the closed-form KL-regularized optimal policy \eqref{eq:regularized_closed_form_optimal_policy} can be viewed as a variation of the exponential mechanism \eqref{eq:exponential_mechanism}. The key difference between them is the multiplicative factor of the reference policy $\pi_0$. This observation leads us to generalize of the privacy guarantees of the exponential mechanism in order to obtain guarantees that can be applied to KL-regularized policies as follows:
\begin{lemma}\label{lem:KL_regularization_privacy}
    Consider an offline decision-making problem over a finite set $\sA$ that takes a dataset $\sD$ as input and outputs an action $a\in \sA$.
    Suppose some utility function $r(a;\sD)$ of the dataset $\sD$ has bounded sensitivity $\Delta r$, then for any reference policy $\pi_0 \in \Delta(\sA)$, the policy $\widetilde{\pi}$ given by
    \begin{equation}\label{eq:lem_KL_regularization_privacy_1}
        \widetilde{\pi}(a;\sD)\propto \pi_0(a) \exp\left(r(a;\sD)/\eta\right)
    \end{equation}
    is $\frac{2\Delta r}{\eta}$-differentially private. Moreover, suppose $\sA$ can be divided into subsets $\mathcal{I}$ and $\bar{\mathcal{I}}=\sA\backslash \mathcal{I}$ such that:
    \begin{enumerate}
        \item All actions in $\mathcal{I}$ have bounded sensitivity $\forall a\in \mathcal{I}, \Delta r(a;\sD)\leq \Delta$;
        \item For all datasets $\sD'$ such that $\|\sD'-\sD\|_1\leq 1$, the probability that $a$ is selected by $\widetilde{\pi}(\cdot;\sD')$ is at most $\delta_0$, or equivalently, $\max_{\sD':\|\sD'-\sD\|_1\leq 1}\sum_{a'\in\sA} \mathbb{I}[a'\in \bar{\mathcal{I}}]\widetilde{\pi}(a;\sD')\leq \delta_0$.
    \end{enumerate}
    Then, the optimal policy $\pi$ is $\left(\frac{2\Delta}{\eta}-\log(1-\delta_0), \delta_0\right)$-differentially private.
\end{lemma}
A proof of \Cref{lem:KL_regularization_privacy} can be found in \Cref{app:proof_KL_regularization_privacy}. \Cref{lem:KL_regularization_privacy} generalizes the privacy guarantee of the exponential mechanism (cf.~\Cref{def:exponential_mechanism}) in two directions. First, it shows that even with an arbitrary reference policy $\pi_0$, privacy still holds---extending the guarantee of the exponential mechanism obtained when $\pi_0$ is the uniform policy in \eqref{eq:lem_KL_regularization_privacy_1}. Second, it allows large or even unbounded sensitivity for certain actions, provided that the probability with which $\widetilde{\pi}(\cdot;\sD)$ selects these actions is small; this is achieved by relaxing pure DP to approximate DP, thereby tolerating a small probability that the event that arms in the tail of $\tilde{\pi}$ are sampled.

As we will show later, \Cref{lem:KL_regularization_privacy} is crucial to obtaining a privacy guarantee under weaker coverage assumptions on $\sD$ across all three settings that we consider.

\subsection{Multi-armed bandits with KL-regularization}

We begin with multi-armed bandits. In offline bandits and RL, \textit{pessimism} is a commonly used principle for obtaining near-optimal performance guarantees \citep{rashidinejad2021bridging, jin2021pessimism, li2022pessimism, shi2022pessimistic}. Pessimism is introduced by subtracting a penalty term (that depends on coverage) from an arm’s empirical reward to obtain a high-probability lower confidence bound on its unknown expected reward. In this work, we subtract a pessimism penalty term
 $\Gamma(a;\sD)  \asymp  1/\sqrt{N(a;\sD)}$ from the empirical reward, yielding an optimal policy of the following form:
\begin{equation}\label{eq:multi_armed_bandits_policy_closed_form}
    \widetilde{\pi}(a;\sD)\propto \pi_0(a)\exp\left((\bar{r}(a;\sD)-\beta_0\Gamma(a;\sD))/\eta\right),
\end{equation}
where $\beta_0>0$ represents the level of the pessimism term, and $\bar{r}(a;\sD)=\frac{\sum_{i=1}^n \mathbb{I}[a_i=a]r_i}{N(a;\sD)}$ is the empirical average reward. Given this, we now introduce our privacy guarantee for $\widetilde{\pi}(a;\sD)$:
\begin{theorem}\label{thm:privacy_multi_armed_bandits}
    The mechanism $\mathcal{M}$ that takes $\sD$ as input, computes $\widetilde{\pi}(\cdot~;\sD)$ through \eqref{eq:multi_armed_bandits_policy_closed_form},
    then samples and outputs action $a$ from $\widetilde{\pi}(\cdot~;\sD)$ is $\epsilon_0$-differentially private where
    \begin{equation}\label{eq:thm_privacy_multi_armed_bandits_pureDP}
        \epsilon_0=\frac{1}{\eta}\Big(\frac{4R}{\min_{a}N(a;\sD)-1}+\frac{\beta_0}{(\min_{a}N(a;\sD)-1)^{3/2}}\Big)
    \end{equation}
    given $\min_{a\in \sA}N(a;\sD)>1$. Furthermore, it is $(\epsilon,\delta)$-approximate differentially private with
    \begin{equation*}\begin{aligned}
        &\epsilon=\frac{1}{\eta}\Big(\frac{4R}{N_0}+\frac{\beta_0}{N_0^{3/2}}\Big)-\log(1-\delta);\\
        &\delta=|\sA|\exp\Big(\min\{C_1,C_2\}-\frac{\beta_0}{\eta\sqrt{N_0+1}} \Big),
    \end{aligned}\end{equation*}
    where
    \begin{equation*}\begin{aligned}
        C_1&=\frac{\beta_0}{\eta}\frac{1}{\sqrt{N(\bar{a};\sD)-1}},\\
        C_2&=\frac{R}{\eta}+\sigma + \frac{\beta_0}{\eta}\frac{1}{\max_a\sqrt{N(a;\sD)-1}},
    \end{aligned}\end{equation*}
    for arbitrarily chosen $N_0> 0$, where $\sigma$ is the regularity of $\pi_0$ and $\bar{a} =\argmax_{a'}\{\bar{r}(a;\sD)+\eta\log \pi_0(a)\}$.
\end{theorem}

\paragraph{Proof outline.} Our proof of \Cref{thm:privacy_multi_armed_bandits} follows three technically involved steps. The first step addresses the difficulty of controlling the probability of rarely chosen arms under $\widetilde{\pi}$ that serves as $\bar{\mathcal{I}}$ in \Cref{lem:KL_regularization_privacy} by leveraging the pessimism penalty term, that yields two distinct upper bounds (corresponding to $C_1$ and $C_2$ under different coverage assumptions). The second step deals with the challenge to upper bound the sensitivities of rewards $\Delta r(a;\sD)$ and pessimism terms $\Delta \Gamma(a;\sD)$ of each arm. The last step uses \Cref{lem:KL_regularization_privacy} to combine the results of the first two steps and obtain the final result. The detailed proof is provided in \Cref{app:proof_privacy_multi_armed_bandits}.

\begin{figure}[htb]
    \centering
    \includegraphics[width=0.8\linewidth]{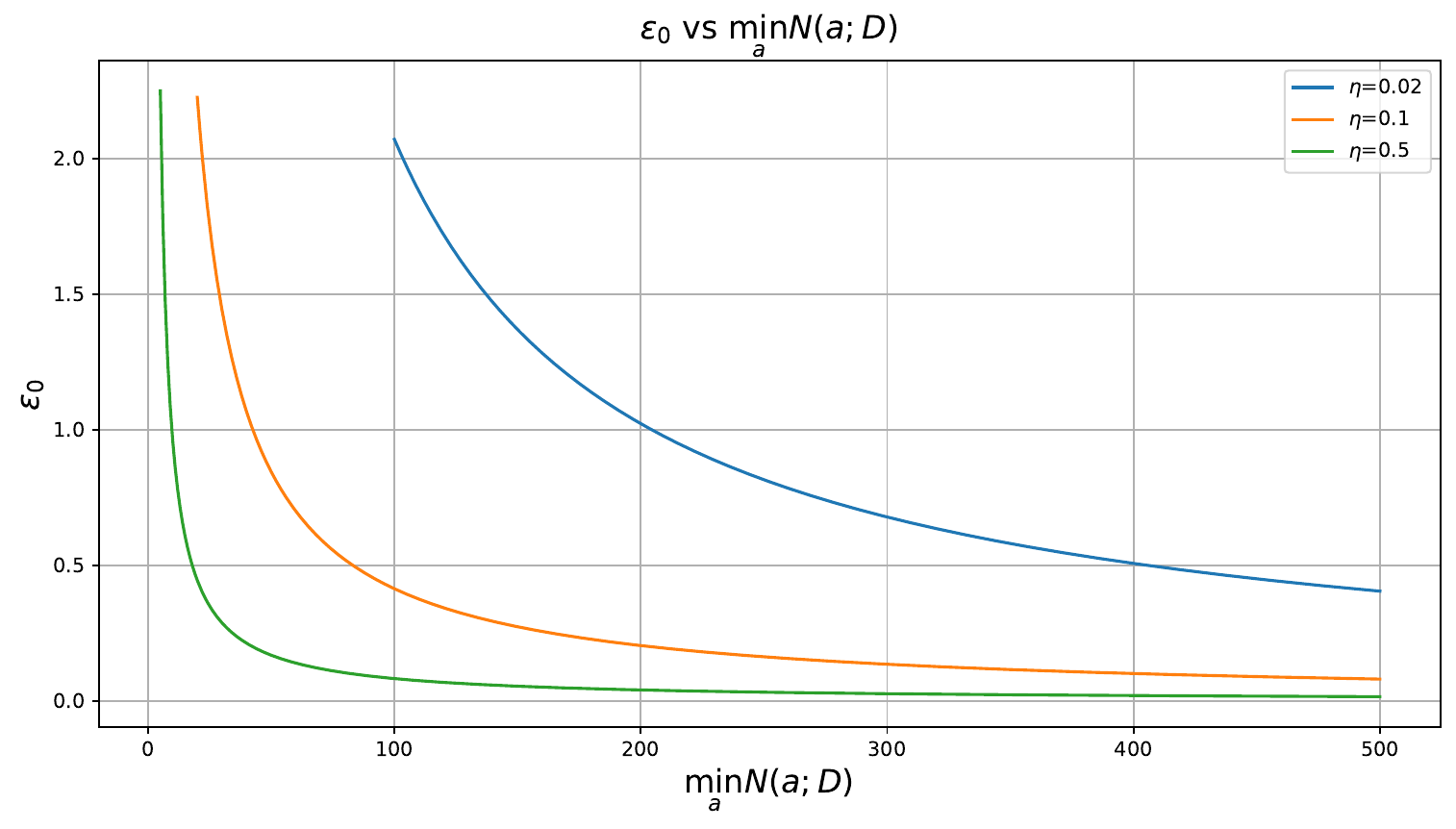}
    \caption{$\epsilon_0$-$\min_a N(a;\sD)$ curves for different $\eta$.}
    \label{fig:MAB_illustration}
\end{figure}

\paragraph{Pure DP with full data coverage.}
\Cref{thm:privacy_multi_armed_bandits} shows that the action sampled from the optimal policy $\widetilde{\pi}(\cdot;\sD)$, derived from the KL-regularized pessimistic objective, satisfies $\epsilon_0$-pure DP guarantees (cf. \eqref{eq:thm_privacy_multi_armed_bandits_pureDP}) with a privacy level on the order of $O\left(\frac{1}{\eta \min_a N(a;\sD)}\right)$ when the dataset covers all arms.
We illustrate the level $\epsilon_0$ of DP with respect to the minimal coverage $\min_a N(a;\sD)$, with $R=1$ and $\beta_0=1$, for different choices of $\eta$ in \Cref{fig:MAB_illustration}. As the regularization strength $\eta$ increases, the privacy guarantee $\epsilon_0$ improves (becomes smaller). This reflects the intuition that a larger $\eta$ keeps the policy closer to the reference policy $\pi_0$ and less sensitive to the private information in $\sD$.

\paragraph{Approximate DP with partial data coverage.}
\Cref{thm:privacy_multi_armed_bandits} ensures that, even in the absence of coverage across all arms (i.e., without requiring $\min_a N(a;\sD) > 0$), $\widetilde{\pi}(\cdot;\sD)$ retains $(\epsilon,\delta)$-approximate-DP guarantees, since the terms $C_1, C_2$ depend only on a specific action $\bar{a}$ and on $\max_a\sqrt{N(a;\sD)}$, respectively. Notably, the offline dataset $\sD$ typically provides sufficient coverage of high-return regions to ensure the feasibility of learning an optimal policy (i.e., the problem is well-posed) \cite{woo2025blessing,shi2022pessimistic}, which often implies a sufficiently large  $N(\bar{a};\sD)$. To achieve a reasonable privacy level with $\delta \ll 1$ at a practical level of $\epsilon$ without degrading the performance of the output $\widetilde{\pi}(a;\sD)$, we can design $N_0$ and demonstrate the reasonable choice of algorithm parameters including the penalty level $\beta_0$ and the regularization level $\eta$ as below:

\begin{corollary}\label{cor:MAB_specification}
With normalized $R=1$ and offline dataset sufficiently covering the high-reward area, satisfying $N(\bar{a};\sD) \geq 5$. With some fixed constant penalty term $\beta_0 > 1$, let $N_0 = \frac{N(\bar{a};\sD) -3}{2} \geq 1$, we can choose $\eta \leq \frac{\beta_0}{4\sqrt{N(\bar{a};\sD)-3} \log\frac{|\sA|}{\delta}}$ to achieve arbitrarily small $\delta$ and sufficient small $\epsilon$ on the order of at most a logarithmic term or constant $O\big( (\frac{1}{\sqrt{N(\bar{a};\sD)} \beta_0}+\frac{1}{N(\bar{a};\sD)}\big) \log\frac{|\sA|}{\delta}\big)$.
\end{corollary}

\begin{figure}[htb]
    \centering
    \includegraphics[width=0.8\linewidth]{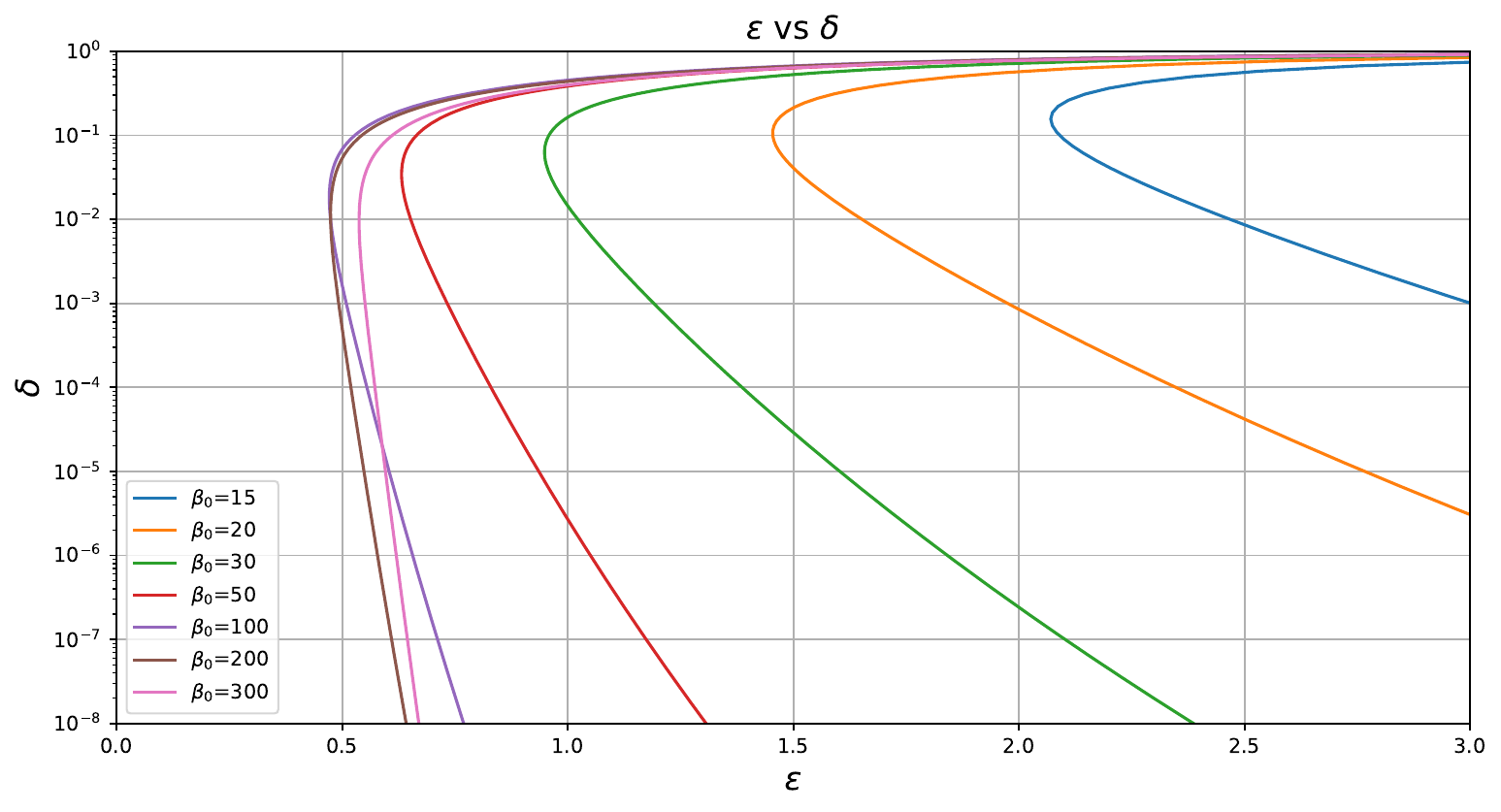}
    \caption{$\epsilon$-$\delta$ curves for different $\beta_0$.}
    \label{fig:MAB_illustration2}
\end{figure}

We provide an illustration the approximate DP result in \Cref{thm:privacy_multi_armed_bandits} (we use $C_2$ when computing $\delta$ that requires weaker coverage assumption) by setting $\eta=0.1$, $R=1$, $N_{\max}(\sD)=\max_a N(a;\sD)=500, \sigma=5$ and $|\sA|=5$ and obtaining the corresponding $\epsilon$ and $\delta$ curves by varying $N_0$ for different values of $\beta_0$ as plotted in \Cref{fig:MAB_illustration2}. We can see that for certain values of $\beta_0$, we are able to control $(\epsilon,\delta)$ in a practical range. For example, if we take $\beta_0=30$, we obtain a $\delta < 10^{-6}$ for a target privacy level $\epsilon=2.0$.

\begin{figure}[htb]
    \centering
    \includegraphics[width=0.8\linewidth]{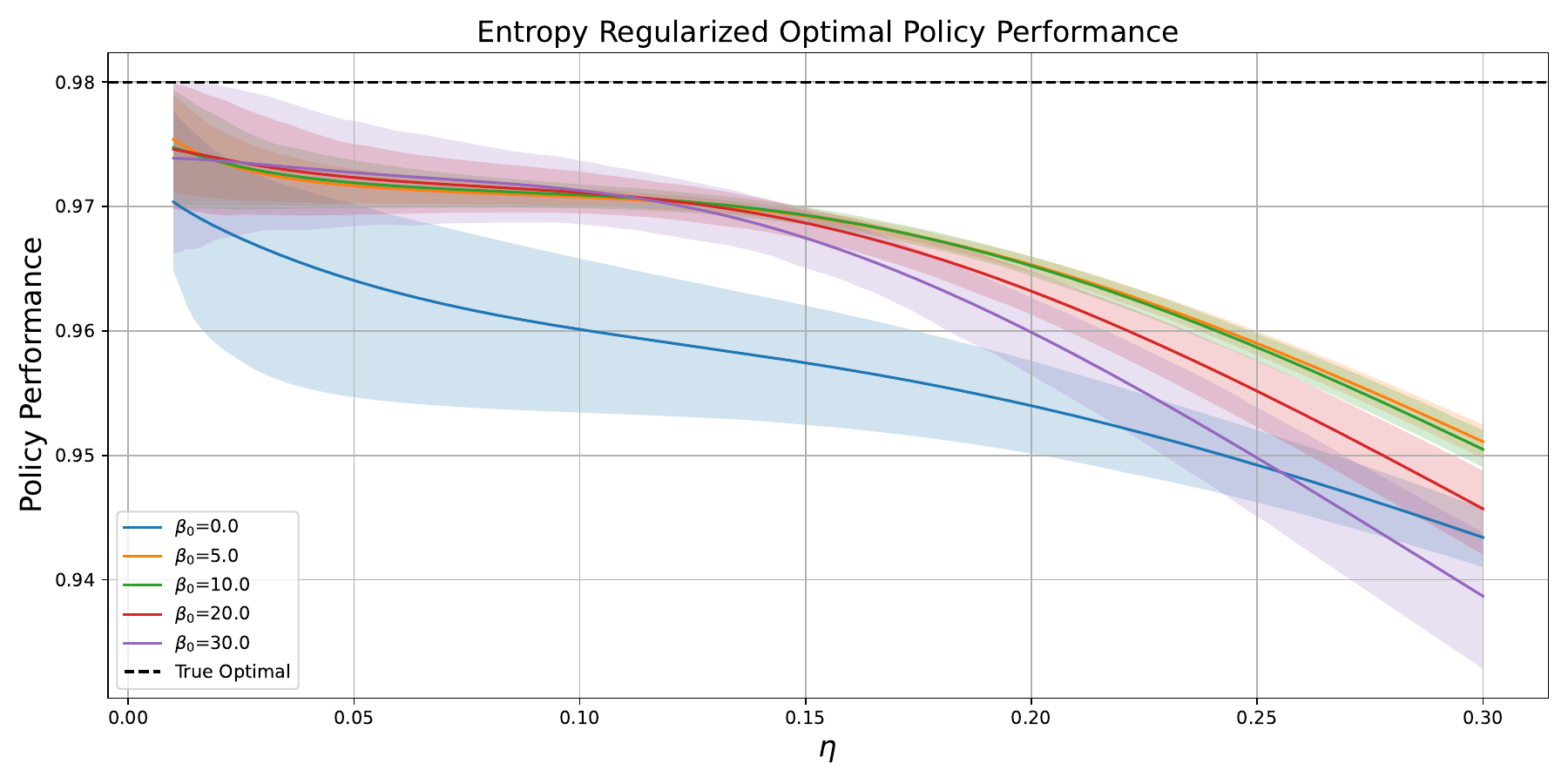}
    \caption{Performance-$\eta$ curves for different $\beta_0$.}
    \label{fig:KL_regularized_policy_performance}
\end{figure}

\paragraph{Balance between performance and DP.}
It is important to notice that the introduction of $\eta$ and $\beta_0$ are both commonly used approaches to enhance performance in bandits, and while larger $\beta_0$ and $\eta$ could lead to better approximate DP guarantee, this does not necessarily lead to huge performance degradation on the resulting policy. We illustrate through the following example: Consider a multi-armed bandit problem with $|\sA|=5$ arms. The reward of each arm follows a Bernoulli distribution with means $\mathbf{\mu}=[0.97, 0.9, 0.98, 0.1, 0.96]$. We construct 50 datasets, each consisting of $n=1500$ sampled rewards from the same behavioral policy $\pi_b=[0.23, 0.02, 0.23, 0.30, 0.22]$. We plot the entropy-regularized policy performance with the mean and 10th-90th quantile of these datasets for different values of $\beta_0$ in \Cref{fig:KL_regularized_policy_performance}. We can see that when $\eta < 0.1$ the slope of mean performance degradation is relatively small, and the 10th quantile performance can even be better as $\eta$ grows. Similarly, for different $\beta_0$, the performance within $\eta\in[0,0.1]$ are similar except for $\beta_0=0$ having worse performance. This indicates that our framework provides privacy guarantees without performance-privacy tradeoff, which is in sharp contrast to most existing privacy mechanisms.

\subsection{Linear contextual bandits with KL-regularization}
In this section we provide our results under the setting of linear contextual bandits. We first generalize \Cref{thm:privacy_multi_armed_bandits} to linear contextual bandits. Given a dataset $\sD$, the confidence penalty on a query $(x,a)$ usually takes the form of an elliptical potential $\Gamma(x,a;\sD)=\|\phi(x,a)\|_{\Sigma_\sD^{-1}}$, which quantifies the uncertainty (tailored to fit the concentration inequalities) in the reward estimate for the feature vector $\phi(x,a)$ \citep{abbasi2011improved, jin2021pessimism, xiong2023iterative}. Define $b_\sD=\sum_{i\in\sD}r_i\phi(x_i,a_i)$, the optimal policy has a similar form:
\begin{equation}\label{eq:linear_bandits_policy_closed_form}
    \widetilde{\pi}(a|x;\sD)\propto \pi_0(a|x)\exp\left(\frac{\bar{r}(x,a;\sD)-\beta_0\Gamma(x,a;\sD)}{\eta}\right)
\end{equation}
where $\bar{r}(x,a;\sD)=b_\sD^T\Sigma_\sD^{-1}\phi(x,a)$ is the least-square estimation of reward.
Let $\lambda_{\max}(\sD)$ and $\lambda_{\min}(\sD)$ be the maximum and minimum eigenvalues of $\Sigma_\sD$, we have:
\begin{theorem}\label{thm:linear_contextual_bandits_privacy}
    Fix state $x$, the action sampled from $\widetilde{\pi}(\cdot|x;\sD)$ is $\epsilon_0$-differentially private where
    \begin{equation*}
        \epsilon_0=\frac{1}{\eta\sqrt{\lambda_{\min}(\sD)}}\left(2\left(1+\sqrt{\frac{(n-1)d}{\lambda_{\min}(\sD)-1}}\right)\frac{R}{\sqrt{\lambda_{\min}(\sD)-1}}+\frac{\beta_0}{\lambda_{\min}(\sD)-1}\right).
    \end{equation*}
    Moreover, for arbitrarily chosen $\lambda_0> 0$, it is $(\epsilon,\delta)$-DP with
    \begin{equation*}
        \begin{aligned}
            \delta=&|\sA|\exp\left(\min\{C_5,C_6\}-\frac{\beta_0}{\eta}\frac{1}{\sqrt{\lambda_0(1+1/\lambda_{\min}(\sD))}} \right);\\
            \epsilon=&\frac{1}{\eta\sqrt{\lambda_0}}\left(\left(1+\sqrt{\frac{(n-1)d}{\lambda_{\min}(\sD)-1}}\right)\frac{2R}{\sqrt{\lambda_{\min}(\sD)-1}}+\frac{\beta_0}{\lambda_{\min}(\sD)-1}\right)-\log(1-\delta).
        \end{aligned}
    \end{equation*}
    where
    \begin{align*}
        C_5=&\frac{\beta_0}{\eta}\frac{\|\phi(x,\bar{a})\|_{\Sigma_\sD^{-1}}}{\sqrt{1-1/\lambda_{\min}(\sD)}},\bar{a}=\argmax_{a'} \{b_{\sD}^T\Sigma_\sD^{-1}\phi(x,a')+\eta\log \pi_0(a'|x)\};\\
        C_6=&\sqrt{\frac{(n-1)d}{\lambda_{\min}(\sD)-1}}\frac{R}{\eta} +\sigma+\frac{\beta_0}{\eta}\frac{\min_a \|\phi(x,a)\|_{\Sigma_{\sD}^{-1}}}{\sqrt{1-1/\lambda_{\min}(\sD)}}.
    \end{align*}
\end{theorem}

Similar to \Cref{thm:privacy_multi_armed_bandits}, 
\Cref{thm:linear_contextual_bandits_privacy} provides pure and approximate DP guarantees to the linear contextual bandit setting and its corresponding optimal solution of the KL-regularized pessimistic objective. However, \Cref{thm:linear_contextual_bandits_privacy} requires the asymptotic growth rate of $\lambda_{\min}(\sD) = \Omega(\sqrt{n})$ even for approximate DP, in order to achieve small $\epsilon$ and near-zero $\delta$.
Recall that $\Sigma_\sD=\lambda I+\sum_{i\in \sD}\phi(x_i,a_i)\phi(x_i,a_i)^T$, where $\lambda I$ serves as a regularization term commonly used in ridge regression, and $\sum_{i\in \sD}\phi(x_i,a_i)\phi(x_i,a_i)^T$ captures the empirical data coverage. When the dataset has sufficient coverage in all directions such that $\sum_{i\in \sD}\phi(x_i,a_i)\phi(x_i,a_i)^T$ covers all directions in the feature space at a rate of at least $\Omega(\sqrt{n})$, we know that we can get arbitrarily small $\epsilon_0$ and $(\epsilon,\delta)$ pair as $n\rightarrow\infty$. Otherwise, when our dataset size $n$ is fixed, we may also select a larger regularization strength $\lambda=\Omega(\sqrt{n})$ as a strong prior and obtain the desirable privacy levels.

To further clarify the dependence of privacy on different parameters, we introduce the following lemma by taking $\lambda_{\min}(\sD)=\Theta(n)$ and $\lambda_0=\lambda_{\min}(\sD)/c^2$ for some $c\in(0,1)$:
\begin{corollary}\label{cor:linear_contextual_bandits_privacy}
    Assume the coverage of dataset $\sD$ satisfies $\lambda_{\min}(\sD)\geq 1+\frac{n}{k_1d}$ for some constant $k_1>0$, the action sampled from $\widetilde{\pi}(\cdot|x;\sD)$ is $\epsilon_0$-differentially private where
    \begin{equation*}
        \epsilon_0=\frac{k_1d}{n\eta}\left(2R(1+\sqrt{k_1}d)+\beta_0\sqrt{k_1d/n}\right).
    \end{equation*}
    Moreover, for arbitrary constant $c\in (0,1)$, if $\|\phi(x,\bar{a})\|_{\Sigma_\sD^{-1}}\leq \sqrt{k_2d/n}$ for some constant $k_2>0$, it is $(\epsilon,\delta)$-DP where
    \begin{equation}\label{eq:cor_linear_bandits_approxDP}\begin{aligned}
        \epsilon=c\epsilon_0-\log(1-\delta);\quad
        \delta=|\sA|\exp\left(\frac{\beta_0}{\eta}\sqrt{\frac{d}{n}}\left(-c\sqrt{k_1}+\sqrt{k_2\Big(1+\frac{k_1d}{n}\Big)}\right)\right).
    \end{aligned}\end{equation}
\end{corollary}

\Cref{cor:linear_contextual_bandits_privacy} considers the case where $\lambda_{\min}(\sD) \geq 1 + \frac{n}{k_1d}$, indicating that each direction in the feature space $\R^d$ is covered by the dataset at an asymptotic rate of $\Theta(n)$ as $n$ grows. Here $k_1$ captures the uniformity of dataset coverage across different directions, with $k_1=1$ indicating uniform coverage and larger $k_1$ implies less uniform coverage. Similarly, $k_2$ captures the dataset coverage on the most likely action $\bar{a}$. The $\epsilon_0$-pure DP result indicates that the more uniform the feature space is covered and the more data points we have, the better privacy $\widetilde{\pi}(\cdot|x;\sD)$ can provide. We plot the $\epsilon_0$-$k_1$ curves for different $n$ fixing $d=2, R=1, \beta_0=1, k_2=1$ and $\eta=0.2$ as follows:
\begin{figure}[htb]
    \centering
    \includegraphics[width=0.8\linewidth]{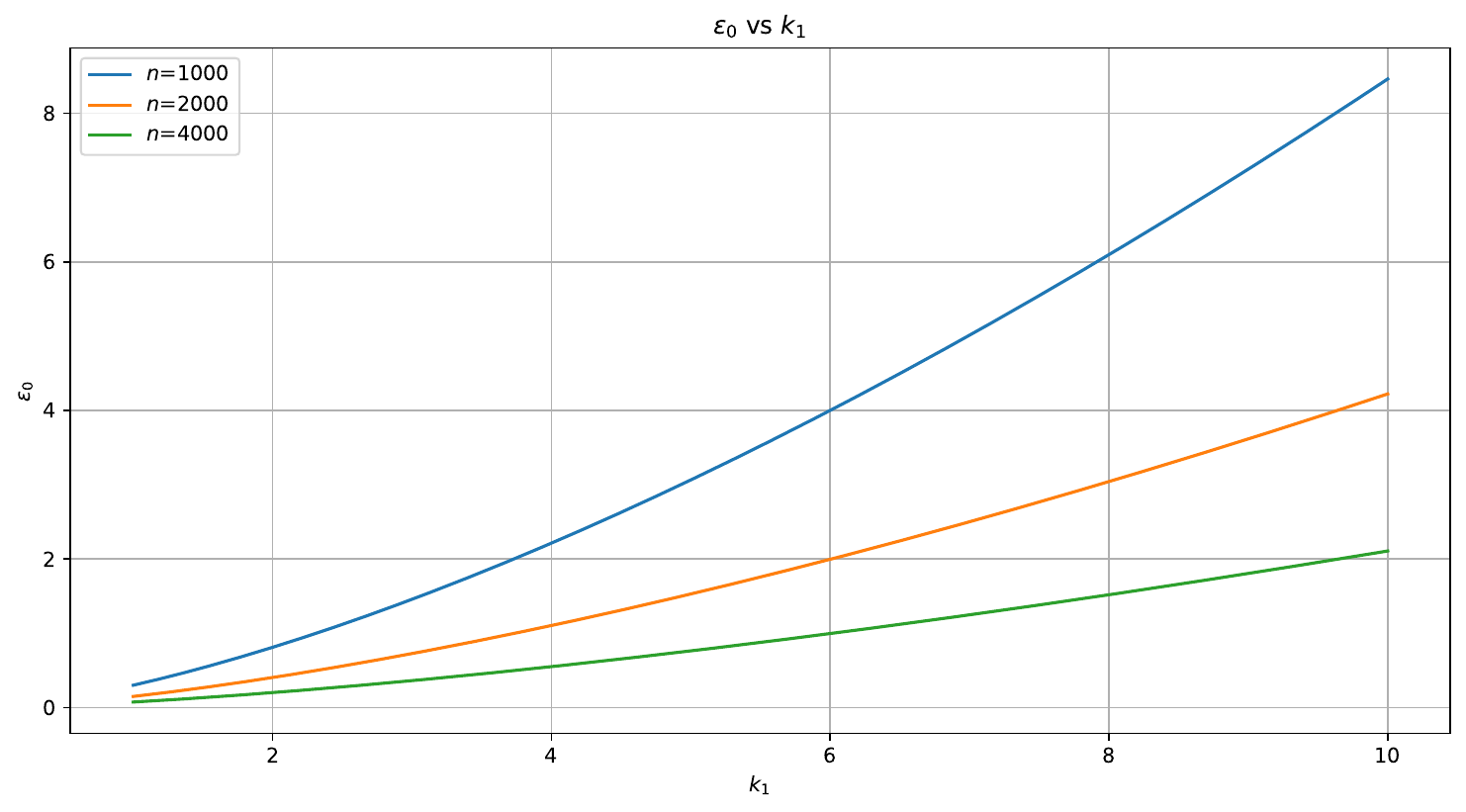}
    \caption{$\epsilon_0$-$k_1$ curves for $n\in\{1000,2000,4000\}$.}
    \label{fig:linear_bandits_illustration1}
\end{figure}

\begin{figure}[htb]
    \centering
    \includegraphics[width=0.8\linewidth]{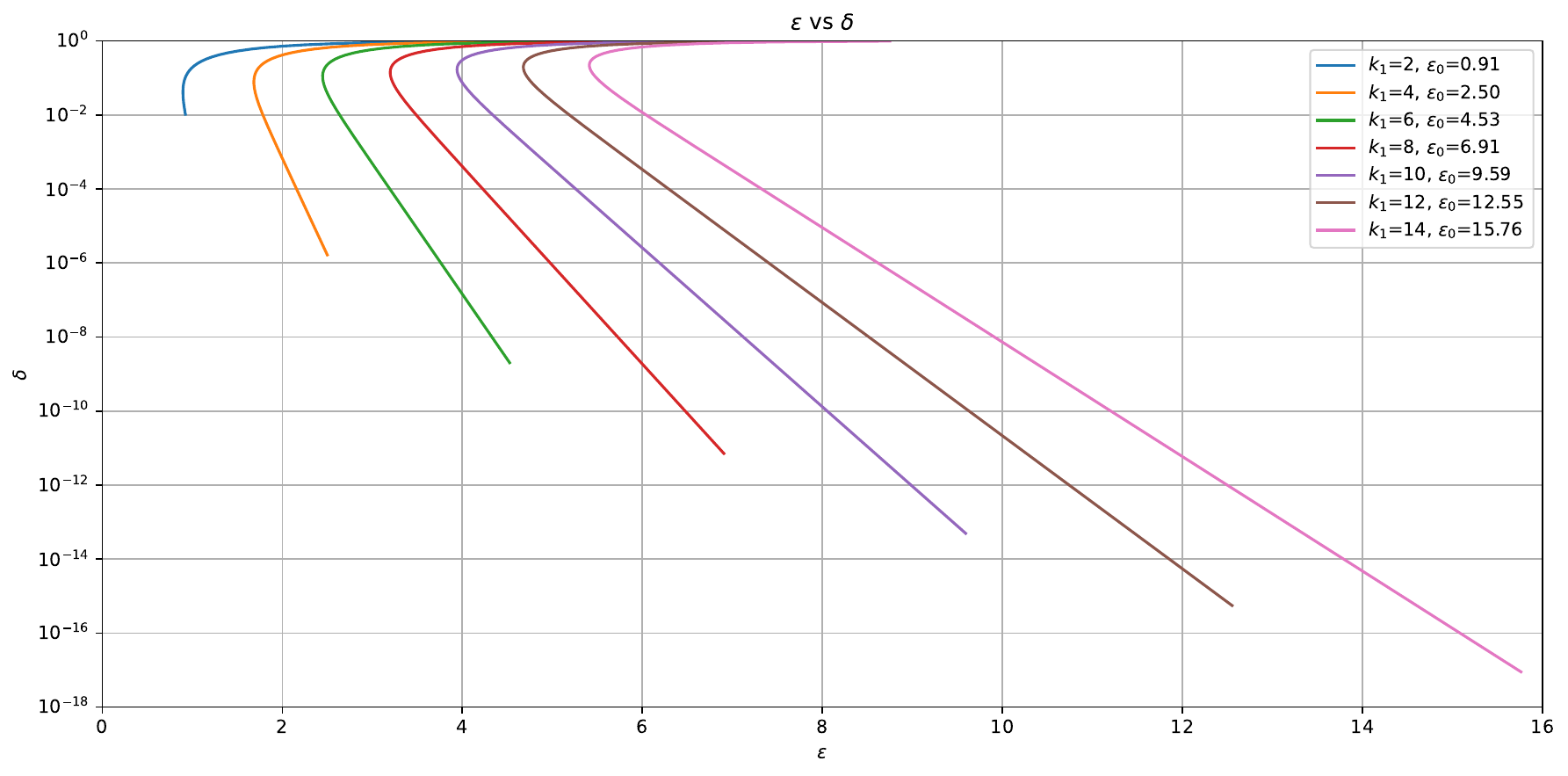}
    \caption{$\epsilon$-$\delta$ curves for $k_1\in\{2, 4, 6, 8, 10, 12, 14\}$.}
    \label{fig:linear_bandits_illustration2}
\end{figure}

In addition to the pure-DP result, \Cref{cor:linear_contextual_bandits_privacy} also gives the trade-off between $\epsilon$ and $\delta$ for obtaining an $\epsilon$ that is $c$ times stronger than $\epsilon_0$ at a rate of approximately $\delta\sim \exp(-c)$ when $\delta$ is small. Similar to that in multi-armed bandits, we plot the $\epsilon$-$\delta$ curves by varying $c$ in range $[0.01, 1]$ for different $k_1$ and obtain \Cref{fig:linear_bandits_illustration2}.
In \Cref{fig:linear_bandits_illustration2} we set $d=5, R=1, \beta_0=30, \eta=0.1, |\sA|=5, k_2=1$ and $n=2000$. We have also shown the pure DP value $\epsilon_0$ for each $k_1$. The plot shows that when $\delta$ is small, the log-scaled $\delta$ increases linearly as $\epsilon$ decreases, as indicated by \eqref{eq:cor_linear_bandits_approxDP}, $\delta\propto \exp(-\epsilon)$. The endpoint of each curve indicates the values of $\epsilon$ when $c=1$ (when pure DP is attained). For larger $k_1$ which means less uniform dataset coverage, the smallest $\delta$ we can get from our bound (corresponding to $c=1$ at which pure DP is more preferable) is exponentially smaller, and we can utilize the tolerance of $\delta$ to get a smaller $\epsilon_0$ while controlling $\delta$ within a practical range. Taking $k_1=14$ as an example, although the pure DP parameter $\epsilon_0=15.76$ is weak, if one can tolerate a failure probability of $\delta=10^{-5}$, the corresponding $\epsilon$ can be improved to approximately $8$, which is much stronger than $\epsilon_0$.

\begin{remark}
    In the illustration of both \Cref{fig:MAB_illustration2} and \Cref{fig:linear_bandits_illustration2}, we set the parameters $\eta$ to be small and $\beta_0$ to be large. This is because the ratio $\beta_0/\eta$ acts as a multiplier that controls the expontially decaying rate of $\delta$. Notice that although the expression of $\epsilon$ also includes multiplicative factors $1/\eta$ and $\beta_0/\eta$, their rates are not exponential and will not affect the final $\epsilon$ value as much as that to $\delta$. In practice, $\eta$ and $\beta_0$ are usually tuned (and fixed afterwards) towards better performance rather than privacy, one may prefer using the approximate DP guarantees when $\beta_0/\eta$ is large, and pure DP guarantees otherwise.
\end{remark}

\begin{figure}[h]
    \centering
    \includegraphics[width=0.8\linewidth]{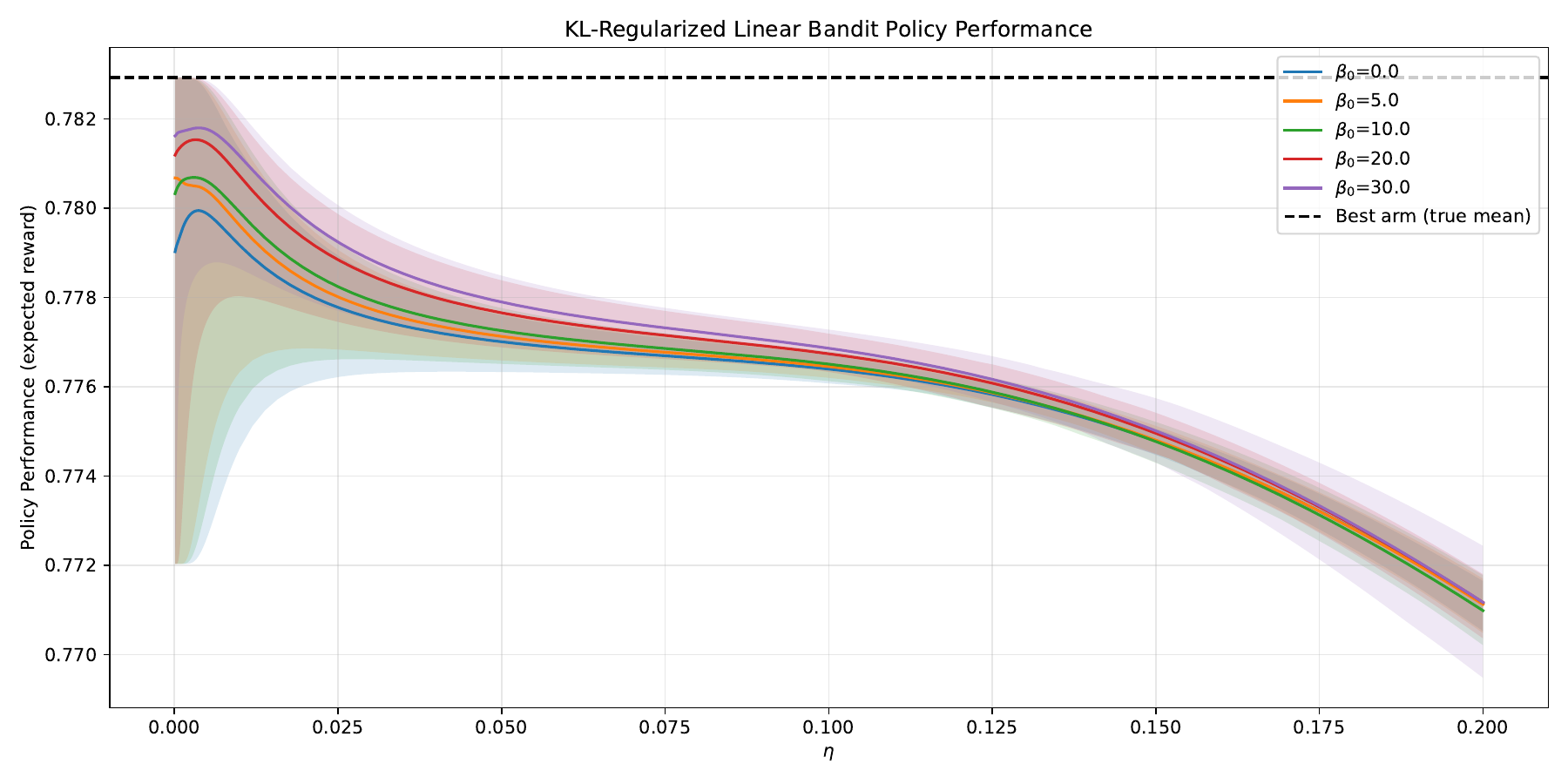}
    \caption{Performance-$\eta$ curves for $\beta_0\in\{0, 5, 10, 20, 30\}$.}
    \label{fig:KL_regularized_linear_bandit_policy_performance}
\end{figure}

Similar to that in \Cref{fig:KL_regularized_policy_performance}, to see that stronger regularization does not necessarily lead to worse empirical performance, we plot the performance-$\eta$ curves for different values of $\beta_0$ in \Cref{fig:KL_regularized_linear_bandit_policy_performance}.
In this example we set $|\sA|=5,d=3, \theta^*=[1.0,1.0, 0.1]$ and construct $50$ datasets each with $n=2000$ sampled rewards from $\pi_b=[0.17, 0.01, 0.16, 0.34, 0.32]$. For simplicity, we drop the context and set the feature of each arm as follows:
\begin{equation*}
    \begin{aligned}
        &\phi(a_1)=[0.2, 1.7, 0.0];
        \phi(a_2)=[0.1, 1.1, 0.0];
        \phi(a_3)=[0.2, 1.7, 0.0];\\
        &\phi(a_4)=[0.0, 0.0, 1.0];
        \phi(a_5)=[1.0, 0.1, 0.0].
    \end{aligned}
\end{equation*}
to maintain boundedness assumption, $\|\phi(a_i)\|_2$ and $\|\theta^*\|_2$ are normalized to be 1. The reward samples of arm $i$ follow Bernoulli distribution with mean $(\theta^*)\phi(a_i)$.

The proof of \Cref{thm:linear_contextual_bandits_privacy} can be found in \Cref{app:proof_linear_contextual_bandits_privacy}.
There are two reasons for the additional requirement of coverage assumption compared to the multi-armed bandit setting. The first one is that the least-square estimate $\bar{r}$ is no longer upper bounded by $R$ as in the bandit case.
The second one is the inner product $\langle \phi(x,a),\phi(x_i,a_i)\rangle_{\Sigma_{\sD}^{-1}}$ scaled by the covariance matrix which appears in the sensitivity bounds of both $\bar{r}(x,a;\sD)$ and $\Gamma(x,a;\sD)$. While this term is zero when $(x_i,a_i)\neq (x,a)$ in multi-armed bandits, this does not generally hold in the linear bandit setting.
Therefore, we have to use Cauchy-Schwarz to turn it into $\|\phi(x,a)\|_{\Sigma_\sD^{-1}}\|\phi(x_i,a_i)\|_{\Sigma_\sD^{-1}}$ and bound $\|\phi(x_i,a_i)\|_{\Sigma_\sD^{-1}}$ by $1/\sqrt{\lambda_{\min}(\sD)}$. To illustrate the intrinsic difference between multi-armed bandits and linear bandits, we provide the following example: Consider a linear bandit (for simplicity, again we drop the dependency on context) with $|\sA|=3, d=2$ and $\theta^*=(10.0, 1.0)$. The feature vector of each arm is as follows:
\begin{equation*}
    \begin{aligned}
        \phi(a_1)=[1.0, 0.0];
        \phi(a_2)=[0.1, 1.0];
        \phi(a_3)=[0.0, 1.0].
    \end{aligned}
\end{equation*}
To ensure boundedness, again$\|\phi(a_i)\|_2$ and $\|\theta^*\|_2$ are normalized to be 1. We consider the dataset $\sD$ with $n=200$ generated by sampling from the policy $\pi_b=[0.0,0.1,0.9]$ and its neighboring dataset $\sD_+$ by adding one data point $(a_1,r=1)$ to $\sD$. We plot the estimated $\theta$ using data from each dataset, and the resulting entropy-regularized optimal policies using different $\beta_0$ in \Cref{fig:linear_privacy_illustration}:
\begin{figure}[htbp]
    \centering
    \begin{subfigure}[b]{0.45\textwidth}
        \centering
        \includegraphics[width=\textwidth]{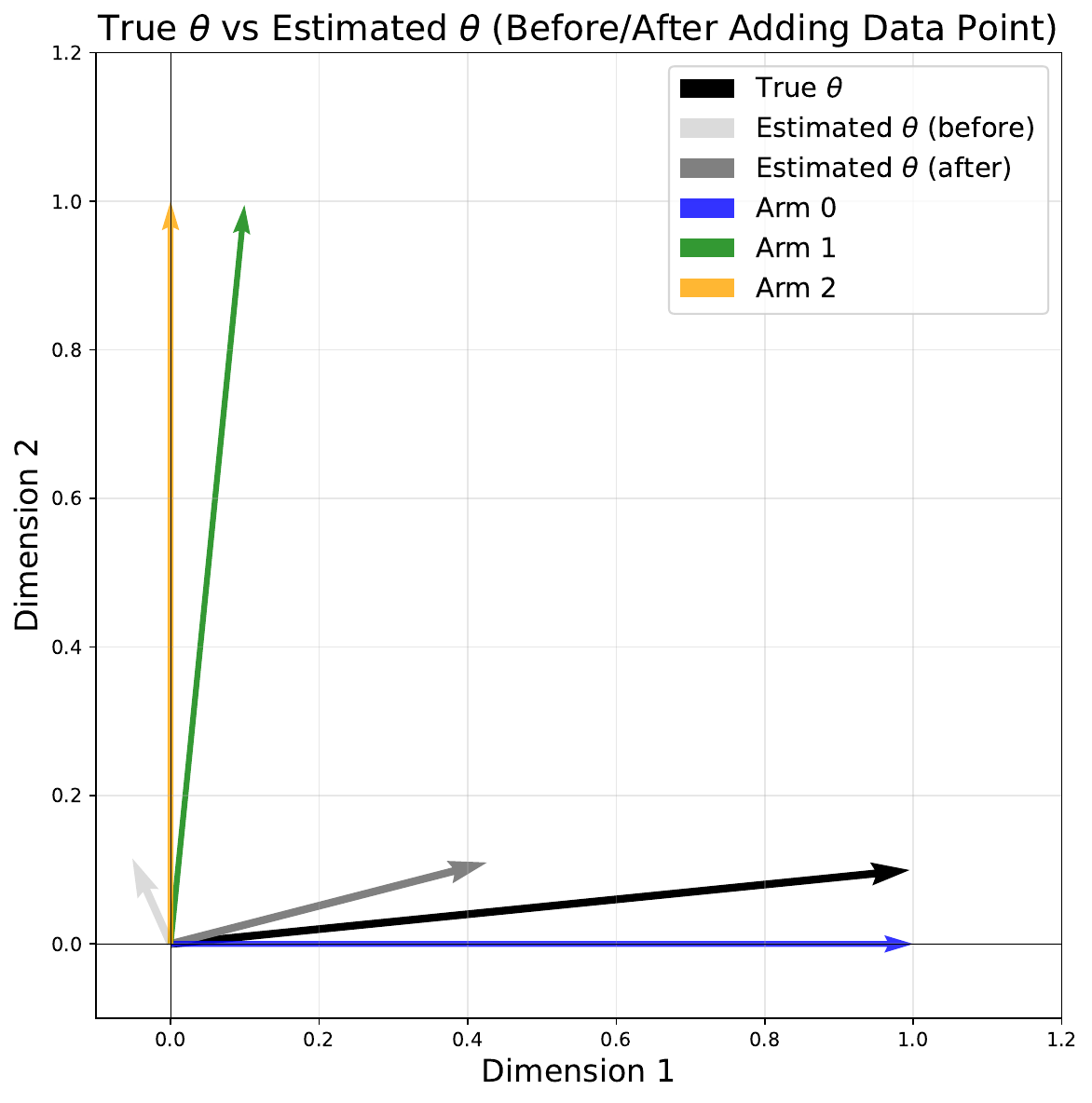}
        \caption{Arms, $\theta^*, \theta(\sD)$ (before) and $\theta(\sD_+)$ (after).}
        \label{fig:arms_and_theta}
    \end{subfigure}
    \hfill
    \begin{subfigure}[b]{0.5\textwidth}
        \centering
        \includegraphics[width=\textwidth]{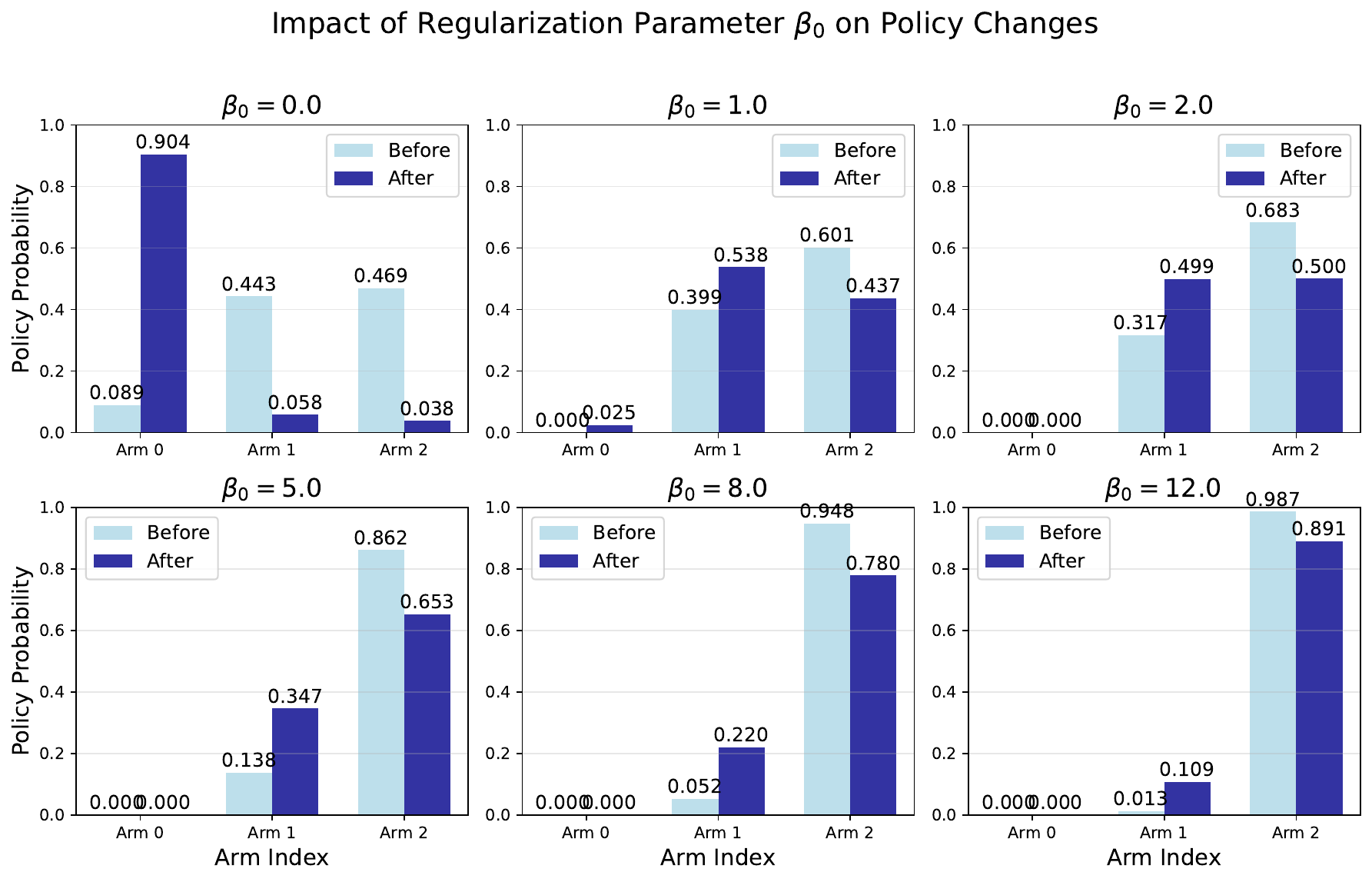}
        \caption{Policy comparison before and after adding 1 data point.}
        \label{fig:before_and_after}
    \end{subfigure}
    \caption{Illustration of privacy in linear bandits.}
    \label{fig:linear_privacy_illustration}
\end{figure}
In our constructed example, arms 1 and 2 are covered by the dataset $\sD$ but produces very weak reward signals, while arm 0 is well-aligned in the direction of $\theta^*$ but not covered by $\sD$. This causes our estimated $\theta(\sD)$ from $\sD$ being negative in dimension 1, which will never happen in multi-armed bandits because the rewards of each arm are orthogonal. After adding one data point of arm 0 into $\sD$, the resulting $\theta(\sD_+)$ differs a lot from $\theta(\sD)$ in dimension 1. If we compare the probabilities of each arm being sampled by the optimal policies $\widetilde{\pi}(\cdot;\sD)$ and $\widetilde{\pi}(\cdot;\sD_+)$ using $\beta_0=0$, we can see that the probability of pulling arm 0 changes drastically. When $\beta_0$ is increased, our pessimism term $\Gamma(\cdot;\sD)$ becomes larger. This leads to the policies before and after getting closer and closer for $\beta_0=1,2$, which indicates a stronger notion of privacy. However, when $\beta_0$ is too large, the sensitivity of $\Gamma(\cdot;\sD)$ becomes the dominating term in the privacy, which makes the policy difference larger again (and accordingly, weaker privacy).

\subsection{Privacy analysis for RLHF}

In this section, we provide privacy guarantees for RLHF using the quantitative metric label-DP (cf.~\Cref{def:label-DP}). We follow the algorithm proposed in \citep[Algorithm 1, Option II]{xiong2023iterative}, which adopts the maximum likelihood estimation $\theta_{MLE}(\sD)=\argmax_\theta \ell_\sD(\theta)$ for the log-likelihood under the Bradley-Terry model:
\begin{equation*}
    \begin{aligned}
        \ell_\mathcal{D}(\theta)=&\sum_{\mathcal{D}}\Big[y\log\left(\textsf{sigmoid}(r_\theta(x,a^1)-r_\theta(x,a^2))\right)
        +(1-y)\log\left(\textsf{sigmoid}(r_\theta(x,a^2)-r_\theta(x,a^1))\right)\Big].
    \end{aligned}
\end{equation*}
This leads to the following optimal policy:
\begin{equation}\label{eq:RLHF_policy_closed_form}
    \widetilde{\pi}(a|x;\sD)\propto \pi_0(a|x)\exp\Big(\frac{r_{MLE}(x,a;\sD)-\beta_0\Gamma(x,a;\sD)}{\eta}\Big),
\end{equation}
where $r_{MLE}(x,a;\sD)=\theta_{MLE}(\sD)^T\phi(x,a)$ is the maximum likelihood estimation of reward on $(s,a)$ and $\Gamma(x,a;\sD)=\|\phi(x,a)\|_{\Sigma_{\sD}^{-1}}$ is the elliptical potential that serves as the confidence penalty term  capturing the coverage on $(x,a)$. Let $\lambda_{\min}(\sD)$ be the minimum eigenvalue of $\Sigma_\sD$, we have the following privacy guarantee:
\begin{theorem}\label{thm:RLHF_privacy}
    Fix state $x$, the action sampled from $\widetilde{\pi}(\cdot|x;\sD)$ is $\epsilon_0$-label differentially private for
    \begin{equation*}
        \epsilon_0=\frac{1}{\eta}\Big(\frac{(2+2\exp(2B))}{\lambda_{\min}(\sD)-\lambda}\Big).
    \end{equation*}
    Moreover, for any $\lambda_0> \lambda$, it is $(\epsilon,\delta)$-label DP with
    \begin{equation*}\begin{aligned}
        \epsilon&=\frac{(2+2\exp(2B))\sqrt{\lambda_{\min}(\sD)}}{\eta\sqrt{\lambda_0}(\lambda_{\min}(\sD)-\lambda)}-\log(1-\delta);\\
        \delta&=|\sA|\exp\Big(\min\{C_3,C_4\}-\frac{\beta_0}{\eta}\frac{1}{\sqrt{\lambda_0}} \Big).
    \end{aligned}\end{equation*}
    where
    \begin{align*}
        &C_3=\frac{\beta_0}{\eta}\|\phi(x,\bar{a})\|_{\Sigma_\sD^{-1}};\\
        &C_4=\frac{2B}{\eta}+\sigma+\frac{\beta_0}{\eta}\min_a\|\phi(x,a)\|_{\Sigma_\sD^{-1}}.
    \end{align*}
    for $\bar{a}=\argmax_{a'}\{r_{MLE}(x,a';\sD)+\eta\log \pi_0(a'|x)\}$.
\end{theorem}
Similar to \Cref{thm:privacy_multi_armed_bandits}, \Cref{thm:RLHF_privacy} provides both pure and approximate label DP guarantees for the problem of RLHF. A more detailed version of \Cref{thm:RLHF_privacy} that considers both conventional DP and label DP as well as its proof can be found in \Cref{app:proof_RLHF_privacy}.

The pure DP result in \Cref{thm:RLHF_privacy} gives a privacy level of $O\left(\frac{1}{\eta(\lambda_{\min}(\sD)-\lambda)}\right)$. Here $\lambda_{\min}(\sD)-\lambda$ represents the minimum coverage in all directions of the parameterized feature space. As a special case, when the feature maps $\phi(x,a)$ for different arms are an orthonormal basis of the feature space, this recovers the standard multi-armed bandit problem and rate of $O\left(\frac{1}{\eta\min_a N(a;\sD)}\right)$ in \Cref{thm:privacy_multi_armed_bandits}, albeit with a different constant term indicating different upper bounds of the reward functions.

When some direction of the feature space is not sufficiently covered, $\lambda_{\min}(\sD)-\lambda$ is small, which leads to a large $\epsilon_0$. In this case, the approximate DP guarantees may be preferable. To better demonstrate the approximate DP guarantees, we provide the following corollary by choosing $\lambda_0=\frac{\lambda_{\min}(\sD)}{c^2}$ for some $c\in(0,1)$:
\begin{corollary}
    For arbitrary $c\in(0,1)$, we have $\epsilon=c\epsilon_0-\log(1-\delta); \delta=\min\{\delta_1,\delta_2\}$
    where
    \begin{align*}
        \delta_1=&|\sA|\exp\Big(\frac{\beta_0}{\eta}(\|\phi(x,\bar{a})\|_{\Sigma_\sD^{-1}}-\frac{c}{\sqrt{\lambda_{\min}(\sD)}})\Big);\\
        \delta_2=&|\sA| \exp\Big(\frac{2B}{\eta}+\sigma+\frac{\beta_0}{\eta}(\min_a\|\phi(x,a)\|_{\Sigma_\sD^{-1}}-\frac{c}{\sqrt{\lambda_{\min}(\sD)}})\Big).
    \end{align*}
\end{corollary}
Notice that $\epsilon$ does not depend on $\beta_0$, and for all $(x,a)$, $\|\phi(x,a)\|_{\Sigma_{\sD}^{-1}}\leq \|\phi(x,a)\|_2/\sqrt{\lambda_{\min}(\sD)}\leq 1/\sqrt{\lambda_{\min}(\sD)}$, with equality holds only if $\phi(x,a)$ lies in the least covered direction in the feature space. As long as there is one ``better" covered action, we have $\min_a\|\phi(x,a)\|_{\Sigma_\sD^{-1}}> 1/\sqrt{\lambda_{\min}(\sD)}$, we can find $c<1$ such that $\min_a\|\phi(x,a)\|_{\Sigma_\sD^{-1}}-c/\sqrt{\lambda_{\min}(\sD)}<0$, and then set $\beta_0$ to be arbitrarily large to obtain an arbitrarily small $\delta$. Therefore, within the feasible range of $c>\sqrt{\lambda_{\min}(\sD)}\min\{\|\phi(x,\bar{a})\|_{\Sigma_\sD^{-1}},\min_a\|\phi(x,a)\|_{\Sigma_\sD^{-1}}\}$, we can increase $\beta_0$ to get an arbitrarily small $\delta$ at an asymptotic rate of $O\Big(|\sA|\exp\Big(-\frac{\beta_0c}{\eta\sqrt{\lambda_{\min}(\sD)}}\Big)\Big)$ and a privacy level of $\epsilon\approx c\epsilon_0$ (as $\log(1-\delta)\approx0$ when $\delta\ll1$).

\paragraph{Proof outline.}
The proof follows the same outline as in \Cref{thm:privacy_multi_armed_bandits} that first constructs a set of low-probability arms $\bar{\mathcal{I}}=\{a\in\sA:\|\phi(x,a)\|_{\Sigma_{\sD'}^{-1}}\geq \frac{1}{\sqrt{\lambda_0}}\}$ for some dataset $\sD'$ with $\|\sD'-\sD\|\leq 1$ and bounds the probability $\delta$ that the sampled arm $a\in\bar{\mathcal{I}}$. This is made possible through the fact that the pessimism penalty $\Gamma(x,a;\sD)$ is large under this event, making $\widetilde{\pi}$ exponentially small. We provide two distinct upper bounds on the probability $\delta$, one under the condition that the empirically optimal arm $\bar{a}$ maximizing $r_{MLE}(x,\cdot;\sD)+\eta\log \pi_0(a'|x)$ (i.e. the arm with high estimated reward and likely to be selected by $\pi_0$ simultaneously) is well-covered by $\sD$, yielding the term $C_3$, and the other under the condition that $\pi_0$ is $\sigma$-regular, which gives the term $C_4$. After constructing the low-probability event, we bound the sensitivity of $r_{MLE}$ (or equivalently, $\theta_{MLE}$) for arms in $\mathcal{I}$. We follow a similar approach as in \citep{chaudhuri2008privacy}  utilizing the fact that $\nabla \ell_\sD(\theta_{MLE}(\sD))=0$ for all datasets $\sD$ and the Taylor expansion:
\begin{equation*}\begin{aligned}
    \langle \nabla \ell_{\sD_+}(\theta(\sD_+)), v\rangle=&\langle \nabla \ell_{\sD_+}(\theta(\sD)), v\rangle+(\theta(\sD_+)-\theta(\sD))^T \nabla^2 \ell_{\sD_+}(\theta') v
\end{aligned}\end{equation*}
for $\sD_+$ obtained by adding one element to $\sD$. Observing that $\ell_\sD$ is strongly convex for all $\sD$ and taking $v=\left(\Sigma_{\sD_+}-\lambda I\right)^{-1} \phi(x,a)$ leads to the final sensitivity bound of $r_{MLE}$. For the penalty term $\Gamma$, notice that in label-DP, the penalty terms are the same for neighboring datasets differing only in one label, the sensitivity of $\Gamma$ is therefore zero. The final step applies \Cref{lem:KL_regularization_privacy} to obtain both pure and approximate DP results.

\subsection{Extensions}\label{sec:extensions}

Several generalizations of the paper's results can be obtained through standard approaches. We discuss i) sampling more than a single action and ii) inexact policy optimization below.

\paragraph{Repeated sampling.}
In our analysis, we view the dataset $\sD$ as input and the sampled action $a\sim \widetilde{\pi}(\cdot;\sD)$ as output. This implies our privacy results only hold when sampling an action once. For multiple samples, one can directly apply existing composition theorems \cite[Theorem 3.20]{dwork2014algorithmic} and obtain:
\begin{proposition}\label{prop:multiple_sample_composition}
    If sampling once from $\widetilde{\pi}$ satisfy $\epsilon_0$ pure DP and $(\epsilon,\delta)$ approximate DP, then sampling $T$ actions from the same $\widetilde{\pi}$ satisfies $T\epsilon_0$-pure DP and $(\sqrt{2T\log(1/\delta')}\epsilon+T\epsilon(\exp(\epsilon)-1), k\delta+\delta')$-approximate DP for arbitrary $\delta'\geq 0$.
\end{proposition}
In comparison, existing privacy results for bandits and RLHF \citep{zheng2020locally, han2021generalized, wu2023privately, chowdhury2024differentially} treat the policy $\pi$ itself as the output, offering a stronger privacy model. There, once the policy is privately learned, one can sample an arbitrarily large number of actions from that policy while still being private. However, these algorithms require adding noise at some stage of the existing algorithms and therefore perform worse than the algorithm we present. This highlights a performance-privacy trade-off when compared to our approach, which keeps the original algorithm unchanged while offering a weaker notion of privacy.

\paragraph{Inexact policy optimization.}
We have assumed throughout that we have access to the exact optimal policy \eqref{eq:multi_armed_bandits_policy_closed_form}, \eqref{eq:linear_bandits_policy_closed_form} and \eqref{eq:RLHF_policy_closed_form}. However, when the action space is large (especially in the RLHF setting), the exact optimal policy is hard to compute because the normalization constant $Z=\sum_a \pi_0(a)\exp(r(a;\sD)/\eta)$ is computationally expensive. Alternatively, one can use existing policy optimization algorithms to obtain a policy that is close to optimal. 
Assume we obtain a policy $\widehat{\pi}(\cdot;\sD)$ close to the optimal policy $\widetilde{\pi}(\cdot;\sD)$ such that $D_{\infty}(\widetilde{\pi}(\cdot;\sD)\|\widehat{\pi}(\cdot;\sD))\leq \widehat{\epsilon}$ and $D_{\infty}(\widehat{\pi}(\cdot;\sD)\|\widetilde{\pi}(\cdot;\sD))\leq \widehat{\epsilon}$ where $D_{\infty}(\pi_1\|\pi_2)=\max_{a\in \sA}\log\frac{\pi_1(a)}{\pi_2(a)}$. \Cref{prop:inexact_policy_optimization} suggests that the resulting policy $\widehat{\pi}$ is $(\epsilon+2\widehat{\epsilon},\exp(\widehat{\epsilon})\delta)$-differentially private. Hence, if our policy optimization algorithm is accurate enough, the resulting policy $\widehat{\pi}(\cdot;\sD)$ remains private.

\begin{proposition}\label{prop:inexact_policy_optimization}
    For any policy $\widehat{\pi}$ such that $D_{\infty}(\widetilde{\pi}(\cdot;\sD)\|\widehat{\pi}(\cdot;\sD))\leq \widehat{\epsilon}$ and $D_{\infty}(\widehat{\pi}(\cdot;\sD)\|\widetilde{\pi}(\cdot;\sD))\leq \widehat{\epsilon}$ where $D_{\infty}(\pi_1\|\pi_2)=\max_{a\in \sA}\log\frac{\pi_1(a)}{\pi_2(a)}$, suppose sampling from $\widetilde{\pi}$ is $(\epsilon,\delta)$-differentially private, sampling from $\widehat{\pi}$ is $(\epsilon+2\widehat{\epsilon},\exp(\widehat{\epsilon})\delta)$-differentially private.
\end{proposition}
The proof of \Cref{prop:inexact_policy_optimization} is deferred to \Cref{app:proof_inexact_policy_optimization}.

\section{Conclusion}\label{sec:conclusion}
This paper studies the inherent privacy guarantees brought by regularization in multi-armed bandits, linear contextual bandits, and RLHF. Our work provides insights towards obtaining privacy guarantees through regularization and motivates potential generalizations in several directions. The settings that we consider in this paper are all offline, single-stage problems, and generalizing our results to online, sequential decision-making problems would be a natural direction for future work. The analysis that we carried out is based on existing algorithms with well-established performance guarantees, and it is interesting to see if the algorithms can be tuned (e.g., by changing the form of the penalty term) for privacy under weaker data coverage assumptions. In addition, the relationship between different privacy notions and different forms of regularization is an important direction for further study.

\section*{Acknowledgements}
The work is supported by the NSF through CNS-2146814, CPS-2136197, CNS-2106403, NGSDI-2105648, IIS-2336236, IIS-2504990, and by the Resnick Sustainability Institute at Caltech.
K. Panaganti acknowledges support from the Resnick Institute and the `PIMCO Postdoctoral Fellow in Data Science' fellowship at Caltech.
The work of L. Shi is supported in part by the Resnick Institute and Computing, Data, and Society Postdoctoral Fellowship at Caltech. The authors thank Heyang Zhao for valuable discussions.

\bibliographystyle{plainnat}   
\bibliography{references}

\begin{thebibliography}{70}
\providecommand{\natexlab}[1]{#1}
\providecommand{\url}[1]{\texttt{#1}}
\expandafter\ifx\csname urlstyle\endcsname\relax
  \providecommand{\doi}[1]{doi: #1}\else
  \providecommand{\doi}{doi: \begingroup \urlstyle{rm}\Url}\fi

\bibitem[Abbasi-Yadkori et~al.(2011)Abbasi-Yadkori, P{\'a}l, and Szepesv{\'a}ri]{abbasi2011improved}
Yasin Abbasi-Yadkori, D{\'a}vid P{\'a}l, and Csaba Szepesv{\'a}ri.
\newblock Improved algorithms for linear stochastic bandits.
\newblock \emph{Advances in neural information processing systems}, 24, 2011.

\bibitem[Agarwal et~al.(2020)Agarwal, Kakade, Lee, and Mahajan]{agarwal2020optimality}
Alekh Agarwal, Sham~M Kakade, Jason~D Lee, and Gaurav Mahajan.
\newblock Optimality and approximation with policy gradient methods in markov decision processes.
\newblock In \emph{Conference on Learning Theory}, pages 64--66. PMLR, 2020.

\bibitem[Azize and Basu(2022)]{azize2022privacy}
Achraf Azize and Debabrota Basu.
\newblock When privacy meets partial information: A refined analysis of differentially private bandits.
\newblock \emph{Advances in Neural Information Processing Systems}, 35:\penalty0 32199--32210, 2022.

\bibitem[Azize and Basu(2024)]{azize2024concentrated}
Achraf Azize and Debabrota Basu.
\newblock Concentrated differential privacy for bandits.
\newblock In \emph{2024 IEEE Conference on Secure and Trustworthy Machine Learning (SaTML)}, pages 78--109. IEEE, 2024.

\bibitem[Bai et~al.(2022)Bai, Jones, Ndousse, Askell, Chen, DasSarma, Drain, Fort, Ganguli, Henighan, et~al.]{bai2022training}
Yuntao Bai, Andy Jones, Kamal Ndousse, Amanda Askell, Anna Chen, Nova DasSarma, Dawn Drain, Stanislav Fort, Deep Ganguli, Tom Henighan, et~al.
\newblock Training a helpful and harmless assistant with reinforcement learning from human feedback.
\newblock \emph{arXiv preprint arXiv:2204.05862}, 2022.

\bibitem[Blocki et~al.(2012)Blocki, Blum, Datta, and Sheffet]{blocki2012johnsonlindenstrausstransformpreservesdifferential}
Jeremiah Blocki, Avrim Blum, Anupam Datta, and Or~Sheffet.
\newblock The johnson-lindenstrauss transform itself preserves differential privacy, 2012.
\newblock URL \url{https://arxiv.org/abs/1204.2136}.

\bibitem[Boyd et~al.(2019)]{boyd2019differential2020}
Danah Boyd et~al.
\newblock Differential privacy in the 2020 decennial census and the implications for available data products.
\newblock \emph{arXiv preprint arXiv:1907.03639}, 2019.

\bibitem[Bradley and Terry(1952)]{bradley1952rank}
Ralph~Allan Bradley and Milton~E Terry.
\newblock Rank analysis of incomplete block designs: I. the method of paired comparisons.
\newblock \emph{Biometrika}, 39\penalty0 (3/4):\penalty0 324--345, 1952.

\bibitem[Cen et~al.(2022)Cen, Cheng, Chen, Wei, and Chi]{cen2022fast}
Shicong Cen, Chen Cheng, Yuxin Chen, Yuting Wei, and Yuejie Chi.
\newblock Fast global convergence of natural policy gradient methods with entropy regularization.
\newblock \emph{Operations Research}, 70\penalty0 (4):\penalty0 2563--2578, 2022.

\bibitem[Chaudhuri and Monteleoni(2008)]{chaudhuri2008privacy}
Kamalika Chaudhuri and Claire Monteleoni.
\newblock Privacy-preserving logistic regression.
\newblock \emph{Advances in neural information processing systems}, 21, 2008.

\bibitem[Chaudhuri et~al.(2011)Chaudhuri, Monteleoni, and Sarwate]{chaudhuri2011differentially}
Kamalika Chaudhuri, Claire Monteleoni, and Anand~D Sarwate.
\newblock Differentially private empirical risk minimization.
\newblock \emph{Journal of Machine Learning Research}, 12\penalty0 (3), 2011.

\bibitem[Chowdhury et~al.(2024)Chowdhury, Zhou, and Natarajan]{chowdhury2024differentially}
Sayak~Ray Chowdhury, Xingyu Zhou, and Nagarajan Natarajan.
\newblock Differentially private reward estimation with preference feedback.
\newblock In \emph{International Conference on Artificial Intelligence and Statistics}, pages 4843--4851. PMLR, 2024.

\bibitem[Cuturi(2013)]{cuturi2013sinkhorn}
Marco Cuturi.
\newblock Sinkhorn distances: Lightspeed computation of optimal transport.
\newblock \emph{Advances in neural information processing systems}, 26, 2013.

\bibitem[Dagan and Feldman(2020)]{dagan2020pac}
Yuval Dagan and Vitaly Feldman.
\newblock Pac learning with stable and private predictions.
\newblock In \emph{Conference on Learning Theory}, pages 1389--1410. PMLR, 2020.

\bibitem[Dankar and El~Emam(2013)]{dankar2013practicing}
Fida~Kamal Dankar and Khaled El~Emam.
\newblock Practicing differential privacy in health care: A review.
\newblock \emph{Trans. Data Priv.}, 6\penalty0 (1):\penalty0 35--67, 2013.

\bibitem[Desfontaines(2021)]{desfontaines2021realworlddp}
Damien Desfontaines.
\newblock A list of real-world uses of differential privacy.
\newblock \url{https://desfontain.es/blog/real-world-differential-privacy.html}, Oct 2021.
\newblock Blog post, updated 2025-08-18.

\bibitem[Dinur and Nissim(2003)]{dinur2003revealing}
Irit Dinur and Kobbi Nissim.
\newblock Revealing information while preserving privacy.
\newblock In \emph{Proceedings of the twenty-second ACM SIGMOD-SIGACT-SIGART symposium on Principles of database systems}, pages 202--210, 2003.

\bibitem[Dwork and Feldman(2018)]{dwork2018privacy}
Cynthia Dwork and Vitaly Feldman.
\newblock Privacy-preserving prediction.
\newblock In \emph{Conference On Learning Theory}, pages 1693--1702. PMLR, 2018.

\bibitem[Dwork et~al.(2006)Dwork, McSherry, Nissim, and Smith]{dwork2006calibrating}
Cynthia Dwork, Frank McSherry, Kobbi Nissim, and Adam Smith.
\newblock Calibrating noise to sensitivity in private data analysis.
\newblock In \emph{Theory of Cryptography: Third Theory of Cryptography Conference, TCC 2006, New York, NY, USA, March 4-7, 2006. Proceedings 3}, pages 265--284. Springer, 2006.

\bibitem[Dwork et~al.(2014)Dwork, Roth, et~al.]{dwork2014algorithmic}
Cynthia Dwork, Aaron Roth, et~al.
\newblock The algorithmic foundations of differential privacy.
\newblock \emph{Foundations and Trends{\textregistered} in Theoretical Computer Science}, 9\penalty0 (3--4):\penalty0 211--407, 2014.

\bibitem[Erlingsson et~al.(2014)Erlingsson, Pihur, and Korolova]{erlingsson2014rappor}
{\'U}lfar Erlingsson, Vasyl Pihur, and Aleksandra Korolova.
\newblock Rappor: Randomized aggregatable privacy-preserving ordinal response.
\newblock In \emph{Proceedings of the 2014 ACM SIGSAC conference on computer and communications security}, pages 1054--1067, 2014.

\bibitem[{Federal Reserve Board}(2024)]{Fed2024ConsumerCompliance}
{Federal Reserve Board}.
\newblock Requirements for commercial products and services.
\newblock \emph{Consumer Compliance Outlook}, \penalty0 (First Issue), 2024.
\newblock URL \url{https://www.consumercomplianceoutlook.org/2024/first-issue/requirements-for-commercial-products-and-services}.
\newblock Accessed September 30, 2025.

\bibitem[{Federal Trade Commission}(2024)]{FTC2024SocialMedia}
{Federal Trade Commission}.
\newblock Examining the data practices of social media and video streaming services.
\newblock Technical report, Federal Trade Commission, September 2024.
\newblock URL \url{https://www.ftc.gov/system/files/ftc_gov/pdf/Social-Media-6b-Report-9-11-2024.pdf}.
\newblock Staff Report.

\bibitem[Fontaine et~al.(2019)Fontaine, Berthet, and Perchet]{fontaine2019regularized}
Xavier Fontaine, Quentin Berthet, and Vianney Perchet.
\newblock Regularized contextual bandits.
\newblock In \emph{The 22nd International Conference on Artificial Intelligence and Statistics}, pages 2144--2153. PMLR, 2019.

\bibitem[Fu et~al.(2025)Fu, Chen, Gao, and Li]{fu2025privacy}
Xingyu Fu, Ningyuan Chen, Pin Gao, and Yang Li.
\newblock Privacy-preserving personalized recommender systems.
\newblock \emph{Manufacturing \& Service Operations Management}, 2025.

\bibitem[Geist et~al.(2019)Geist, Scherrer, and Pietquin]{geist2019theory}
Matthieu Geist, Bruno Scherrer, and Olivier Pietquin.
\newblock A theory of regularized markov decision processes.
\newblock In \emph{International conference on machine learning}, pages 2160--2169. PMLR, 2019.

\bibitem[Gershoff(2025)]{gershoff2025kanonymousabtesting}
Matthew Gershoff.
\newblock K-anonymous a/b testing, 2025.
\newblock URL \url{https://arxiv.org/abs/2501.14329}.

\bibitem[Ghazi et~al.(2021)Ghazi, Golowich, Kumar, Manurangsi, and Zhang]{ghazi2021deep}
Badih Ghazi, Noah Golowich, Ravi Kumar, Pasin Manurangsi, and Chiyuan Zhang.
\newblock Deep learning with label differential privacy.
\newblock \emph{Advances in neural information processing systems}, 34:\penalty0 27131--27145, 2021.

\bibitem[Haarnoja et~al.(2017)Haarnoja, Tang, Abbeel, and Levine]{haarnoja2017reinforcement}
Tuomas Haarnoja, Haoran Tang, Pieter Abbeel, and Sergey Levine.
\newblock Reinforcement learning with deep energy-based policies.
\newblock In \emph{International conference on machine learning}, pages 1352--1361. PMLR, 2017.

\bibitem[Haarnoja et~al.(2018)Haarnoja, Zhou, Abbeel, and Levine]{haarnoja2018soft}
Tuomas Haarnoja, Aurick Zhou, Pieter Abbeel, and Sergey Levine.
\newblock Soft actor-critic: Off-policy maximum entropy deep reinforcement learning with a stochastic actor.
\newblock In \emph{International conference on machine learning}, pages 1861--1870. Pmlr, 2018.

\bibitem[Han et~al.(2021)Han, Liang, Wang, and Zhang]{han2021generalized}
Yuxuan Han, Zhipeng Liang, Yang Wang, and Jiheng Zhang.
\newblock Generalized linear bandits with local differential privacy.
\newblock \emph{Advances in Neural Information Processing Systems}, 34:\penalty0 26511--26522, 2021.

\bibitem[Hassan et~al.(2019)Hassan, Rehmani, and Chen]{hassan2019differential}
Muneeb~Ul Hassan, Mubashir~Husain Rehmani, and Jinjun Chen.
\newblock Differential privacy techniques for cyber physical systems: A survey.
\newblock \emph{IEEE Communications Surveys \& Tutorials}, 22\penalty0 (1):\penalty0 746--789, 2019.

\bibitem[Jin et~al.(2021)Jin, Yang, and Wang]{jin2021pessimism}
Ying Jin, Zhuoran Yang, and Zhaoran Wang.
\newblock Is pessimism provably efficient for offline rl?
\newblock In \emph{International Conference on Machine Learning}, pages 5084--5096. PMLR, 2021.

\bibitem[Kazerouni et~al.(2017)Kazerouni, Ghavamzadeh, Abbasi~Yadkori, and Van~Roy]{kazerouni2017conservative}
Abbas Kazerouni, Mohammad Ghavamzadeh, Yasin Abbasi~Yadkori, and Benjamin Van~Roy.
\newblock Conservative contextual linear bandits.
\newblock \emph{Advances in Neural Information Processing Systems}, 30, 2017.

\bibitem[Kong and Yang(2022)]{kong2022provably}
Dingwen Kong and Lin Yang.
\newblock Provably feedback-efficient reinforcement learning via active reward learning.
\newblock \emph{Advances in Neural Information Processing Systems}, 35:\penalty0 11063--11078, 2022.

\bibitem[Korkmaz and Brown-Cohen(2024)]{korkmaz2024learning}
Ezgi Korkmaz and Jonah Brown-Cohen.
\newblock Learning differentially private rewards from human feedback, 2024.
\newblock URL \url{https://openreview.net/forum?id=reBq1gmlhS}.

\bibitem[Li et~al.(2022)Li, Ma, and Srebro]{li2022pessimism}
Gene Li, Cong Ma, and Nati Srebro.
\newblock Pessimism for offline linear contextual bandits using $\ell_p $ confidence sets.
\newblock \emph{Advances in Neural Information Processing Systems}, 35:\penalty0 20974--20987, 2022.

\bibitem[Li et~al.(2010)Li, Chu, Langford, and Schapire]{li2010contextual}
Lihong Li, Wei Chu, John Langford, and Robert~E Schapire.
\newblock A contextual-bandit approach to personalized news article recommendation.
\newblock In \emph{Proceedings of the 19th international conference on World wide web}, pages 661--670, 2010.

\bibitem[Liu et~al.(2024)Liu, Zhang, Yang, and Meng]{liu2024survey}
WeiKang Liu, Yanchun Zhang, Hong Yang, and Qinxue Meng.
\newblock A survey on differential privacy for medical data analysis.
\newblock \emph{Annals of Data Science}, 11\penalty0 (2):\penalty0 733--747, 2024.

\bibitem[Maniar et~al.(2021)Maniar, Akkinepally, and Sharma]{maniar2021differential}
Tabish Maniar, Alekhya Akkinepally, and Anantha Sharma.
\newblock Differential privacy for credit risk model.
\newblock \emph{arXiv preprint arXiv:2106.15343}, 2021.

\bibitem[McSherry and Talwar(2007)]{4389483}
Frank McSherry and Kunal Talwar.
\newblock Mechanism design via differential privacy.
\newblock In \emph{48th Annual IEEE Symposium on Foundations of Computer Science (FOCS'07)}, pages 94--103, 2007.
\newblock \doi{10.1109/FOCS.2007.66}.

\bibitem[Mei et~al.(2020)Mei, Xiao, Szepesvari, and Schuurmans]{mei2020global}
Jincheng Mei, Chenjun Xiao, Csaba Szepesvari, and Dale Schuurmans.
\newblock On the global convergence rates of softmax policy gradient methods.
\newblock In \emph{International conference on machine learning}, pages 6820--6829. PMLR, 2020.

\bibitem[Ou et~al.(2024)Ou, Medina, and Cummings]{ou2024thompsonsamplingdifferentiallyprivate}
Tingting Ou, Marco~Avella Medina, and Rachel Cummings.
\newblock Thompson sampling itself is differentially private, 2024.
\newblock URL \url{https://arxiv.org/abs/2407.14879}.

\bibitem[Ouyang et~al.(2022{\natexlab{a}})Ouyang, Wu, Jiang, Almeida, Wainwright, Mishkin, Zhang, Agarwal, Slama, Ray, Schulman, Hilton, Kelton, Miller, Simens, Askell, Welinder, Christiano, Leike, and Lowe]{NEURIPS2022_b1efde53}
Long Ouyang, Jeffrey Wu, Xu~Jiang, Diogo Almeida, Carroll Wainwright, Pamela Mishkin, Chong Zhang, Sandhini Agarwal, Katarina Slama, Alex Ray, John Schulman, Jacob Hilton, Fraser Kelton, Luke Miller, Maddie Simens, Amanda Askell, Peter Welinder, Paul~F Christiano, Jan Leike, and Ryan Lowe.
\newblock Training language models to follow instructions with human feedback.
\newblock In S.~Koyejo, S.~Mohamed, A.~Agarwal, D.~Belgrave, K.~Cho, and A.~Oh, editors, \emph{Advances in Neural Information Processing Systems}, volume~35, pages 27730--27744. Curran Associates, Inc., 2022{\natexlab{a}}.
\newblock URL \url{https://proceedings.neurips.cc/paper_files/paper/2022/file/b1efde53be364a73914f58805a001731-Paper-Conference.pdf}.

\bibitem[Ouyang et~al.(2022{\natexlab{b}})Ouyang, Wu, Jiang, Almeida, Wainwright, Mishkin, Zhang, Agarwal, Slama, Ray, et~al.]{ouyang2022training}
Long Ouyang, Jeffrey Wu, Xu~Jiang, Diogo Almeida, Carroll Wainwright, Pamela Mishkin, Chong Zhang, Sandhini Agarwal, Katarina Slama, Alex Ray, et~al.
\newblock Training language models to follow instructions with human feedback.
\newblock \emph{Advances in neural information processing systems}, 35:\penalty0 27730--27744, 2022{\natexlab{b}}.

\bibitem[Ponnoprat(2023)]{ponnoprat2023dirichlet}
Donlapark Ponnoprat.
\newblock Dirichlet mechanism for differentially private {KL} divergence minimization.
\newblock \emph{Transactions on Machine Learning Research}, 2023.
\newblock ISSN 2835-8856.
\newblock URL \url{https://openreview.net/forum?id=lmr2WwlaFc}.

\bibitem[Qiao and Wang(2023)]{qiao2023offlinereinforcementlearningdifferential}
Dan Qiao and Yu-Xiang Wang.
\newblock Offline reinforcement learning with differential privacy, 2023.
\newblock URL \url{https://arxiv.org/abs/2206.00810}.

\bibitem[Rafailov et~al.(2024)Rafailov, Sharma, Mitchell, Ermon, Manning, and Finn]{rafailov2024directpreferenceoptimizationlanguage}
Rafael Rafailov, Archit Sharma, Eric Mitchell, Stefano Ermon, Christopher~D. Manning, and Chelsea Finn.
\newblock Direct preference optimization: Your language model is secretly a reward model, 2024.
\newblock URL \url{https://arxiv.org/abs/2305.18290}.

\bibitem[Rashidinejad et~al.(2021)Rashidinejad, Zhu, Ma, Jiao, and Russell]{rashidinejad2021bridging}
Paria Rashidinejad, Banghua Zhu, Cong Ma, Jiantao Jiao, and Stuart Russell.
\newblock Bridging offline reinforcement learning and imitation learning: A tale of pessimism.
\newblock \emph{Advances in Neural Information Processing Systems}, 34:\penalty0 11702--11716, 2021.

\bibitem[Rio et~al.(2025)Rio, Barlier, and Colin]{rio2025differentially}
Alexandre Rio, Merwan Barlier, and Igor Colin.
\newblock Differentially private policy gradient.
\newblock \emph{arXiv preprint arXiv:2501.19080}, 2025.

\bibitem[Schulman et~al.(2015)Schulman, Levine, Abbeel, Jordan, and Moritz]{schulman2015trust}
John Schulman, Sergey Levine, Pieter Abbeel, Michael Jordan, and Philipp Moritz.
\newblock Trust region policy optimization.
\newblock In \emph{International conference on machine learning}, pages 1889--1897. PMLR, 2015.

\bibitem[Schulman et~al.(2017{\natexlab{a}})Schulman, Levine, Moritz, Jordan, and Abbeel]{schulman2017trustregionpolicyoptimization}
John Schulman, Sergey Levine, Philipp Moritz, Michael~I. Jordan, and Pieter Abbeel.
\newblock Trust region policy optimization, 2017{\natexlab{a}}.
\newblock URL \url{https://arxiv.org/abs/1502.05477}.

\bibitem[Schulman et~al.(2017{\natexlab{b}})Schulman, Wolski, Dhariwal, Radford, and Klimov]{schulman2017proximal}
John Schulman, Filip Wolski, Prafulla Dhariwal, Alec Radford, and Oleg Klimov.
\newblock Proximal policy optimization algorithms.
\newblock \emph{arXiv preprint arXiv:1707.06347}, 2017{\natexlab{b}}.

\bibitem[Shariff and Sheffet(2018)]{shariff2018differentially}
Roshan Shariff and Or~Sheffet.
\newblock Differentially private contextual linear bandits.
\newblock \emph{Advances in Neural Information Processing Systems}, 31, 2018.

\bibitem[Shi et~al.(2022)Shi, Li, Wei, Chen, and Chi]{shi2022pessimistic}
Laixi Shi, Gen Li, Yuting Wei, Yuxin Chen, and Yuejie Chi.
\newblock Pessimistic q-learning for offline reinforcement learning: Towards optimal sample complexity.
\newblock In \emph{International conference on machine learning}, pages 19967--20025. PMLR, 2022.

\bibitem[Song et~al.(2024)Song, Swamy, Singh, Bagnell, and Sun]{song2024importanceonlinedataunderstanding}
Yuda Song, Gokul Swamy, Aarti Singh, J.~Andrew Bagnell, and Wen Sun.
\newblock The importance of online data: Understanding preference fine-tuning via coverage, 2024.
\newblock URL \url{https://arxiv.org/abs/2406.01462}.

\bibitem[Syrgkanis et~al.(2016)Syrgkanis, Krishnamurthy, and Schapire]{syrgkanis2016efficientalgorithmsadversarialcontextual}
Vasilis Syrgkanis, Akshay Krishnamurthy, and Robert~E. Schapire.
\newblock Efficient algorithms for adversarial contextual learning, 2016.
\newblock URL \url{https://arxiv.org/abs/1602.02454}.

\bibitem[van~der Maaten and Hannun(2020)]{van2020trade}
Laurens van~der Maaten and Awni Hannun.
\newblock The trade-offs of private prediction.
\newblock \emph{arXiv preprint arXiv:2007.05089}, 2020.

\bibitem[Vietri et~al.(2020)Vietri, Balle, Krishnamurthy, and Wu]{vietri2020private}
Giuseppe Vietri, Borja Balle, Akshay Krishnamurthy, and Steven Wu.
\newblock Private reinforcement learning with pac and regret guarantees.
\newblock In \emph{International Conference on Machine Learning}, pages 9754--9764. PMLR, 2020.

\bibitem[Villar et~al.(2015)Villar, Bowden, and Wason]{villar2015multi}
Sof{\'\i}a~S Villar, Jack Bowden, and James Wason.
\newblock Multi-armed bandit models for the optimal design of clinical trials: benefits and challenges.
\newblock \emph{Statistical science: a review journal of the Institute of Mathematical Statistics}, 30\penalty0 (2):\penalty0 199, 2015.

\bibitem[Wang and Zhu(2024)]{wang2024optimal}
Siwei Wang and Jun Zhu.
\newblock Optimal learning policies for differential privacy in multi-armed bandits.
\newblock \emph{Journal of Machine Learning Research}, 25\penalty0 (314):\penalty0 1--52, 2024.

\bibitem[Watson et~al.(2020)Watson, Rozemberczki, and Sarkar]{watson2020stability}
Lauren Watson, Benedek Rozemberczki, and Rik Sarkar.
\newblock Stability enhanced privacy and applications in private stochastic gradient descent.
\newblock \emph{arXiv preprint arXiv:2006.14360}, 2020.

\bibitem[Woo et~al.(2025)Woo, Joshi, and Chi]{woo2025blessing}
Jiin Woo, Gauri Joshi, and Yuejie Chi.
\newblock The blessing of heterogeneity in federated q-learning: Linear speedup and beyond.
\newblock \emph{Journal of Machine Learning Research}, 26\penalty0 (26):\penalty0 1--85, 2025.

\bibitem[Wu et~al.(2023)Wu, Inan, Backurs, Chandrasekaran, Kulkarni, and Sim]{wu2023privately}
Fan Wu, Huseyin~A Inan, Arturs Backurs, Varun Chandrasekaran, Janardhan Kulkarni, and Robert Sim.
\newblock Privately aligning language models with reinforcement learning.
\newblock \emph{arXiv preprint arXiv:2310.16960}, 2023.

\bibitem[Xiong et~al.(2023)Xiong, Dong, Ye, Wang, Zhong, Ji, Jiang, and Zhang]{xiong2023iterative}
Wei Xiong, Hanze Dong, Chenlu Ye, Ziqi Wang, Han Zhong, Heng Ji, Nan Jiang, and Tong Zhang.
\newblock Iterative preference learning from human feedback: Bridging theory and practice for rlhf under kl-constraint.
\newblock \emph{arXiv preprint arXiv:2312.11456}, 2023.

\bibitem[Zhang et~al.(2025)Zhang, Lei, Ding, Li, Xiang, Xu, Xu, and Wang]{zhang2025towards}
Jiaming Zhang, Mingxi Lei, Meng Ding, Mengdi Li, Zihang Xiang, Difei Xu, Jinhui Xu, and Di~Wang.
\newblock Towards user-level private reinforcement learning with human feedback.
\newblock \emph{arXiv preprint arXiv:2502.17515}, 2025.

\bibitem[Zhao et~al.(2024)Zhao, Ye, Gu, and Zhang]{zhao2024sharp}
Heyang Zhao, Chenlu Ye, Quanquan Gu, and Tong Zhang.
\newblock Sharp analysis for kl-regularized contextual bandits and rlhf.
\newblock \emph{arXiv preprint arXiv:2411.04625}, 2024.

\bibitem[Zheng et~al.(2020)Zheng, Cai, Huang, Li, and Wang]{zheng2020locally}
Kai Zheng, Tianle Cai, Weiran Huang, Zhenguo Li, and Liwei Wang.
\newblock Locally differentially private (contextual) bandits learning.
\newblock \emph{Advances in Neural Information Processing Systems}, 33:\penalty0 12300--12310, 2020.

\bibitem[Zhu et~al.(2023)Zhu, Jordan, and Jiao]{pmlr-v202-zhu23f}
Banghua Zhu, Michael Jordan, and Jiantao Jiao.
\newblock Principled reinforcement learning with human feedback from pairwise or k-wise comparisons.
\newblock In Andreas Krause, Emma Brunskill, Kyunghyun Cho, Barbara Engelhardt, Sivan Sabato, and Jonathan Scarlett, editors, \emph{Proceedings of the 40th International Conference on Machine Learning}, volume 202 of \emph{Proceedings of Machine Learning Research}, pages 43037--43067. PMLR, 23--29 Jul 2023.
\newblock URL \url{https://proceedings.mlr.press/v202/zhu23f.html}.

\bibitem[Ziegler et~al.(2020)Ziegler, Stiennon, Wu, Brown, Radford, Amodei, Christiano, and Irving]{ziegler2020finetuninglanguagemodelshuman}
Daniel~M. Ziegler, Nisan Stiennon, Jeffrey Wu, Tom~B. Brown, Alec Radford, Dario Amodei, Paul Christiano, and Geoffrey Irving.
\newblock Fine-tuning language models from human preferences, 2020.
\newblock URL \url{https://arxiv.org/abs/1909.08593}.

\end{thebibliography}

\newpage
\appendix

\section{Proof of \Cref{lem:KL_regularization_privacy}}
\label{app:proof_KL_regularization_privacy}
First, notice that for arbitrary subset $\sS\subseteq \sA$ and neighboring datasets $\sD,\sD'$, the ratio
\begin{equation*}
    \frac{\sum_{a\in \sS}\pi_0(a)\exp\left(\frac{r(a;\sD')}{\eta}\right)}{\sum_{a\in \sS}\pi_0(a)\exp\left(\frac{r(a;\sD)}{\eta}\right)}\leq \exp\left(\frac{\Delta r}{\eta}\right),
\end{equation*}
because the bound $\pi_0(a)\exp\left(\frac{r(a;\sD')-r(a;\sD)}{\eta}\right)\leq \exp\left(\frac{\Delta r}{\eta}\right)$ holds for each individual term in this sum.
For the pure DP case, notice that
for neighboring datasets $\sD$ and $\sD'$, fix action $a$ and compute the probability ratio of $a$ being sampled given $x$, we have:
\begin{equation*}\begin{aligned}
    \frac{\widetilde{\pi}(a;\sD)}{\widetilde{\pi}(a;\sD')}=&\frac{\pi_0(a)\exp\left(\frac{r(a;\sD)}{\eta}\right)}{\sum_{a'}\pi_0(a')\exp\left(\frac{r(a';\sD)}{\eta}\right)}\Bigg / \frac{\pi_0(a)\exp\left(\frac{r(a;\sD')}{\eta}\right)}{\sum_{a'}\pi_0(a')\exp\left(\frac{r(a';\sD')}{\eta}\right)}\\
    =&\exp\left(\frac{r(a;\sD)-r(a;\sD')}{\eta}\right)\frac{\sum_{a'}\pi_0(a')\exp\left(\frac{r(a';\sD')}{\eta}\right)}{\sum_{a'}\pi_0(a')\exp\left(\frac{r(a';\sD)}{\eta}\right)}\\
    \leq& \exp\left(\frac{2\Delta r}{\eta}\right).
\end{aligned}\end{equation*}

For the more general case, consider the following ratio, we have:
\begin{equation*}
    \begin{aligned}
        &\frac{\sum_{a'\in\sA}\pi_0(a')\exp\left(\frac{r(a';\sD')}{\eta}\right)}{\sum_{a'\in\sA}\pi_0(a')\exp\left(\frac{r(a';\sD)}{\eta}\right)}\\
        =&\frac{\sum_{a'\in\bar{\mathcal{I}}}\pi_0(a')\exp\left(\frac{r(a';\sD')}{\eta}\right)+\sum_{a'\in\mathcal{I}}\pi_0(a')\exp\left(\frac{r(a';\sD')}{\eta}\right)}{\sum_{a'\in\sA}\pi_0(a')\exp\left(\frac{r(a';\sD)}{\eta}\right)}\\
        \leq & \frac{\sum_{a'\in\mathcal{I}}\pi_0(a')\exp\left(\frac{r(a';\sD')}{\eta}\right)}{\sum_{a'\in\mathcal{I}}\pi_0(a')\exp\left(\frac{r(a';\sD)}{\eta}\right)}+\frac{\sum_{a'\in\bar{\mathcal{I}}}\pi_0(a')\exp\left(\frac{r(a';\sD')}{\eta}\right)}{\sum_{a'\in\sA}\pi_0(a')\exp\left(\frac{r(a';\sD)}{\eta}\right)}\\
        \leq &\frac{\sum_{a'\in\mathcal{I}}\pi_0(a')\exp\left(\frac{r(a';\sD')}{\eta}\right)}{\sum_{a'\in\mathcal{I}}\pi_0(a')\exp\left(\frac{r(a';\sD)}{\eta}\right)}+\delta_0\frac{\sum_{a'\in\mathcal{A}}\pi_0(a')\exp\left(\frac{r(a';\sD')}{\eta}\right)}{\sum_{a'\in\sA}\pi_0(a')\exp\left(\frac{r(a';\sD)}{\eta}\right)}\\
    \end{aligned}
\end{equation*}
where the last inequality follows from the second condition that $\widetilde{\pi}(\cdot;\mathcal{D})\leq \delta_0$. Rearranging terms we obtain:
\begin{equation*}\begin{aligned}
    \frac{\sum_{a'\in\sA}\pi_0(a')\exp\left(\frac{r(a';\sD')}{\eta}\right)}{\sum_{a'\in\sA}\pi_0(a')\exp\left(\frac{r(a';\sD)}{\eta}\right)}\leq &\frac{1}{1-\delta_0}\frac{\sum_{a'\in\mathcal{I}}\pi_0(a')\exp\left(\frac{r(a';\sD')}{\eta}\right)}{\sum_{a'\in\mathcal{I}}\pi_0(a')\exp\left(\frac{r(a';\sD)}{\eta}\right)}\\
    \leq& \frac{1}{1-\delta_0}\exp\left(\frac{\Delta}{\eta}\right)
\end{aligned}\end{equation*}

hence, for all $a\in\mathcal{I}$ we have:
\begin{equation*}
    \begin{aligned}
    &\frac{\widetilde{\pi}(a;\sD)}{\widetilde{\pi}(a;\sD')}\\
    =&\frac{\pi_0(a)\exp\left(\frac{r(a;\sD)}{\eta}\right)}{\sum_{a'\in\sA}\pi_0(a')\exp\left(\frac{r(a';\sD)}{\eta}\right)}\Bigg / \frac{\pi_0(a)\exp\left(\frac{r(a;\sD')}{\eta}\right)}{\sum_{a'\in\sA}\pi_0(a')\exp\left(\frac{r(a';\sD')}{\eta}\right)}\\
    =&\exp\left(\frac{r(a;\sD)-r(a;\sD')}{\eta}\right)\frac{\sum_{a'\in\sA}\pi_0(a')\exp\left(\frac{r(a';\sD')}{\eta}\right)}{\sum_{a'\in\sA}\pi_0(a')\exp\left(\frac{r(a';\sD)}{\eta}\right)}\\
    \leq& \frac{1}{1-\delta_0}\exp\left(\frac{2\Delta}{\eta}\right)
\end{aligned}
\end{equation*}
so that $\widetilde{\pi}(a;\sD)\leq \frac{1}{1-\delta_0}\exp\left(\frac{2\Delta}{\eta}\right) \widetilde{\pi}(a;\sD')$.

For all $a\in\bar{\mathcal{I}}$, $\widetilde{\pi}(a;\sD')\leq \delta_0$, and therefore, for arbitrary $\mathcal{S}\subseteq \sA$, we have:
\begin{equation*}\begin{aligned}
    \Pr\left(\widetilde{\pi}(\cdot;\sD)\in \sS \right)=&\Pr\left(\widetilde{\pi}(\cdot;\sD)\in \sS \cap \mathcal{I}\right)+\Pr\left(\widetilde{\pi}(\cdot;\sD)\in \sS \cap\bar{\mathcal{I}} \right)\\
    \leq & \frac{1}{1-\delta_0}\exp\left(\frac{2\Delta}{\eta}\right)\Pr\left(\widetilde{\pi}(\cdot;\sD')\in \sS \cap \mathcal{I}\right)+\Pr\left(\widetilde{\pi}(\cdot;\sD)\in \bar{\mathcal{I}} \right)\\
    \leq & \frac{1}{1-\delta_0}\exp\left(\frac{2\Delta}{\eta}\right)\Pr\left(\widetilde{\pi}(\cdot;\sD')\in \sS \right)+\delta_0
\end{aligned}\end{equation*}

which completes the proof.

\section{Proof of \Cref{thm:privacy_multi_armed_bandits}}\label{app:proof_privacy_multi_armed_bandits}
Here we provide the proof of \Cref{thm:privacy_multi_armed_bandits}. 
\paragraph{Constructing low probability event.} Given a selected threshold $N_0$ and any dataset $\sD'$, consider the probability that the action $A$ sampled from $\widetilde{\pi}(\cdot;\sD')$ has count upper bounded by $N_0+1$, we have the following bound:
\begin{equation*}\begin{aligned}
    &\Pr_{a\sim\widetilde{\pi}(\cdot;\sD')}(N(a;\sD')\leq N_0+1)\\
    \leq & \frac{\sum_a \pi_0(a) \mathbb{I}[N(a;\sD')\leq N_0+1]\exp\left((\bar{r}(a;\sD')-\beta_0\Gamma(a;\sD'))/\eta\right)}{\sum_a \pi_0(a)\exp\left((\bar{r}(a;\sD')-\beta_0\Gamma(a;\sD'))/\eta\right)}\\
    \leq & \frac{|\sA|\exp\left(\frac{1}{\eta}(\max_{a'}\{\bar{r}(a;\sD')+\eta\log \pi_0(a)\}-\beta_0/\sqrt{N_0+1})\right)}{\exp\left(\frac{1}{\eta}\max_{\hat{a}}\{\bar{r}(\hat{a};\sD')+\eta\log \pi_0(\hat{a})-\beta_0/\sqrt{N(\hat{a};\sD')}\}\right)}\\
    = & |\sA|\exp\left(\frac{1}{\eta}(\max_{a'}\{\bar{r}(a;\sD')+\eta\log \pi_0(a)\}-\beta_0/\sqrt{N_0+1})-\frac{1}{\eta}\max_{\hat{a}}\{\bar{r}(\hat{a};\sD')+\eta\log \pi_0(\hat{a})-\beta_0/\sqrt{N(\hat{a};\sD')}\}\right).
\end{aligned}\end{equation*}
Let $\bar{a}=\argmax_{a'}\{\bar{r}(a;\sD')+\eta\log \pi_0(a)\}$, we have that
\begin{equation*}\begin{aligned}
    &\max_{a'}\{\bar{r}(a;\sD')+\eta\log \pi_0(a)\}-\max_{\hat{a}}\{\bar{r}(\hat{a};\sD')+\eta\log \pi_0(\hat{a})-\beta_0/\sqrt{N(\hat{a};\sD')}\}\\
    =&\bar{r}(\bar{a};\sD')+\eta\log \pi_0(\bar{a})-\max_{\hat{a}}\{\bar{r}(\hat{a};\sD')+\eta\log \pi_0(\hat{a})-\beta_0/\sqrt{N(\hat{a};\sD')}\}\\
    \leq& \bar{r}(\bar{a};\sD')+\eta\log \pi_0(\bar{a})-(\bar{r}(\bar{a};\sD')+\eta\log \pi_0(\bar{a})-\beta_0/\sqrt{N(\bar{a};\sD')})\\
    =&\beta_0/\sqrt{N(\bar{a};\sD')},
\end{aligned}\end{equation*}
and therefore, we have
\begin{equation}\label{eq:multi_armed_bandit_low_prob_event_case1}\begin{aligned}
    &\Pr_{a\sim\widetilde{\pi}(\cdot;\sD')}(N(a;\sD')\leq N_0+1)
    \leq |\sA|\exp\left(\frac{\beta_0}{\eta}\left(\frac{1}{\sqrt{N(\bar{a};\sD')}}-\frac{1}{\sqrt{N_0+1}}\right)\right).
\end{aligned}\end{equation}
Additionally, let $\tilde{a}=\argmin_{\hat{a}}\{\beta_0/\sqrt{N(\hat{a};\sD)}\}$, we have:
\begin{equation*}
    \begin{aligned}
        &\max_{a'}\{\bar{r}(a;\sD')+\eta\log \pi_0(a)\}-\max_{\hat{a}}\{\bar{r}(\hat{a};\sD')+\eta\log \pi_0(\hat{a})-\beta_0/\sqrt{N(\hat{a};\sD')}\}\\
        \leq &\max_{a'}\{\bar{r}(a;\sD')+\eta\log \pi_0(a)\}- (\bar{r}(\tilde{a};\sD')+\eta\log \pi_0(\tilde{a})-\beta_0/\sqrt{N(\tilde{a};\sD')})\\
        \leq & R+\eta\sigma +\beta_0/\sqrt{N(\tilde{a};\sD')}\\
        \leq & R+\eta\sigma +\beta_0/\sqrt{N(\tilde{a};\sD)-1},
    \end{aligned}
\end{equation*}
and as a result,
\begin{equation}\label{eq:multi_armed_bandit_low_prob_event_case2}\begin{aligned}
    &\Pr_{a\sim\widetilde{\pi}(\cdot;\sD')}(N(a;\sD')\leq N_0+1)
    \leq |\sA|\exp\left(\frac{R}{\eta}+\sigma+\frac{\beta_0}{\eta}\left(\frac{1}{\max_a\sqrt{N(a;\sD)-1}}-\frac{1}{\sqrt{N_0+1}}\right)\right).
\end{aligned}\end{equation}

\paragraph{Bounding the sensitivity per arm.} Fix arm $a$, we bound the sensitivity of $\bar{r}(a;\sD)-\beta_0\Gamma(a;\sD)$ through bounding separately $\Delta \bar{r}(a;\sD)$ and $\Delta \Gamma(a;\sD)$. For all $\sD'$ such that $\|\sD-\sD'\|_1=1$, we have
\begin{equation*}\begin{aligned}
    &|\bar{r}(a;\sD)-\bar{r}(a;\sD')|\\
    =&\left|\frac{\sum_\sD \mathbb{I}[a_i=a]r_i}{N(a;\sD)}-\frac{\sum_{\sD'} \mathbb{I}[a_i=a]r_i}{N(a;\sD')}\right|\\
    \leq &\left|\frac{\sum_\sD \mathbb{I}[a_i=a]r_i}{N(a;\sD)}-\frac{\sum_{\sD'} \mathbb{I}[a_i=a]r_i}{N(a;\sD)}\right|+\left|\frac{\sum_{\sD'} \mathbb{I}[a_i=a]r_i}{N(a;\sD)}-\frac{\sum_{\sD'} \mathbb{I}[a_i=a]r_i}{N(a;\sD')}\right|\\
    \leq& \frac{R}{N(a;\sD)}+N(a;\sD')R\left(\frac{1}{N(a;\sD)-1}-\frac{1}{N(a;\sD)}\right)\\
    =&\frac{R}{N(a;\sD)}+\frac{R}{N(a;\sD)-1}\\
    \leq& \frac{2R}{N(a;\sD)-1};
\end{aligned}\end{equation*}
and
\begin{equation*}
    \begin{aligned}
        |\Gamma(a;\sD)-\Gamma(a;\sD')|=&\left|\frac{1}{\sqrt{N(a;\sD)}}-\frac{1}{\sqrt{N(a;\sD)-1}}\right|\\
        =&\frac{\sqrt{N(a;\sD)}-\sqrt{N(a;\sD)-1}}{\sqrt{N(a;\sD)(N(a;\sD)-1)}}\\
        =&\frac{1}{\sqrt{N(a;\sD)(N(a;\sD)-1)}(\sqrt{N(a;\sD)-1}+\sqrt{N(a;\sD)})}\\
        \leq& \frac{1}{2(N(a;\sD)-1)^{\frac{3}{2}}}.
    \end{aligned}
\end{equation*}
Therefore, combining two upper bounds above, we obtain:
\begin{equation}\label{eq:MAB_pure_DP}\begin{aligned}
    &\Delta\left(\bar{r}(a;\sD)-\beta_0\Gamma(a;\sD)\right)\\
    \leq& \max_{\|\sD-\sD'\|_1=1} |\bar{r}(a;\sD)-\beta_0\Gamma(a;\sD)-(\bar{r}(a;\sD')-\beta_0\Gamma(a;\sD'))|\\
    \leq& \max_{\|\sD-\sD'\|_1=1} |\bar{r}(a;\sD)-\bar{r}(a;\sD')|+\beta_0\max_{\|\sD-\sD'\|_1=1}|\Gamma(a;\sD)-\Gamma(a;\sD')|\\
    \leq& \frac{2R}{N(a;\sD)-1} +\frac{\beta_0}{2(N(a;\sD)-1)^{\frac{3}{2}}}.
\end{aligned}\end{equation}

\paragraph{Completing the proof.}
For the threshold $N_0$, and the constructed arm set $\mathcal{I}=\{a\in \sA: N(a;\sD)\geq N_0+1\}$, we have that $\forall a\in \mathcal{I}$,
\begin{equation*}
    \Delta\left(\bar{r}(a;\sD)-\beta_0\Gamma(a;\sD)\right)\leq \frac{2R}{N(a;\sD)-1} +\frac{\beta_0}{2(N(a;\sD)-1)^{\frac{3}{2}}}\leq \frac{2R}{N_0} +\frac{\beta_0}{2N_0^{3/2}},
\end{equation*}
and notice that for all $a\in \sA$ and $\sD'$ satisfying $\|\sD-\sD'\|_1\leq 1$, it holds that
\begin{equation*}
    N(a;\sD')\leq N(a;\sD)+1
\end{equation*}
then $\bar{\mathcal{I}}$ is constructed as:
\begin{equation*}
    \bar{\mathcal{I}}=\sA\backslash \mathcal{I}=\{a\in \sA: N(a;\sD)\leq N_0\}
\end{equation*}
the probability that the sampled action $a\sim \widetilde{\pi}(\cdot;\sD')$ is in $\bar{\mathcal{I}}$ is upper bounded by:
\begin{equation*}\begin{aligned}
    \Pr\left(\widetilde{\pi}(\cdot;\sD')\in \bar{\mathcal{I}} \right)=& \sum_{a\in\bar{\sA}}\widetilde{\pi}(a;\sD')\mathbb{I}[N(a;\sD)\leq N_0]\\
    \leq &\sum_{a\in\bar{\sA}}\widetilde{\pi}(a;\sD')\mathbb{I}[N(a;\sD')-1\leq N_0]\\
    \leq &\sum_{a\in\bar{\sA}}\widetilde{\pi}(a;\sD')\mathbb{I}[N(a;\sD')\leq N_0+1]\\
    =&\Pr_{a\sim\widetilde{\pi}(\cdot;\sD')}(N(a;\sD')\leq N_0+1)
\end{aligned}\end{equation*}
using \eqref{eq:multi_armed_bandit_low_prob_event_case1},\eqref{eq:multi_armed_bandit_low_prob_event_case2} we obtain:
\begin{equation*}\begin{aligned}
    \delta_0=&|\sA|\exp\left(\min\left\{\frac{\beta_0}{\eta}\frac{1}{\sqrt{N(\bar{a};\sD')}},\frac{R}{\eta}+\sigma + \frac{\beta_0}{\eta}\frac{1}{\max_a\sqrt{N(a;\sD')}}\right\}-\frac{\beta_0}{\eta}\frac{1}{\sqrt{N_0+1}}\right)\\
    \leq & |\sA|\exp\left(\min\left\{\frac{\beta_0}{\eta}\frac{1}{\sqrt{N(\bar{a};\sD)-1}},\frac{R}{\eta}+\sigma + \frac{\beta_0}{\eta}\frac{1}{\max_a\sqrt{N(a;\sD)-1}}\right\}-\frac{\beta_0}{\eta}\frac{1}{\sqrt{N_0+1}}\right).
\end{aligned}\end{equation*}
Using \Cref{lem:KL_regularization_privacy} completes the proof of \Cref{thm:privacy_multi_armed_bandits} for the approximate DP part. The pure DP part follows from taking $N_0=\min_a N(a;\sD)-1$ in \eqref{eq:MAB_pure_DP} which eliminates the low probability event and using the pure DP part of \Cref{lem:KL_regularization_privacy}.

\subsection{Proof of \Cref{cor:MAB_specification}}
    With such fixed $\beta_0$, since $\max_a N(a;\sD)-1 \geq O(\frac{n}{|\sA|})$, we can see that either $C_1$ dominate or the term $\frac{R}{\eta}$ in $C_2$ dominates. So we let $N_0 = \frac{N(\bar{a};\sD) -3}{2} \geq 1$ to ensure that $\delta \leq |\sA|\exp\left(-\frac{\beta_0}{4\eta\sqrt{N_0 +1}} \right)$. So as long as $\eta \leq \frac{\beta_0}{4\sqrt{N_0+1} \log\frac{|\sA|}{\delta}}$, we have arbitrary small $\delta$ as we want. But to ensure a small enough $\epsilon$ as well, we have 
\begin{align*}
    \epsilon &= \frac{1}{\eta}\left(\frac{4}{N_0}+\frac{\beta_0}{N_0^{3/2}}\right)-\log(1-\delta) \\ \notag
    & = \frac{4\sqrt{N_0+1} \log\frac{|\sA|}{\delta}}{ \beta_0}\left(\frac{4}{N_0}+\frac{\beta_0}{N_0^{3/2}}\right)-\log(1-\delta) = 4\log\frac{|\sA|}{\delta} \left( \frac{4\sqrt{N_0+1}}{N_0\beta_0}+\frac{\sqrt{N_0+1}}{N_0^{3/2}}\right) \notag \\
    &\leq 8\log\frac{|\sA|}{\delta} \left( \frac{4}{\sqrt{N(\bar{a};\sD) -3} \beta_0}+\frac{1}{N(\bar{a};\sD) -3}\right),
\end{align*}
up to an error $\log(1-\delta)\approx \delta$ when $\delta$ is small.

\section{Proof of \Cref{thm:linear_contextual_bandits_privacy}}\label{app:proof_linear_contextual_bandits_privacy}
For removal DP we consider a pair of neighboring datasets given by $\sD$ and $\sD_-$, where $\sD$ is the original dataset and $\sD_-$ is obtained by deleting one entry $(x_i,a_i,r_i)$ from $\sD$: $\sD_-=\sD\backslash (x_i,a_i,r_i)$, and for addition DP we consider $\sD_+=\sD\cup (x_0,a_0,r_0)$ such that $\sD_+$ is obtained by adding one entry $(x_0,a_0,r_0)$ to $\sD$. We first focus on removal DP and then generalize the result to addition DP.

\paragraph{Basic facts on neighboring datasets.} We first present some useful facts regarding the difference induced by $\sD$ and $\sD_-$. First, considering the difference between $\Sigma_\sD^{-1}-\Sigma_{\sD_-}^{-1}$, since $\Sigma_\sD=\Sigma_{\sD_-}+\phi(x_i,a_i)\phi(x_i,a_i)^T$, we can use the Sherman-Morrison formula to obtain:
\begin{equation}\label{eq:proof_lincontbandit_neighbor_Sigma_difference}
    \Sigma_{\sD}^{-1}-\Sigma_{\sD_-}^{-1}=-\frac{\Sigma_{\sD_-}^{-1}\phi(x_i,a_i)\phi(x_i,a_i)^T\Sigma_{\sD_-}^{-1}}{1+\phi(x_i,a_i)^T\Sigma_{\sD_-}^{-1}\phi(x_i,a_i)}.
\end{equation}
For the eigenvelue of $\Sigma_{\sD_-}$, we have:
\begin{equation}\label{eq:proof_lincontbandit_lambdamin_range}
    \lambda_{\min}(\Sigma_{\sD})\leq \lambda_{\min}(\Sigma_{\sD_-})+\|\phi(x_i,a_i)\|_2^2\leq \lambda_{\min}(\Sigma_{\sD_-})+1,
\end{equation}
so that $\lambda_{\min}(\sD)\geq \lambda_{\min}(\Sigma_{\sD_-})\geq \lambda_{\min}(\sD)-1$.

For the $\Sigma_{\sD}^{-1}$ and $\Sigma_{\sD_-}^{-1}$ induced norm, we have:
\begin{equation}\label{eq:proof_lincontbandit_scaled_norm_bound_D}
    \|\phi(x,a)\|_{\Sigma_{\sD}^{-1}}=\sqrt{\phi(x,a)^T \Sigma_{\sD}^{-1} \phi(x,a)}\leq \frac{1}{\sqrt{\lambda_{\min}(\sD)}}\|\phi(x,a)\|_2\leq \frac{1}{\sqrt{\lambda_{\min}(\sD)}},
\end{equation}
and similarly,
\begin{equation}\label{eq:proof_lincontbandit_scaled_norm_bound_D-}
    \|\phi(x,a)\|_{\Sigma_{\sD_-}^{-1}}\leq \frac{1}{\sqrt{\lambda_{\min}(\sD)-1}},
\end{equation}
for the $\Sigma_{\sD_-}^{-1}$ induced inner product, we have:
\begin{equation}\label{eq:proof_lincontbandit_scaled_innerproduct_transfromation}\begin{aligned}
&\langle \phi(x,a),\phi(x_i,a_i)\rangle_{\Sigma_{\sD}^{-1}}\\
    =&\phi(x,a)^T\Sigma_{\sD}^{-1}\phi(x_i,a_i)\\
    =&\phi(x,a)^T\left(\Sigma_{\sD_-}^{-1}-\frac{\Sigma_{\sD_-}^{-1} \phi(x_i,a_i)\phi(x_i,a_i)^T \Sigma_{\sD_-}^{-1}}{1+\phi(x_i,a_i)^T\Sigma_{{\sD_-}}^{-1}\phi(x_i,a_i)}\right)\phi(x_i,a_i)\\
    =&\langle \phi(x,a),\phi(x_i,a_i)\rangle_{\Sigma_{\sD_-}^{-1}}-\frac{\langle \phi(x,a),\phi(x_i,a_i)\rangle_{\Sigma_{\sD_-}^{-1}} \|\phi(x_i,a_i)\|_{\Sigma_{\sD_-}^{-1}}^2}{1+\|\phi(x_i,a_i)\|_{\Sigma_{\sD_-}^{-1}}^2}\\
    =&\frac{\langle \phi(x,a),\phi(x_i,a_i)\rangle_{\Sigma_{\sD_-}^{-1}}}{1+\|\phi(x_i,a_i)\|_{\Sigma_{\sD_-}^{-1}}^2},
\end{aligned}\end{equation}
and similarly,
\begin{equation}\label{eq:proof_lincontbandit_scaled_innerproduct_transfromation_inv}
    \langle \phi(x,a),\phi(x_i,a_i)\rangle_{\Sigma_{\sD_-}^{-1}}=\frac{\langle \phi(x,a),\phi(x_i,a_i)\rangle_{\Sigma_{\sD}^{-1}}}{1-\|\phi(x_i,a_i)\|_{\Sigma_{\sD}^{-1}}^2}.
\end{equation}

\paragraph{Bounding the sensitivity of $\bar{r}$.} We first bound the sensitivity of $\theta(\sD)=\Sigma_\sD^{-1}b_\sD$. We have
\begin{equation*}
    \begin{aligned}
        &\theta(\sD)-\theta(\sD_-)\\
        =&\Sigma_\sD^{-1}b_\sD-\Sigma_{\sD_-}^{-1}b_{\sD_-}\\
        =&(\Sigma_{\sD}^{-1}-\Sigma_{\sD_-}^{-1})b_{\sD_-}+\Sigma_\sD^{-1}(b_\sD-b_{\sD_-})\\
        \stackrel{\text{(i)}}{=}&-\frac{\Sigma_{\sD_-}^{-1}\phi(x_i,a_i)\phi(x_i,a_i)^T\Sigma_{\sD_-}^{-1}}{1+\phi(x_i,a_i)^T\Sigma_{\sD_-}^{-1}\phi(x_i,a_i)}b_{\sD_-}+\Sigma_\sD^{-1}\phi(x_i,a_i)r_i\\
        \stackrel{\text{(ii)}}{=}&-\frac{\Sigma_{\sD_-}^{-1}\phi(x_i,a_i)\phi(x_i,a_i)^T\Sigma_{\sD_-}^{-1}}{1+\phi(x_i,a_i)^T\Sigma_{\sD_-}^{-1}\phi(x_i,a_i)}b_{\sD_-}+\left(\Sigma_{\sD_-}^{-1}-\frac{\Sigma_{\sD_-}^{-1}\phi(x_i,a_i)\phi(x_i,a_i)^T\Sigma_{\sD_-}^{-1}}{1+\phi(x_i,a_i)^T\Sigma_{\sD_-}^{-1}\phi(x_i,a_i)}\right)\phi(x_i,a_i)r_i\\
        =&-\frac{\Sigma_{\sD_-}^{-1}\phi(x_i,a_i)\phi(x_i,a_i)^T\Sigma_{\sD_-}^{-1}}{1+\phi(x_i,a_i)^T\Sigma_{\sD_-}^{-1}\phi(x_i,a_i)}b_{\sD_-}+\frac{\Sigma_{\sD_-}^{-1}\phi(x_i,a_i)r_i}{1+\phi(x_i,a_i)^T\Sigma_{\sD_-}^{-1}\phi(x_i,a_i)}\\
        =&\frac{\Sigma_{\sD_-}^{-1}\phi(x_i,a_i)(r_i-b_{\sD_-}^T\Sigma_{\sD_-}^{-1}\phi(x_i,a_i))}{1+\phi(x_i,a_i)^T\Sigma_{\sD_-}^{-1}\phi(x_i,a_i)},
    \end{aligned}
\end{equation*}
where (i) and (ii) hold because of \eqref{eq:proof_lincontbandit_neighbor_Sigma_difference}. Therefore,
\begin{equation}\label{eq:proof_lincontbandit_expanded_reward_difference}
    \begin{aligned}
        &\bar{r}(x,a;\sD)-\bar{r}(x,a;\sD_-)\\
        =&\left(\theta(\sD)-\theta(\sD_-)\right)^T\phi(x,a)\\
        =&\frac{\phi(x,a)^T\Sigma_{\sD_-}^{-1}\phi(x_i,a_i)(r_i-b_{\sD_-}^T\Sigma_{\sD_-}^{-1}\phi(x_i,a_i))}{1+\phi(x_i,a_i)^T\Sigma_{\sD_-}^{-1}\phi(x_i,a_i)}\\
        =&\frac{\langle \phi(x,a),\phi(x_i,a_i)\rangle_{\Sigma_{\sD_-}^{-1}}}{1+\|\phi(x_i,a_i)\|_{\Sigma_{\sD_-}^{-1}}^2}(r_i-b_{\sD_-}^T\Sigma_{\sD_-}^{-1}\phi(x_i,a_i))\\
        \stackrel{\text{(i)}}{=}&\langle \phi(x,a),\phi(x_i,a_i)\rangle_{\Sigma_{\sD}^{-1}}(r_i-b_{\sD_-}^T\Sigma_{\sD_-}^{-1}\phi(x_i,a_i)),
    \end{aligned}
\end{equation}
where (i) follows from \eqref{eq:proof_lincontbandit_scaled_innerproduct_transfromation}. Since
\begin{equation*}
    \begin{aligned}
        &b_{\sD_-}^T\Sigma_{\sD_-}^{-1}\phi(x_i,a_i)\\
        = &\sum_{j\neq i} r_j\langle \phi(x_j,a_j),\phi(x_i,a_i)\rangle_{\Sigma_{\sD_-}^{-1}}\\
        \leq &R\|\phi(x_i,a_i) \|_{\Sigma_{\sD_-}^{-1}}\sum_{j\neq i} \|\phi(x_j,a_j) \|_{\Sigma_{\sD_-}^{-1}}\\
        =&R\|\phi(x_i,a_i) \|_{\Sigma_{\sD_-}^{-1}}\sum_{j\neq i} \sqrt{\phi(x_j,a_j)^T\Sigma_{\sD_-}^{-1}\phi(x_j,a_j)}\\
        \leq & \sqrt{n-1} R\|\phi(x_i,a_i) \|_{\Sigma_{\sD_-}^{-1}}\sqrt{\sum_{j\neq i}\phi(x_j,a_j)^T\Sigma_{\sD_-}^{-1}\phi(x_j,a_j)}.
    \end{aligned}
\end{equation*}
For the term $\sum_{j\neq i}\phi(x_j,a_j)^T\Sigma_{\sD_-}^{-1}\phi(x_j,a_j)$, we have:
\begin{equation*}\begin{aligned}
    &\sum_{j\neq i}\phi(x_j,a_j)^T\Sigma_{\sD_-}^{-1}\phi(x_j,a_j)\\
    \leq& \Tr\left(\Sigma_{\sD_-}^{-1}\left(\sum_{j\neq i}\phi(x_j,a_j)\phi(x_j,a_j)^T\right)\right)\\
    =&\Tr\left(\Sigma_{\sD_-}^{-1}\left(\Sigma_{\sD_-}-\lambda I\right)\right)\\
    =& \Tr(I)-\lambda \Tr(\Sigma_{\sD_-}^{-1})\\
    \leq& d,
\end{aligned}\end{equation*}
which yields:
\begin{equation}\label{eq:proof_lincontbandit_regoutcome_bound}
    \begin{aligned}
        b_{\sD_-}^T\Sigma_{\sD_-}^{-1}\phi(x_i,a_i)\leq &\sqrt{(n-1)d} R\|\phi(x_i,a_i) \|_{\Sigma_{\sD_-}^{-1}}\\
        \leq &\sqrt{\frac{(n-1)d}{\lambda_{\min}(\sD)-1}}R.
    \end{aligned}
\end{equation}
Combining \eqref{eq:proof_lincontbandit_expanded_reward_difference} and \eqref{eq:proof_lincontbandit_regoutcome_bound}, we have:
\begin{equation}\label{eq:reward_sensitivity_bound}
    \begin{aligned}
        &|\bar{r}(x,a;\sD)-\bar{r}(x,a;\sD_-)|\\
        \leq& |\langle \phi(x,a),\phi(x_i,a_i)\rangle_{\Sigma_{\sD}^{-1}}|(1+\sqrt{\frac{(n-1)d}{\lambda_{\min}(\sD)-1}})R\\
        \leq & \|\phi(x,a)\|_{\Sigma_{\sD}^{-1}}\|\phi(x_i,a_i)\|_{\Sigma_{\sD}^{-1}}(1+\sqrt{\frac{(n-1)d}{\lambda_{\min}(\sD)-1}})R\\
        \leq & (1+\sqrt{\frac{(n-1)d}{\lambda_{\min}(\sD)-1}})\frac{R}{\sqrt{\lambda_{\min}(\sD)}}\|\phi(x,a)\|_{\Sigma_{\sD}^{-1}}.
    \end{aligned}
\end{equation}
Similarly, we obtain the bound for $\sD_+=\sD\cup (x_0,a_0,r_0)$ as:
\begin{equation*}
    \theta(\sD)-\theta(\sD_+)=\frac{\Sigma_{\sD_+}^{-1}\phi(x_0,a_0)(-r_0+b_{\sD_+}^T\Sigma_{\sD_+}^{-1}\phi(x_0,a_0))}{1-\phi(x_0,a_0)^T\Sigma_{\sD_+}^{-1}\phi(x_0,a_0)},
\end{equation*}
and correspondingly,
\begin{equation*}\begin{aligned}
    |\bar{r}(x,a;\sD_+)-\bar{r}(x,a;\sD)|=&|\langle \phi(x,a),\phi(x_0,a_0)\rangle_{\Sigma_{\sD}^{-1}}|\cdot|(r_0-b_{\sD_+}^T\Sigma_{\sD_+}^{-1}\phi(x_0,a_0))|\\
    \leq & (1+\sqrt{\frac{nd}{\lambda_{\min}(\sD)}})\frac{R}{\sqrt{\lambda_{\min}(\sD)}}\|\phi(x,a)\|_{\Sigma_{\sD}^{-1}}\\
    \leq & (1+\sqrt{\frac{(n-1)d}{\lambda_{\min}(\sD)-1}})\frac{R}{\sqrt{\lambda_{\min}(\sD)}}\|\phi(x,a)\|_{\Sigma_{\sD}^{-1}}.
\end{aligned}\end{equation*}

\paragraph{Bounding the sensitivity of $\Gamma$.} We now bound the sensitivity of $\Gamma(x,a;\sD)=\|\phi(x,a)\|_{\Sigma_\sD^{-1}}$. We can rewrite the difference as:
\begin{equation*}
    \begin{aligned}
        &\Gamma(x,a;\sD_-)-\Gamma(x,a;\sD)\\
        =&\|\phi(x,a)\|_{\Sigma_{\sD_-}^{-1}}-\|\phi(x,a)\|_{\Sigma_{\sD}^{-1}}\\
        =&\frac{\|\phi(x,a)\|^2_{\Sigma_{\sD_-}^{-1}}-\|\phi(x,a)\|^2_{\Sigma_{\sD}^{-1}}}{\|\phi(x,a)\|_{\Sigma_\sD^{-1}}+\|\phi(x,a)\|_{\Sigma_{\sD_-}^{-1}}}\\
        =&\frac{\phi(x,a)^T\left(\Sigma_{\sD_-}^{-1}-\Sigma_{\sD}^{-1}\right)\phi(x,a)}{\|\phi(x,a)\|_{\Sigma_\sD^{-1}}+\|\phi(x,a)\|_{\Sigma_{\sD_-}^{-1}}}\\
        \stackrel{\text{(i)}}{=}&\frac{1}{\|\phi(x,a)\|_{\Sigma_\sD^{-1}}+\|\phi(x,a)\|_{\Sigma_{\sD_-}^{-1}}}\frac{\langle \phi(x,a),\phi(x_i,a_i)\rangle^2_{\Sigma_{\sD_-}^{-1}}}{1+\|\phi(x_i,a_i)\|^2_{\Sigma_{\sD_-}^{-1}}}\\
        \stackrel{\text{(ii)}}{=}&\frac{\langle \phi(x,a),\phi(x_i,a_i)\rangle_{\Sigma_{\sD_-}^{-1}}}{\|\phi(x,a)\|_{\Sigma_\sD^{-1}}+\|\phi(x,a)\|_{\Sigma_{\sD_-}^{-1}}}\langle \phi(x,a),\phi(x_i,a_i)\rangle_{\Sigma_{\sD}^{-1}}\\
        \stackrel{\text{(iii)}}{=}&\frac{1}{\|\phi(x,a)\|_{\Sigma_\sD^{-1}}+\|\phi(x,a)\|_{\Sigma_{\sD_-}^{-1}}}\frac{\langle \phi(x,a),\phi(x_i,a_i)\rangle^2_{\Sigma_{\sD}^{-1}}}{1-\|\phi(x_i,a_i)\|^2_{\Sigma_{\sD}^{-1}}},
    \end{aligned}
\end{equation*}
where we have used \eqref{eq:proof_lincontbandit_neighbor_Sigma_difference} in (i), \eqref{eq:proof_lincontbandit_scaled_innerproduct_transfromation} in (ii) and \eqref{eq:proof_lincontbandit_scaled_innerproduct_transfromation_inv} in (iii). For the latter inner-product term, we have the upper bound:
\begin{equation*}
    \begin{aligned}
        \frac{\langle \phi(x,a),\phi(x_i,a_i)\rangle^2_{\Sigma_{\sD}^{-1}}}{1-\|\phi(x_i,a_i)\|^2_{\Sigma_{\sD}^{-1}}}\stackrel{\text{(i)}}{\leq} &\frac{\langle \phi(x,a),\phi(x_i,a_i)\rangle^2_{\Sigma_{\sD}^{-1}}}{1-1/\lambda_{\min}(\sD)}\\
        \stackrel{\text{(ii)}}{\leq} & \frac{\|\phi(x,a)\|^2_{\Sigma_{\sD}^{-1}}\|\phi(x_i,a_i)\|^2_{\Sigma_{\sD}^{-1}}}{1-1/\lambda_{\min}(\sD)},
    \end{aligned}
\end{equation*}
where we have used \eqref{eq:proof_lincontbandit_scaled_norm_bound_D} in (i) and Cauchy-Schwarz inequality in (ii).
Therefore, we obtain the final bound on the sensitivity of $\Gamma$ (notice that $\Gamma(x,a;\sD_-)-\Gamma(x,a;\sD)\geq 0$):
\begin{equation}\label{eq:gamma_sensitivity_bound}
    \begin{aligned}
        &\Gamma(x,a;\sD_-)-\Gamma(x,a;\sD)\\
        =&\frac{1}{\|\phi(x,a)\|_{\Sigma_\sD^{-1}}+\|\phi(x,a)\|_{\Sigma_{\sD_-}^{-1}}}\frac{\langle \phi(x,a),\phi(x_i,a_i)\rangle^2_{\Sigma_{\sD}^{-1}}}{1-\|\phi(x_i,a_i)\|^2_{\Sigma_{\sD}^{-1}}}\\
        \leq & \frac{1}{\|\phi(x,a)\|_{\Sigma_\sD^{-1}}+\sqrt{1+1/\lambda_{\min}(\sD)}\|\phi(x,a)\|_{\Sigma_{\sD}^{-1}}}\frac{\|\phi(x,a)\|^2_{\Sigma_{\sD}^{-1}}\|\phi(x_i,a_i)\|^2_{\Sigma_{\sD}^{-1}}}{1-1/\lambda_{\min}(\sD)}\\
        \leq &\frac{1/\lambda_{\min}(\sD)}{(1-1/\lambda_{\min}(\sD))(1+\sqrt{1+1/\lambda_{\min}(\sD)})}\|\phi(x,a)\|_{\Sigma_{\sD}^{-1}}\\
        \leq & \frac{1}{2(\lambda_{\min}(\sD)-1)}\|\phi(x,a)\|_{\Sigma_{\sD}^{-1}}.
    \end{aligned}
\end{equation}
Similarly, we can obtain the bound for $\sD_+=\sD\cup (x_0,a_0,r_0)$ as:
\begin{equation}\label{eq:gamma_sensitivity_bound_D+}\begin{aligned}
    \Gamma(x,a;\sD)-\Gamma(x,a;\sD_+)=&\frac{\langle \phi(x,a),\phi(x_0,a_0)\rangle_{\Sigma_{\sD_+}^{-1}}}{\|\phi(x,a)\|_{\Sigma_\sD^{-1}}+\|\phi(x,a)\|_{\Sigma_{\sD_+}^{-1}}}\langle \phi(x,a),\phi(x_0,a_0)\rangle_{\Sigma_{\sD}^{-1}}\\
    \leq &\frac{\|\phi(x,a)\|_{\Sigma_{\sD_+}^{-1}}\|\phi(x_0,a_0)\|_{\Sigma_{\sD_+}^{-1}}}{2\|\phi(x,a)\|_{\Sigma_{\sD_+}^{-1}}} \|\phi(x,a)\|_{\Sigma_{\sD}^{-1}}\|\phi(x_0,a_0)\|_{\Sigma_{\sD}^{-1}}\\
    \leq &\frac{\|\phi(x_0,a_0)\|_{\Sigma_{\sD}^{-1}}}{2} \|\phi(x,a)\|_{\Sigma_{\sD}^{-1}}\\
    \leq&\frac{1}{2\lambda_{\min}(\sD)}\|\phi(x,a)\|_{\Sigma_{\sD}^{-1}}.
\end{aligned}\end{equation}

\paragraph{Establishing the low-probability event.} For the last step, we construct an event that the selected action has low coverage, more specifically we select a threshold $\lambda_0$ and bound the probability of the event that $\|\phi(x,A)\|_{\Sigma_{\sD'}^{-1}}> \frac{1}{\sqrt{\lambda_0}}$ where $A$ is the action sampled from the policy $\widetilde{\pi}(\cdot|x;{\sD'})$ for dataset $\sD'\in\{\sD_+,\sD,\sD_-\}$. Denoting $\bar{a}=\argmax_{a'} \{b_{{\sD'}}^T\Sigma_{\sD'}^{-1}\phi(x,a')+\eta\log \pi_0(a'|x)\}$ we have that:
\begin{equation}\label{eq:contextual_bandit_low_prob_event}\begin{aligned}
    &\Pr_{a\sim\widetilde{\pi}(\cdot|x,{\sD'})}(\|\phi(x,a)\|_{\Sigma_{\sD'}^{-1}}> \frac{1}{\sqrt{\lambda_0}})\\
    \leq & \frac{\sum_a \pi_0(a|x) \mathbb{I}[\|\phi(x,a)\|_{\Sigma_{\sD'}^{-1}}> \frac{1}{\sqrt{\lambda_0}}]\exp\left((b_{\sD'}^T \Sigma_{\sD'}^{-1}\phi(x,a)-\beta_0\|\phi(x,a)\|_{\Sigma_{\sD'}^{-1}})/\eta\right)}{\sum_a \pi_0(a|x)\exp\left((b_{\sD'}^T \Sigma_{\sD'}^{-1}\phi(x,a)-\beta_0\|\phi(x,a)\|_{\Sigma_{\sD'}^{-1}})/\eta\right)}\\
    \leq & \frac{|\sA|\exp\left(\frac{1}{\eta}(\max_{a'}\{b_{{\sD'}}^T\Sigma_{\sD'}^{-1}\phi(x,a')+\eta\log \pi_0(a'|x)\}-\beta_0\frac{1}{\sqrt{\lambda_0}})\right)}{\exp\left(\frac{1}{\eta}\max_{\hat{a}}\{b_{{\sD'}}^T\Sigma_{\sD'}^{-1}\phi(x,\hat{a})+\eta\log \pi_0(\hat{a}|x)-\beta_0\|\phi(x,\hat{a})\|_{\Sigma_{\sD'}^{-1}}\}\right)}\\
    = & |\sA|\exp(\frac{1}{\eta}(\max_{a'}\{b_{{\sD'}}^T\Sigma_{\sD'}^{-1}\phi(x,a')+\eta\log \pi_0(a'|x)\}-\beta_0\frac{1}{\sqrt{\lambda_0}}\\
    &-\max_{\hat{a}}\{b_{{\sD'}}^T\Sigma_{\sD'}^{-1}\phi(x,\hat{a})+\eta\log \pi_0(\hat{a}|x)-\beta_0\|\phi(x,\hat{a})\|_{\Sigma_{\sD'}^{-1}}\}))\\
    \leq & |\sA|\exp\left(\frac{\beta_0}{\eta}\left( \|\phi(x,\bar{a})\|_{\Sigma_{\sD'}^{-1}}-\frac{1}{\sqrt{\lambda_0}}\right)\right),
\end{aligned}\end{equation}
and in the mean time:
\begin{equation}\label{eq:contextual_bandit_low_prob_event_2}
    \begin{aligned}
        &\Pr_{a\sim\widetilde{\pi}(\cdot|x,{\sD'})}(\|\phi(x,a)\|_{\Sigma_{\sD'}^{-1}}> \frac{1}{\sqrt{\lambda_0}})\\
        \leq & |\sA|\exp\Bigg(\frac{1}{\eta}(\max_{a'}\{b_{{\sD'}}^T\Sigma_{\sD'}^{-1}\phi(x,a')+\eta\log \pi_0(a'|x)\}-\beta_0\frac{1}{\sqrt{\lambda_0}}\\
    &-\max_{\hat{a}}\{b_{{\sD'}}^T\Sigma_{\sD'}^{-1}\phi(x,\hat{a})+\eta\log \pi_0(\hat{a}|x)-\beta_0\|\phi(x,\hat{a})\|_{\Sigma_{\sD'}^{-1}}\})\Bigg)\\
    \leq & |\sA|\exp\left(\sqrt{\frac{(n-1)d}{\lambda_{\min}(\sD)-1}}\frac{R}{\eta} +\sigma+\frac{\beta_0}{\eta} (\min_a \|\phi(x,a)\|_{\Sigma_{\sD'}^{-1}}-\frac{1}{\sqrt{\lambda_0}})\right)
    \end{aligned}
\end{equation}
where we have used \eqref{eq:proof_lincontbandit_regoutcome_bound} in the second inequality.
\eqref{eq:contextual_bandit_low_prob_event} and \eqref{eq:contextual_bandit_low_prob_event_2} suggest that this event happens with exponentially small probability with the selection of a proper $\lambda_0$.

\paragraph{Completing the proof.} With the results above in hand, we are ready to use \Cref{lem:KL_regularization_privacy} to complete the proof of \Cref{thm:linear_contextual_bandits_privacy}. Given dataset $\sD$, select a threshold $\lambda_0$, let
\begin{equation*}
    \mathcal{I}=\{a\in \sA: \|\phi(x,a)\|_{\Sigma_\sD^{-1}}\leq \frac{1}{\sqrt{\lambda_0}}\}
\end{equation*}
denote the set of arms that are sufficiently covered. Consider $\sD'\in\{\sD_+,\sD_-\}$, notice that the following identity holds:
\begin{equation*}
    \|\phi(x,a)\|_{\Sigma_\sD^{-1}}\leq \|\phi(x,a)\|_{\Sigma_{\sD_-}^{-1}}\leq \frac{1}{\sqrt{1-\|\phi(x_i,a_i)\|^2_{\Sigma_{\sD}^{-1}}}}\|\phi(x,a)\|_{\Sigma_\sD^{-1}}
\end{equation*}
and
\begin{equation*}
    \frac{1}{\sqrt{1+\|\phi(x_0,a_0)\|^2_{\Sigma_{\sD}^{-1}}}}\|\phi(x,a)\|_{\Sigma_\sD^{-1}}\leq \|\phi(x,a)\|_{\Sigma_{\sD_+}^{-1}}\leq \|\phi(x,a)\|_{\Sigma_\sD^{-1}},
\end{equation*}
which leads to:
\begin{equation*}
    \|\phi(x,a)\|_{\Sigma_{\sD_-}^{-1}}\leq \frac{\|\phi(x,a)\|_{\Sigma_\sD^{-1}}}{\sqrt{1-\|\phi(x_i,a_i)\|^2_{\Sigma_{\sD}^{-1}}}}\leq \frac{\|\phi(x,a)\|_{\Sigma_\sD^{-1}}}{\sqrt{1-1/\lambda_{\min}(\sD)}}
\end{equation*}
and
\begin{equation*}
    \|\phi(x,a)\|_{\Sigma_{\sD_+}^{-1}}\geq \frac{\|\phi(x,a)\|_{\Sigma_{\sD}^{-1}}}{\sqrt{1+\|\phi(x_0,a_0)\|_{\Sigma_{\sD}^{-1}}^2}}\geq\frac{\|\phi(x,a)\|_{\Sigma_{\sD}^{-1}}}{\sqrt{1+1/\lambda_{\min}(\sD)}}
\end{equation*}
the probability that the sampled action $a\sim \widetilde{\pi}(\cdot;\sD_-)$ is in $\bar{\mathcal{I}}$ can be upper bounded by:
\begin{equation*}\begin{aligned}
    \Pr\left(\widetilde{\pi}(\cdot;\sD_-)\in \bar{\mathcal{I}} \right)=&\sum_{a\in\bar{\sA}}\widetilde{\pi}(a;\sD_-)\mathbb{I}[\|\phi(x,a)\|_{\Sigma_{\sD}^{-1}}>\frac{1}{\sqrt{\lambda_0}}]\\
    \leq&\sum_{a\in\bar{\sA}}\widetilde{\pi}(a;\sD_-)\mathbb{I}\left[\|\phi(x,a)\|_{\Sigma_{\sD_-}^{-1}}>\frac{1}{\sqrt{\lambda_0}}\right]\\
    =&\Pr_{a\sim\widetilde{\pi}(\cdot;\sD_-)}\left(\|\phi(x,a)\|_{\Sigma_{\sD_-}^{-1}}>\frac{1}{\sqrt{\lambda_0}}\right )
\end{aligned}\end{equation*}
and
\begin{equation*}\begin{aligned}
    \Pr\left(\widetilde{\pi}(\cdot;\sD_+)\in \bar{\mathcal{I}} \right)=&\sum_{a\in\bar{\sA}}\widetilde{\pi}(a;\sD_+)\mathbb{I}[\|\phi(x,a)\|_{\Sigma_{\sD}^{-1}}>\frac{1}{\sqrt{\lambda_0}}]\\
    \leq&\sum_{a\in\bar{\sA}}\widetilde{\pi}(a;\sD_+)\mathbb{I}\left[\|\phi(x,a)\|_{\Sigma_{\sD_+}^{-1}}>\frac{1}{\sqrt{\lambda_0(1+1/\lambda_{\min}(\sD))}}\right]\\
    =&\Pr_{a\sim\widetilde{\pi}(\cdot;\sD_+)}\left(\|\phi(x,a)\|_{\Sigma_{\sD_+}^{-1}}>\frac{1}{\sqrt{\lambda_0(1+1/\lambda_{\min}(\sD))}}\right )
\end{aligned}\end{equation*}

We can see that \eqref{eq:contextual_bandit_low_prob_event} implies that:
\begin{equation*}\begin{aligned}
    &\Pr_{a\sim\widetilde{\pi}(\cdot;\sD_-)}\left(\|\phi(x,a)\|_{\Sigma_{\sD_-}^{-1}}>\frac{1}{\sqrt{\lambda_0}}\right )\\
    \leq& |\sA|\exp\left( \frac{\beta_0}{\eta}\left(\|\phi(x,\bar{a})\|_{\Sigma_{\sD_-}^{-1}}-\frac{1}{\sqrt{\lambda_0}}\right)\right)\\
    \leq & |\sA|\exp\left( \frac{\beta_0}{\eta}\left(\frac{\|\phi(x,\bar{a})\|_{\Sigma_\sD^{-1}}}{\sqrt{1-1/\lambda_{\min}(\sD)}}-\frac{1}{\sqrt{\lambda_0}}\right)\right)
\end{aligned}\end{equation*}
and
\begin{equation*}\begin{aligned}
    &\Pr_{a\sim\widetilde{\pi}(\cdot;\sD_+)}\left(\|\phi(x,a)\|_{\Sigma_{\sD_+}^{-1}}>\frac{1}{\sqrt{\lambda_0(1+1/\lambda_{\min}(\sD))}}\right )\\
    \leq &|\sA|\exp\left(\frac{\beta_0}{\eta}\left(\|\phi(x,\bar{a})\|_{\Sigma_{\sD_+}^{-1}}-\frac{1}{\sqrt{\lambda_0(1+1/\lambda_{\min}(\sD))}}\right)\right)\\
    \leq & |\sA|\exp\left(\frac{\beta_0}{\eta}\left(\|\phi(x,\bar{a})\|_{\Sigma_{\sD}^{-1}}-\frac{1}{\sqrt{\lambda_0(1+1/\lambda_{\min}(\sD))}}\right)\right)
\end{aligned}\end{equation*}
therefore, one form of probability is:
\begin{equation*}
    \delta_1=|\sA|\exp\left(\frac{\beta_0}{\eta}\left(\frac{\|\phi(x,\bar{a})\|_{\Sigma_\sD^{-1}}}{\sqrt{1-1/\lambda_{\min}(\sD)}}- \frac{1}{\sqrt{\lambda_0(1+1/\lambda_{\min}(\sD))}}\right) \right).
\end{equation*}
similarly, \eqref{eq:contextual_bandit_low_prob_event_2} yields:
\begin{equation*}\begin{aligned}
    &\Pr_{a\sim\widetilde{\pi}(\cdot;\sD_-)}\left(\|\phi(x,a)\|_{\Sigma_{\sD_-}^{-1}}>\frac{1}{\sqrt{\lambda_0}}\right )\\
    \leq& |\sA|\exp\left(\sqrt{\frac{(n-1)d}{\lambda_{\min}(\sD)-1}}\frac{R}{\eta} +\sigma+\frac{\beta_0}{\eta} (\min_a \|\phi(x,a)\|_{\Sigma_{\sD_-}^{-1}}-\frac{1}{\sqrt{\lambda_0}})\right)\\
    \leq& |\sA|\exp\left(\sqrt{\frac{(n-1)d}{\lambda_{\min}(\sD)-1}}\frac{R}{\eta} +\sigma+\frac{\beta_0}{\eta} (\frac{\min_a \|\phi(x,a)\|_{\Sigma_{\sD}^{-1}}}{\sqrt{1-1/\lambda_{\min}(\sD)}}-\frac{1}{\sqrt{\lambda_0}})\right)
\end{aligned}\end{equation*}
and
\begin{equation*}
    \begin{aligned}
        &\Pr_{a\sim\widetilde{\pi}(\cdot;\sD_+)}\left(\|\phi(x,a)\|_{\Sigma_{\sD_+}^{-1}}>\frac{1}{\sqrt{\lambda_0(1+1/\lambda_{\min}(\sD))}}\right )\\
        \leq & |\sA|\exp\left(\sqrt{\frac{(n-1)d}{\lambda_{\min}(\sD)-1}}\frac{R}{\eta} +\sigma+\frac{\beta_0}{\eta} (\min_a \|\phi(x,a)\|_{\Sigma_{\sD_+}^{-1}}-\frac{1}{\sqrt{\lambda_0(1+1/\lambda_{\min}(\sD))}})\right)\\
        \leq & |\sA|\exp\left(\sqrt{\frac{(n-1)d}{\lambda_{\min}(\sD)-1}}\frac{R}{\eta} +\sigma+\frac{\beta_0}{\eta} (\frac{\min_a \|\phi(x,a)\|_{\Sigma_{\sD}^{-1}}}{\sqrt{1-1/\lambda_{\min}(\sD)}}-\frac{1}{\sqrt{\lambda_0(1+1/\lambda_{\min}(\sD))}})\right)
    \end{aligned}
\end{equation*}
which leads to another form of $\delta$:
\begin{equation*}
    \delta_2=|\sA|\exp\left(\sqrt{\frac{(n-1)d}{\lambda_{\min}(\sD)-1}}\frac{R}{\eta} +\sigma+\frac{\beta_0}{\eta} (\frac{\min_a \|\phi(x,a)\|_{\Sigma_{\sD}^{-1}}}{\sqrt{1-1/\lambda_{\min}(\sD)}}-\frac{1}{\sqrt{\lambda_0(1+1/\lambda_{\min}(\sD))}})\right)
\end{equation*}

For all $a\in\mathcal{I}$ the sensitivity is bounded by:
\begin{equation*}\begin{aligned}
    \Delta=&\max_{x,a}\max_{\sD;\sD'\in\{\sD_+,\sD_-\}}|(\bar{r}(x,a;\sD)-\bar{r}(x,a;\sD'))-\beta_0(\Gamma(x,a;\sD')-\Gamma(x,a;\sD))|\\
    \leq &(1+\sqrt{\frac{nd}{\lambda_{\min}(\sD)-1}})\frac{R}{\sqrt{\lambda_{\min}(\sD)}}\|\phi(x,a)\|_{\Sigma_{\sD}^{-1}}+\frac{\beta_0}{2(\lambda_{\min}(\sD)-1)}\|\phi(x,a)\|_{\Sigma_{\sD}^{-1}}\\
    \leq &\left((1+\sqrt{\frac{nd}{\lambda_{\min}(\sD)-1}})\frac{R}{\sqrt{\lambda_{\min}(\sD)}}+\frac{\beta_0}{2(\lambda_{\min}(\sD)-1)}\right)\frac{1}{\sqrt{\lambda_0}},
\end{aligned}\end{equation*}
Therefore, we can use \Cref{lem:KL_regularization_privacy} to obtain $(\epsilon,\delta)$-DP, where
\begin{equation*}
    \delta=\min\{\delta_1,\delta_2\},
\end{equation*}
and
\begin{equation*}
    \epsilon=\frac{1}{\eta\sqrt{\lambda_0}}\left((1+\sqrt{\frac{nd}{\lambda_{\min}(\sD)-1}})\frac{2R}{\sqrt{\lambda_{\min}(\sD)}}+\frac{\beta_0}{\lambda_{\min}(\sD)-1}\right)-\log(1-\delta).
\end{equation*}
This completes the proof of the approximate DP part of \Cref{thm:linear_contextual_bandits_privacy}. For the pure DP part, one can simply take $\lambda_0=\lambda_{\min}(\sD)$ to eliminate the low probability event.

\section{Proof of \Cref{thm:RLHF_privacy}}\label{app:proof_RLHF_privacy}
We first state a generalized version of \Cref{thm:RLHF_privacy} which includes both conventional DP and label DP:
\begin{theorem}[\Cref{thm:RLHF_privacy} Restated.]
    Fix state $x$, the action sampled from $\widetilde{\pi}(\cdot|x;\sD)$ is $\epsilon_0$-differentially private and $\epsilon_0'$-label differentially private for
    \begin{align*}
        \epsilon_0'=&\frac{1}{\eta\sqrt{\lambda_{\min}(\sD)}}\left(\frac{(1+\exp(2B))\sqrt{\lambda_{\min}(\sD)}}{\lambda_{\min}(\sD)-\lambda}+\frac{2\beta_0}{\lambda_{\min}(\sD)-1}\right);\\
        \epsilon_0=&\frac{1}{\eta\sqrt{\lambda_{\min}(\sD)}}\left(\frac{(2+2\exp(2B))\sqrt{\lambda_{\min}(\sD)}}{\lambda_{\min}(\sD)-\lambda}\right).
    \end{align*}
    Moreover, for arbitrarily chosen $\lambda_0> \lambda$, it is $(\epsilon,\delta)$-DP with
    \begin{align*}
        \epsilon'=&\frac{1}{\eta\sqrt{\lambda_0}}\left(\frac{(1+\exp(2B))\sqrt{\lambda_{\min}(\sD)}}{\lambda_{\min}(\sD)-\lambda}+\frac{2\beta_0}{\lambda_{\min}(\sD)-1}\right)-\log(1-\delta);\\
        \delta'=&|\sA|\exp\left(\min\{C_5',C_6'\}-\frac{\beta_0}{\eta}\frac{1}{\sqrt{\lambda_0(1+4/\lambda_{\min}(\sD))}} \right).
    \end{align*}
    where
    \begin{align*}
        C'_5=&\frac{\beta_0}{\eta}\frac{\|\phi(x,\bar{a})\|_{\Sigma_\sD^{-1}}}{\sqrt{1-4/\lambda_{\min}(\sD)}};\\
        C'_6=&\frac{2B}{\eta}+\sigma+\frac{\beta_0}{\eta}\frac{\min_a\|\phi(x,a)\|_{\Sigma_\sD^{-1}}}{\sqrt{1-4/\lambda_{\min}(\sD)}}.
    \end{align*}
    Also, it is $(\epsilon',\delta')$-label DP with
    \begin{align*}
        \epsilon=&\frac{1}{\eta\sqrt{\lambda_0}}\left(\frac{(2+2\exp(2B))\sqrt{\lambda_{\min}(\sD)}}{\lambda_{\min}(\sD)-\lambda}\right)-\log(1-\delta);\\
        \delta=&|\sA|\exp\left(\min\{C_5,C_6\}-\frac{\beta_0}{\eta}\frac{1}{\sqrt{\lambda_0}} \right).
    \end{align*}
    where
    \begin{align*}
        C_5=&\frac{\beta_0}{\eta}\|\phi(x,\bar{a})\|_{\Sigma_\sD^{-1}};\\
        C_6=&\frac{2B}{\eta}+\sigma+\frac{\beta_0}{\eta}\min_a\|\phi(x,a)\|_{\Sigma_\sD^{-1}}.
    \end{align*}
    Here $\bar{a}=\argmax_{a'}\{r_{MLE}(x,a';\sD)+\eta\log \pi_0(a'|x)\}$.
\end{theorem}

For add-remove DP, similar to the proof of \Cref{thm:linear_contextual_bandits_privacy}, we still consider add-remove DP. We use the same notation $\sD$ for the original dataset and $\sD_+=\sD\cup (x_0,a_0^1,a_0^2,y_0)$ for addition DP. We first present our results to addition DP and then generalize to removal DP.

\paragraph{Bounding the sensitivity of the MLE estimator.} Here we bound the sensitivity of $r_{MLE}(x,a;\sD)=\theta_{MLE}(\sD)^T\phi(x,a)$. Recall the log likelihood of $\theta$ given $\sD$:
\begin{equation*}
    \begin{aligned}
        &\ell_\mathcal{D}(\theta)=\sum_{\mathcal{D}}\left[y\log\left(\sigma(r_\theta(x,a^1)-r_\theta(x,a^2))\right)+(1-y)\log\left(\sigma(r_\theta(x,a^2)-r_\theta(x,a^1))\right)\right]\\
&=\sum_{\mathcal{D}}\left[y_i\log\left(\sigma(\langle \theta, \phi(x_i,a_i^1)-\phi(x_i,a_i^2) \rangle)\right)+(1-y_i)\log\left(\sigma(\langle \theta, \phi(x_i,a_i^2)-\phi(x_i,a_i^1) \rangle)\right)\right];
    \end{aligned}
\end{equation*}
its gradient is in the form:
\begin{equation*}\begin{aligned}
    &\nabla \ell_{\mathcal{D}}(\theta)=\\
    &\sum_{\mathcal{D}} \left(\frac{y_i}{1+\exp(\langle \theta, \phi(x_i,a_i^1)-\phi(x_i,a_i^2) \rangle)}-\frac{1-y_i}{1+\exp(\langle \theta, \phi(x_i,a_i^2)-\phi(x_i,a_i^1) \rangle)}\right)(\phi(x_i,a_i^1)-\phi(x_i,a_i^2))
\end{aligned}\end{equation*}
at $\theta_{MLE}(\sD)$, the gradient of $\ell_\sD(\theta)$ should be zero, that is,
\begin{equation}\label{eq:MLE_grad_zero}
    \nabla\ell_\sD(\theta_{MLE}(\sD))=0.
\end{equation}
Observe that the Hessian of $\ell_\sD$ is:
\begin{equation*}
    \begin{aligned}
        &\nabla^2 \ell_\sD(\theta)=\\
    -&\sum_{\mathcal{D}} \left(\frac{y_i\exp(\langle \theta, \phi(x_i,a_i^1)-\phi(x_i,a_i^2) \rangle)}{(1+\exp(\langle \theta, \phi(x_i,a_i^1)-\phi(x_i,a_i^2) \rangle))^2}+\frac{(1-y_i)\exp(\langle \theta, \phi(x_i,a_i^2)-\phi(x_i,a_i^1) \rangle)}{(1+\exp(\langle \theta, \phi(x_i,a_i^2)-\phi(x_i,a_i^1) \rangle))^2}\right)\\
    &(\phi(x_i,a_i^1)-\phi(x_i,a_i^2))(\phi(x_i,a_i^1)-\phi(x_i,a_i^2))^T\\
    =&-\sum_{\mathcal{D}}\frac{\exp(\langle \theta, \phi(x_i,a_i^1)-\phi(x_i,a_i^2) \rangle)}{(1+\exp(\langle \theta, \phi(x_i,a_i^1)-\phi(x_i,a_i^2) \rangle))^2}(\phi(x_i,a_i^1)-\phi(x_i,a_i^2))(\phi(x_i,a_i^1)-\phi(x_i,a_i^2))^T;
    \end{aligned}
\end{equation*}
with the condition $\max_{(x,a)}\|\phi(x,a)\|_2\leq 1$ and $\|\theta\|_2\leq B$, we have
\begin{equation*}
    \frac{\exp(\langle \theta,(\phi(x_i,a_i^1)-\phi(x_i,a_i^2))\rangle)}{(1+\exp(\langle \theta, \phi(x_i,a_i^1)-\phi(x_i,a_i^2) \rangle))^2}\geq \frac{1}{2+\exp(-2B)+\exp(2B)}:=\gamma.
\end{equation*}
Therefore, we can see that $\ell_\sD$ is strongly concave with
\begin{equation}\label{eq:MLE_strong_concavity}
    \nabla^2 \ell_\sD(\theta)\leq -\gamma\sum_{\mathcal{D}}(\phi(x_i,a_i^1)-\phi(x_i,a_i^2))(\phi(x_i,a_i^1)-\phi(x_i,a_i^2))^T.
\end{equation}
When context is clear we use the abbreviate notation $\theta(\sD)$ to denote $\theta_{MLE}(\sD)$, to bound the sensitivity of $r_{MLE}$, notice that
\begin{equation*}
    r_{MLE}(x,a;\sD_+)-r_{MLE}(x,a;\sD)=(\theta(\sD_+)-\theta(\sD))^T\phi(x,a),
\end{equation*}
it suffices to bound $\theta(\sD_+)-\theta(\sD)$, which is given by $\ell_{\sD_+}(\theta)$ and $\ell_\sD(\theta)$. We can write $\ell_{\sD_+}(\theta)$ and its gradient using $\ell_\sD(\theta)$ as:
\begin{equation*}
    \ell_{\sD_+}(\theta)=\ell_\sD(\theta)+\left[y_0\log\left(\sigma(\langle \theta, \phi(x_0,a_0^1)-\phi(x_0,a_0^2) \rangle)\right)+(1-y_0)\log\left(\sigma(\langle \theta, \phi(x_0,a_0^2)-\phi(x_0,a_0^1) \rangle)\right)\right],
\end{equation*}
and
\begin{equation}\label{eq:loglikelihood_grad_difference}\begin{aligned}
    &\nabla \ell_{\sD_+}(\theta)
    =\nabla \ell_\sD(\theta)\\
    &+\left(\frac{y_0}{1+\exp(\langle \theta, \phi(x_0,a_0^1)-\phi(x_0,a_0^2) \rangle)}-\frac{1-y_0}{1+\exp(\langle \theta, \phi(x_0,a_0^2)-\phi(x_0,a_0^1) \rangle)}\right)(\phi(x_0,a_0^1)-\phi(x_0,a_0^2)).
\end{aligned}\end{equation}
Consider the Taylor series expansion of $\langle \nabla \ell_{\sD_+}(\theta(\sD_+)), v\rangle$ for some vector $v$ we have:
\begin{equation}\label{eq:loglikelihood_Taylor_expansion}
    \langle \nabla \ell_{\sD_+}(\theta(\sD_+)), v\rangle=\langle \nabla \ell_{\sD_+}(\theta(\sD)), v\rangle+(\theta(\sD_+)-\theta(\sD))^T \nabla^2 \ell_{\sD_+}(\theta') v,
\end{equation}
where $\theta'$ is a convex combination of $\theta(\sD_+)$ and $\theta(\sD)$.
Using strong concavity \eqref{eq:MLE_strong_concavity}, we obtain:
\begin{equation*}
     \nabla^2 \ell_{\sD_+}(\theta') \leq-\gamma\sum_{\sD_+}(\phi(x_i,a_i^1)-\phi(x_i,a_i^2))(\phi(x_i,a_i^1)-\phi(x_i,a_i^2))^T=-\gamma (\Sigma_{\sD_+}-\lambda I).
\end{equation*}
Therefore, rearranging \eqref{eq:loglikelihood_Taylor_expansion} we get:
\begin{equation*}
\begin{aligned}
    &\gamma(\theta(\sD_+)-\theta(\sD))^T\left(\Sigma_{\sD_+}-\lambda I\right)v\\
    \leq & \langle \nabla \ell_{\sD_+}(\theta(\sD))-\nabla \ell_{\sD_+}(\theta(\sD_+)),v \rangle\\
    \stackrel{\text{(i)}}{\leq}& \langle \left(\nabla \ell_{\sD_+}-\nabla \ell_\sD\right)\left(\theta(\sD)\right), v\rangle\\
    \stackrel{\text{(ii)}}{=}&\left(\frac{y_0}{1+\exp(\langle \theta, \phi(x_0,a_0^1)-\phi(x_0,a_0^2) \rangle)}-\frac{1-y_0}{1+\exp(\langle \theta, \phi(x_0,a_0^2)-\phi(x_0,a_0^1) \rangle)}\right)\langle \phi(x_0,a_0^1)-\phi(x_0,a_0^2),v\rangle,
\end{aligned}\end{equation*}
where we have used the facts \eqref{eq:MLE_grad_zero} in (i) and \eqref{eq:loglikelihood_grad_difference} in (ii).
Taking $v=\left(\Sigma_{\sD_+}-\lambda I\right)^{-1} \phi(x,a)$, we obtain:
\begin{equation}\begin{aligned}
    &\gamma(\theta(\sD_+)-\theta(\sD))^T\phi(x,a)\\
    \leq & \langle \left(\nabla \ell_{\sD_+}-\nabla \ell_\sD\right)(\theta(\sD)), \left(\Sigma_{\sD_+}-\lambda I\right)^{-1} \phi(x,a)\rangle\\
    = & \left(\frac{y_0}{1+\exp(\langle \theta, \phi(x_0,a_0^1)-\phi(x_0,a_0^2) \rangle)}-\frac{1-y_0}{1+\exp(\langle \theta, \phi(x_0,a_0^2)-\phi(x_0,a_0^1) \rangle)}\right)\\
    &\langle \phi(x_0,a_0^1)-\phi(x_0,a_0^2),\phi(x,a)\rangle_{(\Sigma_{\sD_+}-\lambda I)^{-1}}\\
    \leq & \frac{\langle \phi(x_0,a_0^1)-\phi(x_0,a_0^2),\phi(x,a)\rangle_{(\Sigma_{\sD_+}-\lambda I)^{-1}}}{1+\exp(-2B)}.
\end{aligned}\end{equation}
This leads to the final bound of
\begin{equation}\label{eq:rMLE_sensitivity_addition}\begin{aligned}
    &|r_{MLE}(x,a;\sD_+)-r_{MLE}(x,a;\sD)|\\&\leq \frac{2+\exp(-2B)+\exp(2B)}{1+\exp(-2B)}|\langle \phi(x_0,a_0^1)-\phi(x_0,a_0^2),\phi(x,a)\rangle_{(\Sigma_{\sD_+}-\lambda I)^{-1}}|\\
    &=(1+\exp(2B))|\langle \phi(x_0,a_0^1)-\phi(x_0,a_0^2),\phi(x,a)\rangle_{(\Sigma_{\sD_+}-\lambda I)^{-1}}|\\
    &\leq (1+\exp(2B))\|\phi(x_0,a_0^1)-\phi(x_0,a_0^2)\|_{(\Sigma_{\sD_+}-\lambda I)^{-1}}\|\phi(x,a)\|_{(\Sigma_{\sD_+}-\lambda I)^{-1}}\\
    &\leq (1+\exp(2B))\|\phi(x_0,a_0^1)-\phi(x_0,a_0^2)\|_{(\Sigma_{\sD}-\lambda I)^{-1}}\|\phi(x,a)\|_{(\Sigma_{\sD}-\lambda I)^{-1}}\\
    &\leq \frac{1+\exp(2B)}{\sqrt{\lambda_{\min}(\sD)-\lambda}}\|\phi(x,a)\|_{(\Sigma_{\sD}-\lambda I)^{-1}}\\
    &\leq \frac{(1+\exp(2B))\sqrt{\lambda_{\min}(\sD)}}{\lambda_{\min}(\sD)-\lambda}\|\phi(x,a)\|_{\Sigma_{\sD}^{-1}}.
\end{aligned}\end{equation}
where the last inequality holds due to:
\begin{equation*}
    \begin{aligned}
        &\phi(x,a)^T(\Sigma_{\sD}-\lambda I)^{-1}\phi(x,a)\\
        \leq &\phi(x,a)^T\Sigma_{\sD}^{-1}(I-\lambda\Sigma_\sD^{-1})^{-1}\phi(x,a)\\
        \leq &\frac{1}{1-\lambda/\lambda_{\min}(\sD)}\phi(x,a)^T\Sigma_{\sD}^{-1}\phi(x,a)\\
        =&\frac{\lambda_{\min}(\sD)}{\lambda_{\min}(\sD)-\lambda}\phi(x,a)^T\Sigma_{\sD}^{-1}\phi(x,a).
    \end{aligned}
\end{equation*}
This bounds the sensitivity of $r_{MLE}$ in the addition case. In the removal case, simply use \eqref{eq:rMLE_sensitivity_removal} on $\sD_-=\sD\backslash (x_i,a_i^1,a_i^2,y_i)$ and $\sD$, we have:
\begin{equation}\label{eq:rMLE_sensitivity_removal}
    \begin{aligned}
        &|r_{MLE}(x,a;\sD_-)-r_{MLE}(x,a;\sD)|\\
        \leq & (1+\exp(2B))\|\phi(x_i,a_i^1)-\phi(x_i,a_i^2)\|_{(\Sigma_{\sD}-\lambda I)^{-1}}\|\phi(x,a)\|_{(\Sigma_{\sD}-\lambda I)^{-1}}\\
        \leq & \frac{(1+\exp(2B))\sqrt{\lambda_{\min}(\sD)}}{\lambda_{\min}(\sD)-\lambda}\|\phi(x,a)\|_{\Sigma_{\sD}^{-1}}.
    \end{aligned}
\end{equation}

\paragraph{Bounding the sensitivity of $\Gamma$.} Similar to that in the proof of \Cref{thm:linear_contextual_bandits_privacy}, we have:
\begin{equation}\label{eq:RLHF_Gamma_sensitivity_addition}\begin{aligned}
    &\Gamma(x,a;\sD)-\Gamma(x,a;\sD_+)\\
    =&\frac{\langle \phi(x,a),\phi(x_0,a_0^1)-\phi(x_0,a_0^2)\rangle_{\Sigma_{\sD_+}^{-1}}}{\|\phi(x,a)\|_{\Sigma_\sD^{-1}}+\|\phi(x,a)\|_{\Sigma_{\sD_+}^{-1}}}\langle \phi(x,a),\phi(x_0,a_0^1)-\phi(x_0,a_0^2)\rangle_{\Sigma_{\sD}^{-1}}\\
    \leq&\frac{\|\phi(x,a)\|_{\Sigma_{\sD_+}^{-1}}\|\phi(x_0,a_0^1)-\phi(x_0,a_0^2)\|_{\Sigma_{\sD_+}^{-1}}}{2\|\phi(x,a)\|_{\Sigma_{\sD_+}^{-1}}}\|\phi(x,a)\|_{\Sigma_{\sD}^{-1}}\|\phi(x_0,a_0^1)-\phi(x_0,a_0^2)\|_{\Sigma_{\sD}^{-1}}\\
    \leq& \frac{2}{\lambda_{\min}(\sD)}\|\phi(x,a)\|_{\Sigma_\sD^{-1}},
\end{aligned}\end{equation}
and similarly,
\begin{equation}\label{eq:RLHF_Gamma_sensitivity_removal}
    \Gamma(x,a;\sD_-)-\Gamma(x,a;\sD)\leq \frac{2}{\lambda_{\min}(\sD)-1}\|\phi(x,a)\|_{\Sigma_\sD^{-1}}.
\end{equation}

\paragraph{Establishing the low-probability event.} Similar to that in linear bandits, we bound the probability of the event $\|\phi(x,a)\|_{\Sigma_{\sD'}^{-1}}\geq \frac{1}{\sqrt{\lambda_0}}$ under both label DP and conventional DP. For all possible neighboring dataset $\sD'$ to $\sD$ under the setting of either conventional or label DP, we have:
\begin{equation}\label{eq:proof_RLHF_low_prob_1}\begin{aligned}
    &\Pr_{a\sim\widetilde{\pi}(\cdot|x,{\sD'})}(\|\phi(x,a)\|_{\Sigma_{\sD'}^{-1}}> \frac{1}{\sqrt{\lambda_0}})\\
    \leq & \frac{\sum_a \pi_0(a|x) \mathbb{I}[\|\phi(x,a)\|_{\Sigma_{\sD'}^{-1}}> \frac{1}{\sqrt{\lambda_0}}]\exp\left((r_{MLE}(x,a;\sD')-\beta_0\|\phi(x,a)\|_{\Sigma_{\sD'}^{-1}})/\eta\right)}{\sum_a \pi_0(a|x)\exp\left((r_{MLE}(x,a;\sD')-\beta_0\|\phi(x,a)\|_{\Sigma_{\sD'}^{-1}})/\eta\right)}\\
    \leq & \frac{|\sA|\exp\left(\frac{1}{\eta}(\max_{a'}\{r_{MLE}(x,a';\sD')+\eta\log \pi_0(a'|x)\}-\beta_0\frac{1}{\sqrt{\lambda_0}})\right)}{\exp\left(\frac{1}{\eta}\max_{\hat{a}}\{r_{MLE}(x,\hat{a};\sD')+\eta\log \pi_0(\hat{a}|x)-\beta_0\|\phi(x,\hat{a})\|_{\Sigma_{\sD'}^{-1}}\}\right)}\\
    = & |\sA|\exp(\frac{1}{\eta}(\max_{a'}\{r_{MLE}(x,a';\sD')+\eta\log \pi_0(a'|x)\}-\beta_0\frac{1}{\sqrt{\lambda_0}}\\
    &-\max_{\hat{a}}\{r_{MLE}(x,\hat{a};\sD')+\eta\log \pi_0(\hat{a}|x)-\beta_0\|\phi(x,\hat{a})\|_{\Sigma_{\sD'}^{-1}}\}))\\
    \leq & |\sA|\exp\left(\frac{\beta_0}{\eta}\left( \|\phi(x,\bar{a})\|_{\Sigma_{\sD'}^{-1}}-\frac{1}{\sqrt{\lambda_0}}\right)\right),
\end{aligned}\end{equation}
where $\bar{a}=\argmax_{a'}\{r_{MLE}(x,a';\sD')+\eta\log \pi_0(a'|x)\}$, or using the alternative bound of:
\begin{equation}\label{eq:proof_RLHF_low_prob_2}\begin{aligned}
    &\Pr_{a\sim\widetilde{\pi}(\cdot|x,{\sD'})}(\|\phi(x,a)\|_{\Sigma_{\sD'}^{-1}}> \frac{1}{\sqrt{\lambda_0}})\\
    \leq & \frac{\sum_a \pi_0(a|x) \mathbb{I}[\|\phi(x,a)\|_{\Sigma_{\sD'}^{-1}}> \frac{1}{\sqrt{\lambda_0}}]\exp\left((r_{MLE}(x,a;\sD')-\beta_0\|\phi(x,a)\|_{\Sigma_{\sD'}^{-1}})/\eta\right)}{\sum_a \pi_0(a|x)\exp\left((r_{MLE}(x,a;\sD')-\beta_0\|\phi(x,a)\|_{\Sigma_{\sD'}^{-1}})/\eta\right)}\\
    \leq & \frac{|\sA|\exp\left(\frac{1}{\eta}(\max_{a'}\{r_{MLE}(x,a';\sD')+\eta\log \pi_0(a'|x)\}-\beta_0\frac{1}{\sqrt{\lambda_0}})\right)}{\exp\left(\frac{1}{\eta}\max_{\hat{a}}\{r_{MLE}(x,\hat{a};\sD')+\eta\log \pi_0(\hat{a}|x)-\beta_0\|\phi(x,\hat{a})\|_{\Sigma_{\sD'}^{-1}}\}\right)}\\
    = & |\sA|\exp(\frac{1}{\eta}(\max_{a'}\{r_{MLE}(x,a';\sD')+\eta\log \pi_0(a'|x)\}-\beta_0\frac{1}{\sqrt{\lambda_0}}\\
    &-\max_{\hat{a}}\{r_{MLE}(x,\hat{a};\sD')+\eta\log \pi_0(\hat{a}|x)-\beta_0\|\phi(x,\hat{a})\|_{\Sigma_{\sD'}^{-1}}\}))\\
    \leq & |\sA|\exp\left( \frac{2B}{\eta}+\sigma+\frac{\beta_0}{\eta}(\min_a\|\phi(x,a)\|_{\Sigma_{\sD'}^{-1}}-\frac{1}{\sqrt{\lambda_0}})\right),
\end{aligned}\end{equation}

\paragraph{Obtaining the final bound.} 
For conventional DP, notice that:
\begin{equation*}
    \|\phi(x,a)\|_{\Sigma_\sD^{-1}}\leq \|\phi(x,a)\|_{\Sigma_{\sD_-}^{-1}}\leq \frac{1}{\sqrt{1-\|\phi(x_i,a_i^1)-\phi(x_i,a_i^2)\|^2_{\Sigma_{\sD}^{-1}}}}\|\phi(x,a)\|_{\Sigma_\sD^{-1}}
\end{equation*}
and
\begin{equation*}
    \frac{1}{\sqrt{1+\|\phi(x_0,a_0^1)-\phi(x_0,a_0^2)\|^2_{\Sigma_{\sD}^{-1}}}}\|\phi(x,a)\|_{\Sigma_\sD^{-1}}\leq \|\phi(x,a)\|_{\Sigma_{\sD_+}^{-1}}\leq \|\phi(x,a)\|_{\Sigma_\sD^{-1}},
\end{equation*}
which yields:
\begin{equation*}
    \|\phi(x,a)\|_{\Sigma_{\sD_-}^{-1}}\leq \frac{\|\phi(x,a)\|_{\Sigma_\sD^{-1}}}{\sqrt{1-\|\phi(x_i,a_i^1)-\phi(x_i,a_i^2)\|^2_{\Sigma_{\sD}^{-1}}}}\leq \frac{\|\phi(x,a)\|_{\Sigma_\sD^{-1}}}{\sqrt{1-4/\lambda_{\min}(\sD)}}
\end{equation*}
and
\begin{equation*}
    \frac{\|\phi(x,a)\|_{\Sigma_\sD^{-1}}}{\sqrt{1+4/\lambda_{\min}(\sD)}}\leq\frac{\|\phi(x,a)\|_{\Sigma_\sD^{-1}}}{\sqrt{1+\|\phi(x_0,a_0^1)-\phi(x_0,a_0^2)\|^2_{\Sigma_{\sD}^{-1}}}}\leq \|\phi(x,a)\|_{\Sigma_{\sD_+}^{-1}}
\end{equation*}
using \eqref{eq:proof_RLHF_low_prob_1}, we have:
\begin{equation*}
    \begin{aligned}
        \Pr\left(\widetilde{\pi}(\cdot;\sD_-)\in \bar{\mathcal{I}} \right)\leq&\Pr_{a\sim\widetilde{\pi}(\cdot;\sD_-)}\left(\|\phi(x,a)\|_{\Sigma_{\sD_-}^{-1}}>\frac{1}{\sqrt{\lambda_0}}\right )\\
        \leq & |\sA|\exp\left(\frac{\beta_0}{\eta}\left( \|\phi(x,\bar{a})\|_{\Sigma_{\sD_-}^{-1}}-\frac{1}{\sqrt{\lambda_0}}\right)\right)\\
        \leq & |\sA|\exp\left(\frac{\beta_0}{\eta}\left( \frac{\|\phi(x,\bar{a})\|_{\Sigma_\sD^{-1}}}{\sqrt{1-4/\lambda_{\min}(\sD)}}-\frac{1}{\sqrt{\lambda_0}}\right)\right)
    \end{aligned}
\end{equation*}
and
\begin{equation*}
    \begin{aligned}
        \Pr\left(\widetilde{\pi}(\cdot;\sD_+)\in \bar{\mathcal{I}} \right)\leq & \Pr_{a\sim\widetilde{\pi}(\cdot;\sD_+)}\left(\|\phi(x,a)\|_{\Sigma_{\sD_+}^{-1}}>\frac{1}{\sqrt{\lambda_0(1+4/\lambda_{\min}(\sD))}}\right )\\
        \leq & |\sA|\exp\left(\frac{\beta_0}{\eta}\left(\|\phi(x,\bar{a})\|_{\Sigma_{\sD_+}^{-1}}-\frac{1}{\sqrt{\lambda_0(1+4/\lambda_{\min}(\sD))}}\right)\right)\\
    \leq & |\sA|\exp\left(\frac{\beta_0}{\eta}\left(\|\phi(x,\bar{a})\|_{\Sigma_{\sD}^{-1}}-\frac{1}{\sqrt{\lambda_0(1+4/\lambda_{\min}(\sD))}}\right)\right)
    \end{aligned}
\end{equation*}
therefore, we obtain a probability upper bound:
\begin{equation*}
    \delta_1=|\sA|\exp\left(\frac{\beta_0}{\eta}\left(\frac{\|\phi(x,\bar{a})\|_{\Sigma_\sD^{-1}}}{\sqrt{1-4/\lambda_{\min}(\sD)}}-\frac{1}{\sqrt{\lambda_0(1+4/\lambda_{\min}(\sD))}}\right)\right)
\end{equation*}
and similarly, using \eqref{eq:proof_RLHF_low_prob_2}, we obtain another upper bound:
\begin{equation*}
    \delta_2=|\sA|\exp\left(\frac{2B}{\eta}+\sigma+\frac{\beta_0}{\eta}\left(\frac{\min_a\|\phi(x,a)\|_{\Sigma_\sD^{-1}}}{\sqrt{1-4/\lambda_{\min}(\sD)}}-\frac{1}{\sqrt{\lambda_0(1+4/\lambda_{\min}(\sD))}}\right)\right)
\end{equation*}

We can use the same approach as in the proof of \Cref{thm:linear_contextual_bandits_privacy} to achieve the $(\epsilon,\delta)$-privacy guarantee through \eqref{eq:rMLE_sensitivity_addition}, \eqref{eq:rMLE_sensitivity_removal}, \eqref{eq:RLHF_Gamma_sensitivity_addition} and \eqref{eq:RLHF_Gamma_sensitivity_removal} that
\begin{equation*}
    \delta=\min\{\delta_1,\delta_2\};
\end{equation*}
and
\begin{equation*}
    \epsilon=\frac{1}{\eta\sqrt{\lambda_0}}\left(\frac{(1+\exp(2B))\sqrt{\lambda_{\min}(\sD)}}{\lambda_{\min}(\sD)-\lambda}+\frac{2\beta_0}{(\lambda_{\min}(\sD)-1)}\right)-\log(1-\delta).
\end{equation*}

If we consider label DP instead, notice that $\Gamma(x,a;\sD)=\Gamma(x,a;\sD')$ for $\sD$ and $\sD'$ differing in only one label, meaning that we don't have the sensitivity term in $\Gamma$ any more. For the low probability event, we have $\|\phi(x,a)\|_{\Sigma_{\sD'}^{-1}}=\|\phi(x,a)\|_{\Sigma_{\sD}^{-1}}, \forall (x,a)$ and therefore, we have:
\begin{equation*}
    \delta_1=|\sA|\exp\left(\frac{\beta_0}{\eta}\left(\|\phi(x,\bar{a})\|_{\Sigma_\sD^{-1}}-\frac{1}{\sqrt{\lambda_0}}\right)\right)
\end{equation*}
and
\begin{equation*}
    \delta_2=|\sA|\exp\left(\frac{2B}{\eta}+\sigma+\frac{\beta_0}{\eta}\left(\min_a\|\phi(x,a)\|_{\Sigma_\sD^{-1}}-\frac{1}{\sqrt{\lambda_0}}\right)\right)
\end{equation*}

Removing corresponding terms and regarding the label change as first deleting an entry and then adding back the entry with flipped label yields the result.

\section{Proof of \Cref{prop:inexact_policy_optimization}}\label{app:proof_inexact_policy_optimization}
The condition that $D_{\infty}(\widetilde{\pi}(\cdot;\sD)\|\widehat{\pi}(\cdot;\sD))\leq \widehat{\epsilon}$ and $D_{\infty}(\widehat{\pi}(\cdot;\sD)\|\widetilde{\pi}(\cdot;\sD))\leq \widehat{\epsilon}$ yields:
\begin{equation*}
    \exp(-\widehat{\epsilon}) \widehat{\pi}(a;\sD)\leq \widetilde{\pi}(a;\sD)\leq \exp(\widehat{\epsilon}) \widehat{\pi}(a;\sD), \forall a\in \sA
\end{equation*}
and the condition that sampling from $\widetilde{\pi}$ is $(\epsilon,\delta)$-differentially private yields, for neighboring dataset $\sD,\sD'$ and arbitrary subset of arms $\sS\subseteq \sA$:
\begin{equation*}\begin{aligned}
    \sum_{a\in \sS}\widetilde{\pi}(a;\sD)=&\Pr_{a\sim \widetilde{\pi}(\cdot;\sD)}\left(a\in \sS\right)\\
    \leq& \delta+\exp(\epsilon)\Pr_{a\sim \widetilde{\pi}(\cdot;\sD')}\left(a\in \sS\right) \\
    =&\delta+\exp(\epsilon)\left(\sum_{a\in \sS}\widetilde{\pi}(a;\sD')\right)
\end{aligned}\end{equation*}
Therefore, we have:
\begin{equation*}\begin{aligned}
    \Pr_{a\sim\widehat{\pi}(\cdot|\sD)}(a\in\sS)=&\sum_{a\in \sS}\widehat{\pi}(a;\sD)\\
    \leq &\exp(\widehat{\epsilon})\sum_{a\in \sS}\widetilde{\pi}(a;\sD)\\
    \leq &\exp(\widehat{\epsilon})\left(\delta+\exp(\epsilon)\left(\sum_{a\in \sS}\widetilde{\pi}(a;\sD')\right)\right)\\
    \leq &\exp(\widehat{\epsilon})\left(\delta+\exp(\epsilon)\left(\exp(\widehat{\epsilon})\sum_{a\in \sS}\widehat{\pi}(a;\sD')\right)\right)\\
    \leq &\exp(\epsilon+2\widehat{\epsilon})\Pr_{a\sim\widehat{\pi}(\cdot|\sD')}(a\in\sS)+\exp(\widehat{\epsilon})\delta
\end{aligned}\end{equation*}
which shows that sampling from $\widehat{\pi}$ is $(\epsilon+2\widehat{\epsilon},\exp(\widehat{\epsilon})\delta)$-differentially private.


\end{document}